\documentclass{article}

\newcommand{\overbar}[1]{\mkern 1.5mu\overline{\mkern-1.5mu#1\mkern-1.5mu}\mkern 1.5mu}
\usepackage{svg}
\PassOptionsToPackage{numbers, compress}{natbib}
\usepackage{float}
\usepackage{times,graphicx}
\usepackage{relsize}
\usepackage{xcolor}
\usepackage{soul}
\usepackage[inline]{enumitem}
\usepackage{wrapfig}
\usepackage{subcaption}
\usepackage{todonotes}
\usepackage{color}

\usepackage[symbol]{footmisc}

\usepackage[final]{neurips_2020}   

\usepackage[]{neurips_2020}
\usepackage[utf8]{inputenc} 
\usepackage[T1]{fontenc}    
\usepackage{hyperref}       
\usepackage{url}            
\usepackage{booktabs}       
\usepackage{amsfonts}       
\usepackage{nicefrac}       
\usepackage{microtype}      
\usepackage{diagbox}
\usepackage{microtype}
\usepackage{graphicx}
\usepackage{booktabs} 
\usepackage{lipsum}
\usepackage{graphicx}
\usepackage{multirow}
\usepackage{amsmath}
\usepackage{float}
\usepackage{adjustbox}
\usepackage{hyperref}
\usepackage{xcolor}
\usepackage[ruled,vlined]{algorithm2e}
\usepackage{placeins}
\SetKwInput{KwInput}{Given}             
\SetKwInput{KwOutput}{Output}     

\title{Benchmarking Deep Learning Interpretability in Time Series Predictions}

\author{%
  Aya Abdelsalam Ismail,  \space Mohamed Gunady, \space
  \bf H\'{e}ctor Corrada Bravo\thanks{Authors contributed equally}\space  \space , \space  \bf Soheil Feizi \footnotemark[1]\space \\
    \texttt{\{asalam,mgunady,sfeizi\}@cs.umd.edu, \space hcorrada@umiacs.umd.edu} \\
 Department of Computer Science, University of Maryland\\

}

\begin{document}
\definecolor{ao}{rgb}{0.0, 0.5, 0.0}
\maketitle

\begin{abstract}

Saliency methods are used extensively to highlight the importance of input features in model predictions. These methods are mostly used in vision and language tasks, and their applications to time series data is relatively unexplored. In this paper, we set out to extensively compare the performance of various saliency-based interpretability methods across diverse neural architectures, including Recurrent Neural Network, Temporal Convolutional Networks, and Transformers in a new benchmark \footnote{Code: \url{https://github.com/ayaabdelsalam91/TS-Interpretability-Benchmark}} of synthetic time series data. We propose and report multiple metrics to empirically evaluate the performance of saliency methods for detecting feature importance over time using both precision (i.e., whether identified features contain meaningful signals) and recall (i.e., the number of features with signal identified as important). Through several experiments, we show that (i) in general, network architectures and saliency methods fail to reliably and accurately identify feature importance over time in time series data, (ii) this failure is mainly due to the conflation of time and feature domains, and (iii) the quality of saliency maps can be improved substantially by using our proposed two-step temporal saliency rescaling (TSR) approach that first calculates the importance of each time step before calculating the importance of each feature at a time step.
\end{abstract}

 \section{Introduction}

 As the use of Machine Learning models increases in various domains \cite{rich2016machine,obermeyer2016predicting}, the need for reliable model explanations is crucial \cite{caruana2015intelligible,lipton2018mythos}. This need has resulted in the development of numerous interpretability methods that estimate feature importance \cite{baehrens2010explain,LRP,Simonyan2013DeepIC,kindermans2016investigating,sundararajan2017axiomatic,smilkov2017smoothgrad,shrikumar2017learning,lundberg2017unified,levine2019certifiably}. As opposed to the task of understanding the prediction performance of a model, measuring and understanding the performance of interpretability methods is challenging \cite{hooker2019benchmark,ghorbani2019interpretation,adebayo2018sanity,kindermans2019reliability,singla2019understanding} since there is no ground truth to use for such comparisons. For instance, while one could identify sets of informative features for a specific task a priori, models may not necessarily have to draw information from these features to make accurate predictions.  In multivariate time series data, these challenges are even more profound since we cannot rely on human perception as one would when visualizing interpretations by overlaying saliency maps over images or when highlighting relevant words in a sentence.

\begin{figure*}[hbt!]
\centering
   \includegraphics[width=0.8\textwidth]{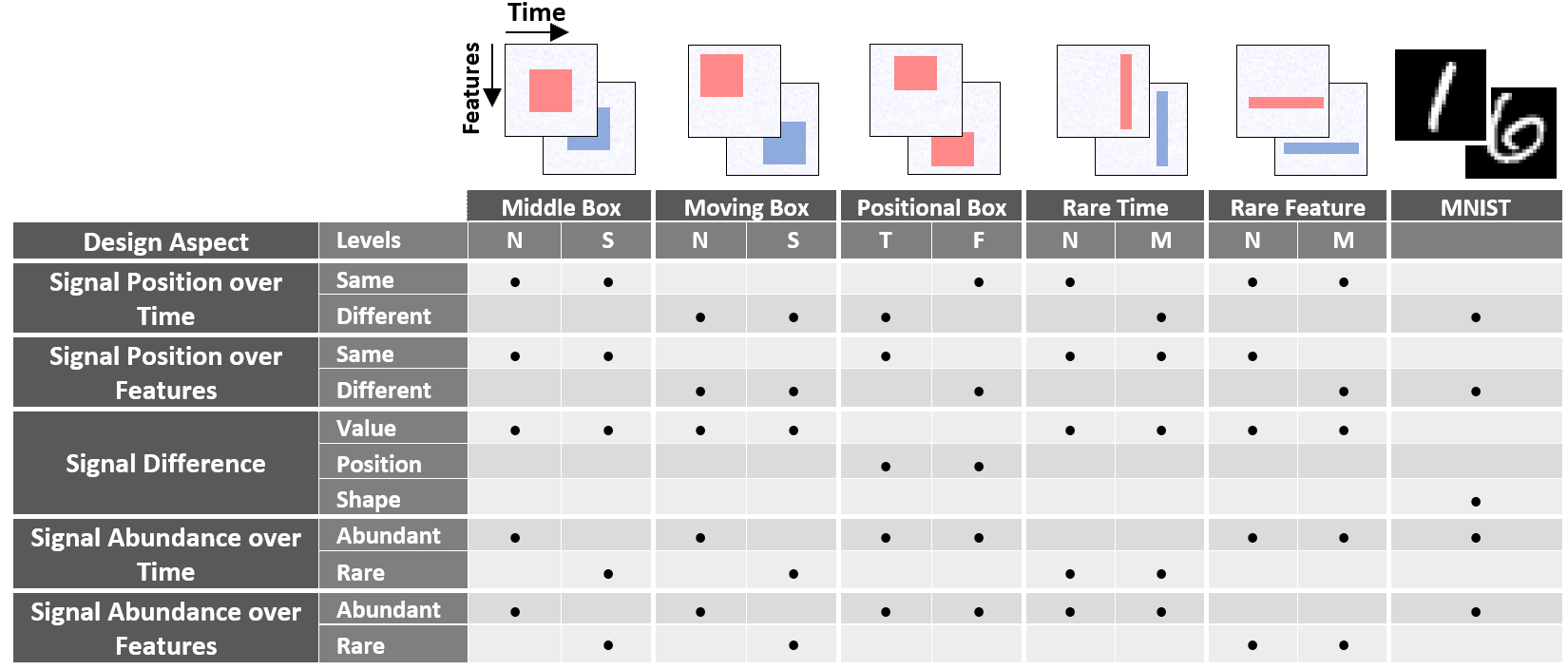}
\caption{Different evaluation datasets used for benchmarking saliency methods.
Some datasets have multiple variations shown as sub-levels. N/S: normal and small shapes,
 T/F: temporal and feature positions, M: moving shape. All datasets are trained for binary classification, except MNIST. Examples are shown above each dataset, where dark red/blue shapes represent informative features.}
\label{fig:DS_DIST}
\end{figure*}

 In this work, we compare the performance of different interpretability methods both perturbation-based and gradient-based methods, across diverse neural architectures including Recurrent Neural Network, Temporal Convolutional Networks, and Transformers when applied to the classification of multivariate time series. We quantify the performance of every (architectures, estimator) pair  for time series data in a systematic way.
We design and generate multiple synthetic datasets to capture different temporal-spatial aspects (e.g., Figure \ref{fig:DS_DIST}). Saliency methods must be able to distinguish important and non-important features at a given time, and capture changes in the importance of features over time. The positions of informative features in our synthetic datasets are known a priori (colored boxes in Figure \ref{fig:DS_DIST}); however, the model might not need \textit{all} informative features to make a prediction. To identify features \textit{needed} by the model, we progressively mask the features identified as important by each interpretability method and measure the accuracy degradation of the trained model.
We then calculate the precision and recall for (architectures, estimator) pairs
at different masks by comparing them to the known set of informative features.

Based on our extensive experiments, we report the following observations: (i) feature importance estimators that produce high-quality saliency maps in images often fail to provide similar high-quality interpretation in time series data, (ii) saliency methods tend to fail to distinguish important vs. non-important features in a given time step; if a feature in a given time is assigned to high saliency, then almost all other features in that time step tend to have high saliency regardless of their actual values, (iii) model architectures have significant effects on the quality of saliency maps. 

After the aforementioned analysis and to improve the quality of saliency methods in time series data, we propose a two-step {\bf T}emporal {\bf S}aliency {\bf R}escaling ({\bf TSR}) approach that can be used on top of any existing saliency method adapting it to time series data. Briefly, the approach works as follows: (a) we first calculate the {\it time-relevance score} for each time by computing the total change in saliency values if that time step is masked; then (b) in each time-step whose time-relevance score is above a certain threshold, we calculate the {\it feature-relevance score} for each feature by computing the total change in saliency values if that feature is masked.
The final (time, feature) importance score is the product of associated time and feature relevance scores. This approach substantially improves the quality of saliency maps produced by various methods when applied to time series data. Figure \ref{fig:saliencyMaps} shows the initial performance of multiple methods, while Figure \ref{fig:NewsaliencyMaps} shows their performance coupled with our proposed TSR method.

\begin{figure*}[hbt!]
\centering
   \includegraphics[width=0.8\textwidth]{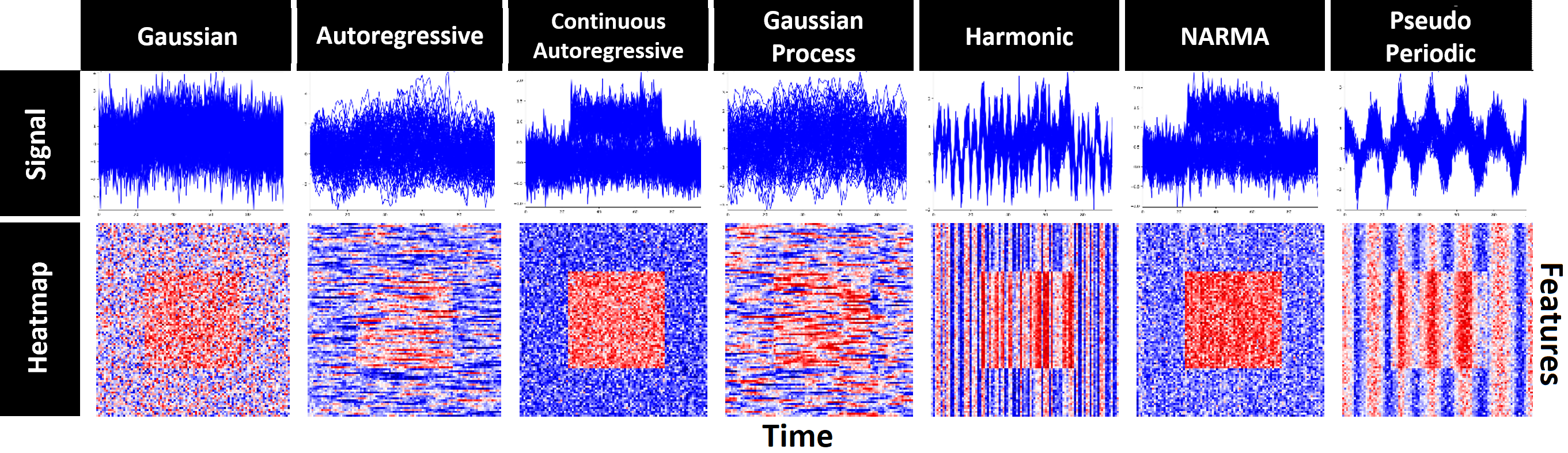}
\caption{Middle box dataset generated by different time series processes.
The first row shows how each feature changes over time when independently sampled from time series processes. The bottom row corresponds to the heatmap of each sample where red represents informative features.}
\label{fig:generation_methods}
\end{figure*}

 \section{Background and Related Work}

The interest in interpretability resulted in several diverse lines of research, all with a common goal of understanding how a network makes a prediction. \cite{ba2014deep, frosst2017distilling, ross2017right, wu2018beyond,inputCellAttention} focus on making neural models more interpretable. \cite{zeiler2014visualizing, sundararajan2017axiomatic, shrikumar2017learning, LRP, Simonyan2013DeepIC, ancona2017towards} estimate the importance of an input feature for a specified output. \citet{kim2017interpretability}  provides an interpretation in terms of human concepts. One key question is whether or not interpretability methods are reliable. \citet{kindermans2019reliability} shows that the explanation can be manipulated by transformations that do not affect the decision-making process.
\citet{ghorbani2019interpretation} introduces an adversarial attack that changes the interpretation without changing the prediction. \citet{adebayo2018sanity} measures changes in the attribute when randomizing model parameters or labels.

Similar to our line of work, modification-based evaluation methods \cite{samek2016evaluating, petsiuk2018rise, kindermans2017learning} involves: applying saliency method, ranking features according to the saliency values, recursively eliminating higher ranked features and measure degradation to the trained model accuracy. \citet{hooker2019benchmark} proposes retraining the model after feature elimination.

Recent work \cite{inputCellAttention,tonekaboni2020went,hardt2020explaining} have identified some limitations in time series interpretability. We provide the first benchmark that systematically evaluates different saliency methods across multiple neural architectures in a multivariate time series setting, identifies common limitations, and proposes a solution to adapt existing methods to time series.

 \subsection{Saliency Methods} \label{sec:Saliency_Methods}
 We compare popular backpropagation-based  and  perturbation based post-hoc saliency methods; each method provides feature importance, or relevance, at a given time step to each input feature. All methods are compared with \textbf{random assignment} as a baseline control.

 In this benchmark, the following saliency methods\footnote[2]{
 \href{https://captum.ai/}{Captum} implementation of different methods was used.} are included: 
 \begin{itemize}
     \item \textbf{Gradient-based:}
     \textit{\textbf{Gradient (GRAD)}}   \cite{baehrens2010explain} the gradient of the output with respect to the input. \textit{\textbf{Integrated Gradients (IG)}} \cite{sundararajan2017axiomatic} the average
gradient while input changes from a non-informative reference point. \textit{\textbf{SmoothGrad (SG)}} \cite{smilkov2017smoothgrad}
 the gradient is computed $n$ times, adding noise to the input  each time. \textit{\textbf{DeepLIFT (DL)}} \cite{shrikumar2017learning} defines a reference point, relevance is the  difference between the activation of each neuron to its reference activation.
\textit{\textbf{Gradient SHAP (GS)}} \cite{lundberg2017unified} adds noise to each input, selects a point along the path between a reference point and input, and computes the gradient of outputs with respect to those points. \textit{\textbf{Deep SHAP (DeepLIFT + Shapley values) (DLS)}} \cite{lundberg2017unified}  takes a distribution of baselines computes the attribution for each input-baseline pair and averages the resulting attributions per input.

     \item \textbf{Perturbation-based:}
     \textit{\textbf{Feature Occlusion (FO)}} \cite{zeiler2014visualizing}  computes attribution as the difference in output after replacing each contiguous region with a given baseline. For time series we considered continuous regions as features with in same time step or multiple time steps grouped together.
     \textit{\textbf{Feature Ablation (FA)}} \cite{suresh2017clinical}
computes attribution as the difference in output after replacing each feature with a baseline. Input features can also be grouped and ablated together rather than individually.
\textit{\textbf{Feature permutation (FP)}} \cite{molnar2020interpretable}
randomly permutes the feature value individually, within a batch and computes the change in output as a result of this modification.
 
 \item \textbf{Other:}  \textit{\textbf{Shapley Value Sampling (SVS)}} \cite{castro2009polynomial} an approximation of Shapley values
       that  involves sampling some random permutations  of the input features and average the marginal contribution of features based the differences on these permutations.
      
 \end{itemize}

 \subsection{Neural Net Architectures}
In this benchmark, we consider 3 main neural architectures groups; Recurrent networks, Convolution neural networks (CNN) and Transformer. For each group we investigate a subset of models that are commonly used for time series data.  Recurrent models include: \textbf{LSTM} \cite{hochreiter1997long} and \textbf{LSTM with Input-Cell Attention} \cite{inputCellAttention}  a variant of LSTM with that attends to inputs from different time steps. For CNN,  
\textbf{Temporal Convolutional Network (TCN)}  \cite{oord2016wavenet, TCN2017, bai2018empirical}  a CNN that handles long sequence time series.  Finally,  we consider the original 
\textbf{Transformers} \cite{vaswani2017attention} implementation.

\section{Problem Definition}

We study a time series classification problem where all time steps 
contribute to making the final output; labels are available after the last time step. In this setting, a network takes multivariate time series input  $X= \lbrack x_1,\dots,x_T \rbrack \in \mathbb{R}^{N\times T}$, where $T$ is the number of time steps and $N$ is the number of features. Let $x_{i,t}$ be the input feature $i$ at time $t$. Similarly, let $X_{:,t}\in \mathbb{R}^N$ and $X_{i,:}\in \mathbb{R}^T$ be the feature vector at time $t$, and the time vector for feature $i$, respectively. The network produces an output  $S(X) = \lbrack S_1(X), ..., S_C (X) \rbrack $, where $C$ is the total number of classes (i.e. outputs). Given a target class $c$, the saliency method finds the relevance $R(X)\in \mathbb{R}^{N\times T}$ which assigns relevance scores $R_{i,t}(X)$ for input feature $i$ at time $t$.

\section{Benchmark Design and Evaluation Metrics} \label{sec:Benchmark}

\subsection{Dataset Design}
Since evaluating interpretability through saliency maps in multivariate time series datasets is nontrivial, we design multiple synthetic datasets where we can control and examine different design aspects that emerge in typical time series datasets. We extend the synthetic data proposed by \citet{inputCellAttention} for binary classification. 
We consider how the discriminating signal is distributed over both time and feature axes, reflecting the importance of time and feature dimensions separately. We also examine how the signal is distributed between classes: difference in value, position, or shape. Additionally, we modify the classification difficulty by decreasing the number of informative features (reducing feature redundancy), i.e., \textit{small box datasets}. Along with synthetic datasets, we included MNIST as a multivariate time series as a more general case (treating one of the image axes as time). Different dataset combinations are shown in Figure \ref{fig:DS_DIST}.



Each synthetic dataset is generated by seven different processes as shown in Figure \ref{fig:generation_methods}, giving a total of 70 datasets. Each feature is independently sampled from either: (a) Gaussian with zero mean and unit variance. (b) Independent sequences of a standard autoregressive time series with Gaussian noise. (c) A standard continuous autoregressive time series with Gaussian noise. (d) Sampled according to a Gaussian Process mixture model. (e) Nonuniformly sampled from a harmonic function. (f) Sequences of standard non–linear autoregressive moving average (NARMA) time series with  Gaussian noise. (g) Nonuniformly sampled from a pseudo period function with  Gaussian noise.
Informative features are then highlighted by the addition of a constant $\mu$
to positive class and subtraction of $\mu$ from negative class (unless specified, $\mu=1$); the embedding size for each sample is $N=50$, and the number of time steps is $T =50$. Figures throughout the paper show data generated as Gaussian noise unless otherwise specified. Further details are provided in the supplementary material.


\subsection{Feature Importance Identification}

\begin{figure*}[hbtp!]
\centering
   \includegraphics[width=0.9\textwidth]{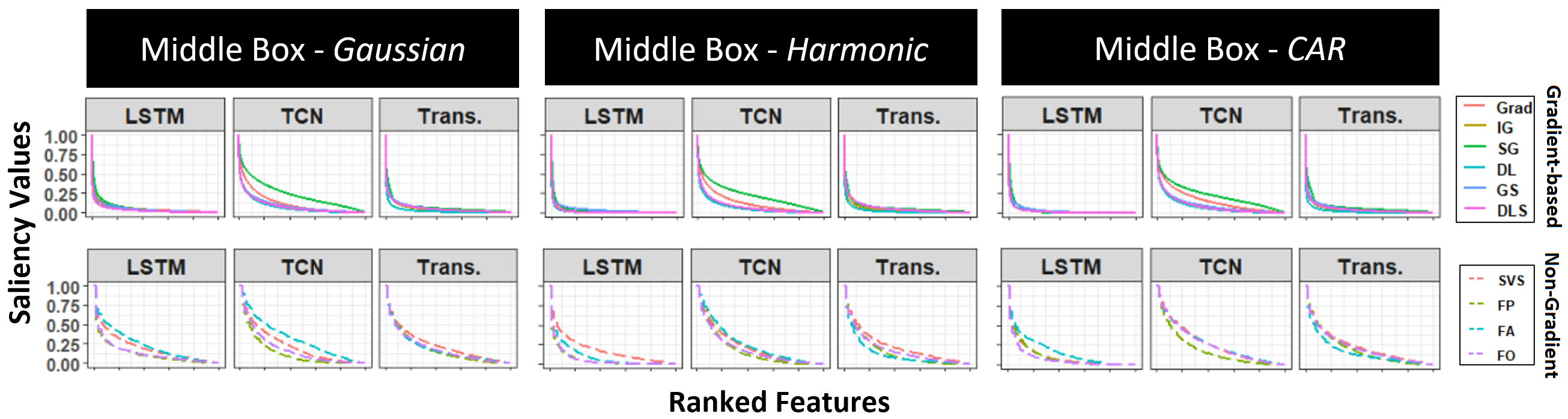}
\caption{The saliency distribution of ranked features produced by different saliency methods for three variations of the Middle Box dataset (Gaussian, Harmonic, Continous Autoregressive (CAR)). Top row shows gradient-based saliency methods while bottom row shows the rest.}
\label{fig:Sal_dist}
\end{figure*}

 Modification-based evaluation metrics \cite{samek2016evaluating, petsiuk2018rise, kindermans2017learning}  have two main issues. First, they assume that feature ranking based on saliency faithfully represents feature importance. Consider the saliency distributions shown in Figure \ref{fig:Sal_dist}. Saliency decays exponentially with feature ranking, meaning that features that are closely ranked might have substantially different saliency values. A second issue, as discussed by \citet{hooker2019benchmark}, is that eliminating features changes the test data distribution violating the assumption that both training and testing data are independent and identically distributed (i.i.d.).  Hence, model accuracy degradation may be a result of changing data distribution rather than removing salient features. In our synthetic dataset benchmark, we address these two issues by the following:
\begin{itemize}
    \item  Sort relevance $R(X)$, so that $R_{e}\left(x_{i,t}\right)$ is the $e^{th}$ element in ordered set $\{R_{e}\left(x_{i,t}\right) \}_{e=1}^{T\times N}$.
 
   \item  Find top $k$ relevant features in the order set such that $\frac{\sum_{e=1}^{k}R_{e}\left(x_{i,t}\right)} {\sum_{i=1,t=1}^{N,T} R(x_{i,t})} \approx d$ (where $d$ is a pre-determined percentage).

   \item  Replace $x_{i,t}$, where $R\left(x_{i,t}\right) \in \{R_{e}\left(x_{i,t}\right) \}_{e=1}^{k}$ with the original distribution (known since this is a synthetic dataset).
   
   \item  Calculate the drop in model accuracy after the masking, this is repeated at different values of $d=\lbrack 0,10,\dots,100\rbrack$.
 
\end{itemize}

We address the first issue by removing features that represent a certain percentage of the overall saliency rather than removing a constant number of features. Since we are using synthetic data and masking using the original data distribution, we are not violating i.i.d. assumptions.
\subsection{Performance Evaluation Metrics}

Masking salient features can result in 
(a) a steep drop in accuracy, meaning that the removed feature is \textit{necessary} for a correct prediction or
(b) unchanged accuracy. The latter may result from 
the saliency method incorrectly identifying the feature as important, or that the removal of that feature is not \textit{sufficient} for the model to behave incorrectly. Some neural architectures tend to use more feature information when making a prediction (i.e., have more recall in terms of importance); this may be the desired behavior in many time series applications where importance changes over time, and the goal of using an interpretability measure is to \textit{detect} all relevant features across time. On the other hand, in some situations, where sparse explanations are preferred, then this behavior may not be appropriate. This in mind, one should not compare saliency methods solely on the loss of accuracy after masking. Instead, we should look into features identified as salient and answer the following questions: (1) \textit{\textbf{Are all features identified as salient informative?}} \textit{(precision)} (2) \textit{\textbf{Was the saliency method able to identify all informative features?}} \textit{(recall)}

We choice to report the \textit{weighted} precision and recall of each \textit{(neural architecture, saliency method)} pair, since,  the saliency value varies dramatically across features Figure \ref{fig:Sal_dist} (detailed calculations are available in the supplementary material).

Through our experiments, we report area under the precision curve (AUP), the area under the recall curve (AUR), and area under precision and recall (AUPR). The curves are calculated by the precision/recall values at different levels of degradation. We also consider feature/time precision and recall (a feature is considered informative if it has information at any time step and vice versa). For the random baseline, we stochastically select a saliency method then permute the saliency values producing arbitrary ranking.

\section{Saliency Methods Fail in Time Series Data}\label{sec:Sal_method_fail}
Due to space limitations, only a subset of the results is reported below; the full set is available in the supplementary material. The results reported in the following section are for models that produce accuracy above 95\% in the classification task.

\subsection{Saliency Map Quality}
\begin{figure*}[htb!]
\centering
\includegraphics[width=0.8\textwidth]{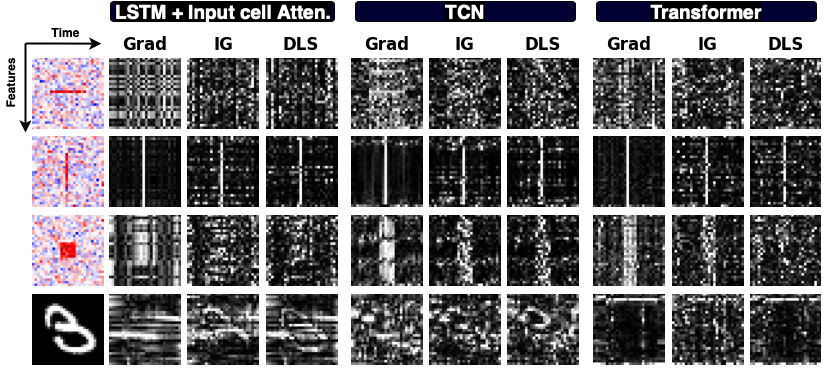}
\caption{Saliency maps produced by Grad, Integrated Gradients, and DeepSHAP for 3 different models on synthetic data and time series MNIST (white represents high saliency). Saliency seems to highlight the correct time step in some cases but fails to identify informative features in a given time.}
\label{fig:saliencyMaps}
\end{figure*}

Consider synthetic examples in Figure \ref{fig:saliencyMaps}; given that the model was able to classify all the samples correctly, one would expect a saliency method to highlight only informative features. However, we find that for the \textit{Middle Box} and  \textit{Rare Feature} datasets, many different (neural architecture, saliency method) pairs are unable to identify informative features. For  \textit{Rare time}, methods identify the correct time steps but are unable to distinguish informative features within those times. Similarly, methods were not able to provide quality saliency maps produced for the multivariate time series MNIST digit.
Overall most (neural architecture, saliency method) pairs fail to identify importance over time.

\begin{figure}[htb!]
\centering
\includegraphics[width=0.8\textwidth]{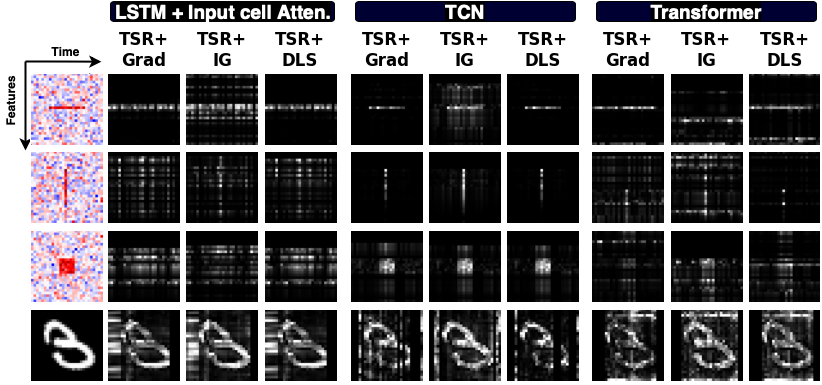}
\caption[Two numerical solutions]{ Saliency maps when applying the proposed Temporal Saliency Rescaling (TSR) approach.}
\label{fig:NewsaliencyMaps} 
\end{figure}


\subsection{Saliency Methods versus Random Ranking}

Here we look into distinctions between each saliency method and a random ranking baseline. The effect of masking salient features on the model accuracy is shown in Figure \ref{fig:MaskAcc}.  In a given panel, the leftmost curve indicates the saliency method that highlights a small number of features that impact accuracy severely (if correct, this method should have high precision); the rightmost curve indicates the saliency method that highlights a large number of features that impact accuracy severely (if correct, this method should show high recall).

\subsubsection*{Model Accuracy Drop}
We were unable to identify a consistent trend for saliency methods across all neural architectures throughout experiments. Instead, saliency methods for a given architecture behave similarly across datasets. E.g., in TCN Grad and SmoothGrad had steepest accuracy drop across all datasets while LSTM showed no clear distinction between random assignment and non-random saliency method curves (this means that LSTM is very hard to interpret regardless of the saliency method used as \cite{inputCellAttention})
Variance in performance between methods can be explained by the dataset itself rather than the methods. E.g., the \textit{Moving box} dataset showed minimal variance across all methods, while \textit{Rare time} dataset showed the highest.

 



\subsubsection*{Precision and Recall}
\begin{figure*}[hbtp!]
\centering
\includegraphics[width=\textwidth]{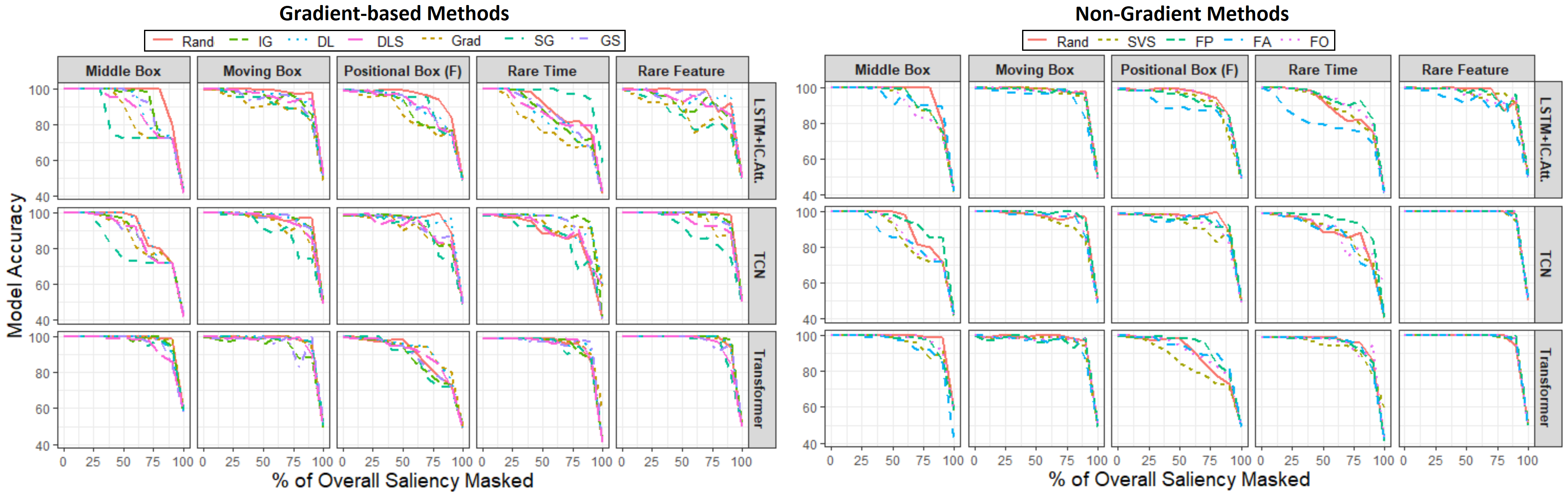}
\caption{The effect of masking features identified as salient by different  methods against a random baseline.
Gradient-based and non-gradient based saliency methods are shown in the left and right plots, respectively. The rate of accuracy drop is not consistent; in many cases there is not much improvement over random baseline.}
\label{fig:MaskAcc}
\end{figure*}

Looking at precision and recall distribution box plots Figure \ref{fig:Precision/Recall} (the precision and recall graphs per dataset are available in the supplementary materials), we observe the following: (a) Model architecture has the largest effect on precision and recall. (b)  Results do not show clear distinctions between saliency methods. (c) Methods can identify informative time steps while fail to identify informative features; AUPR in the time domain (second-row Figure \ref{fig:Precision/Recall}) is  higher than that in the feature domain  (third-row Figure \ref{fig:Precision/Recall}).
(d) Methods identify most features in an informative time step as salient, AUR in feature domain is very high while having very low AUP. This is consistent with what we see in Figure \ref{fig:saliencyMaps}, where all features in informative time steps are highlighted regardless of there actual values. (e) Looking at AUP, AUR, and AUPR values, we find that the steepness in accuracy drop depends on the dataset. A steep drop in model accuracy does not indicate that a saliency method is correctly identifying features used by the model since, in most cases, saliency methods with leftmost curves in Figure \ref{fig:MaskAcc} have the lowest precision and recall values.

    

    

\begin{figure}[htbp!]
\centering
\begin{tabular}{cc}
\includegraphics[width=.47\textwidth]{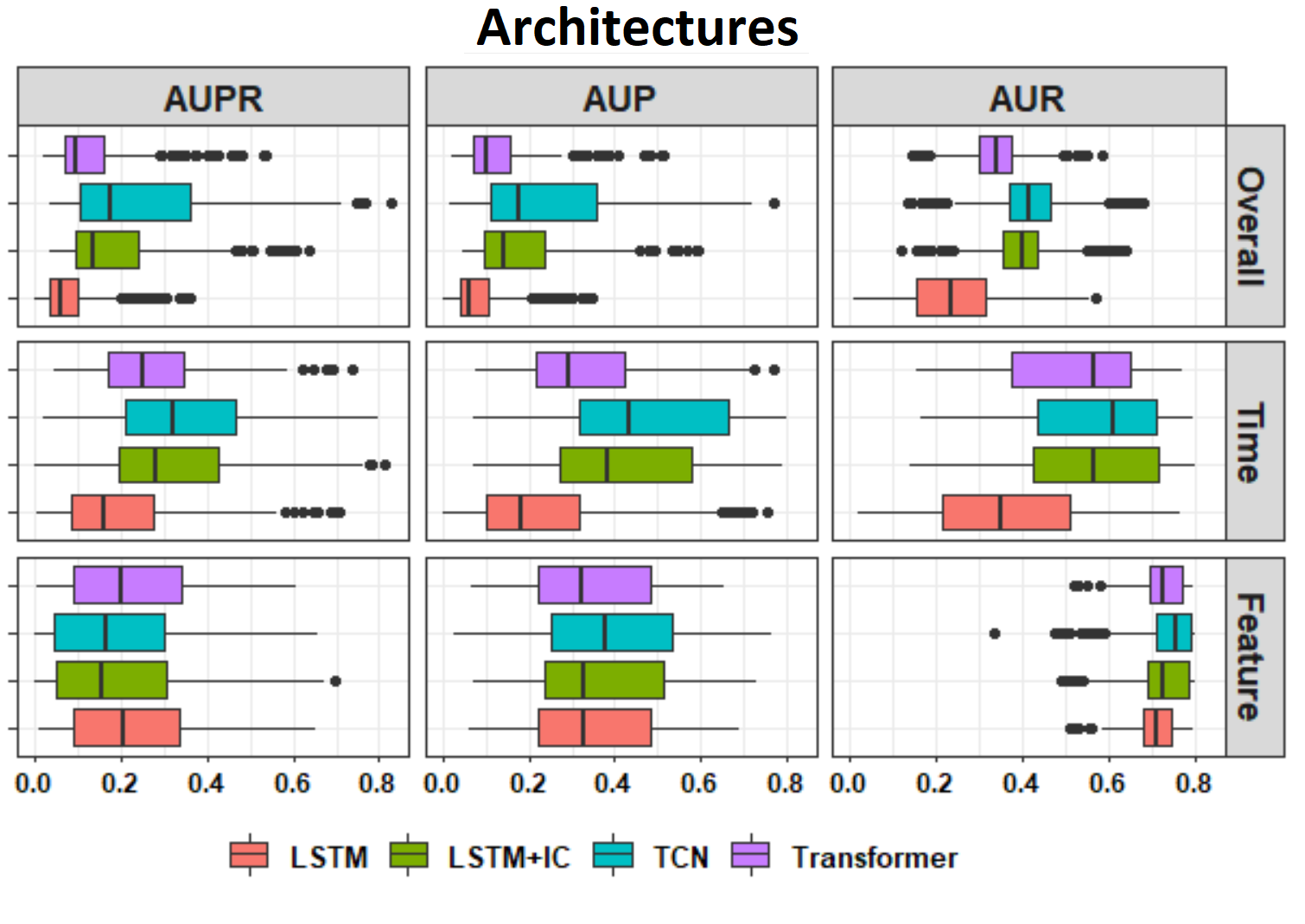}&
\includegraphics[width=.47\textwidth]{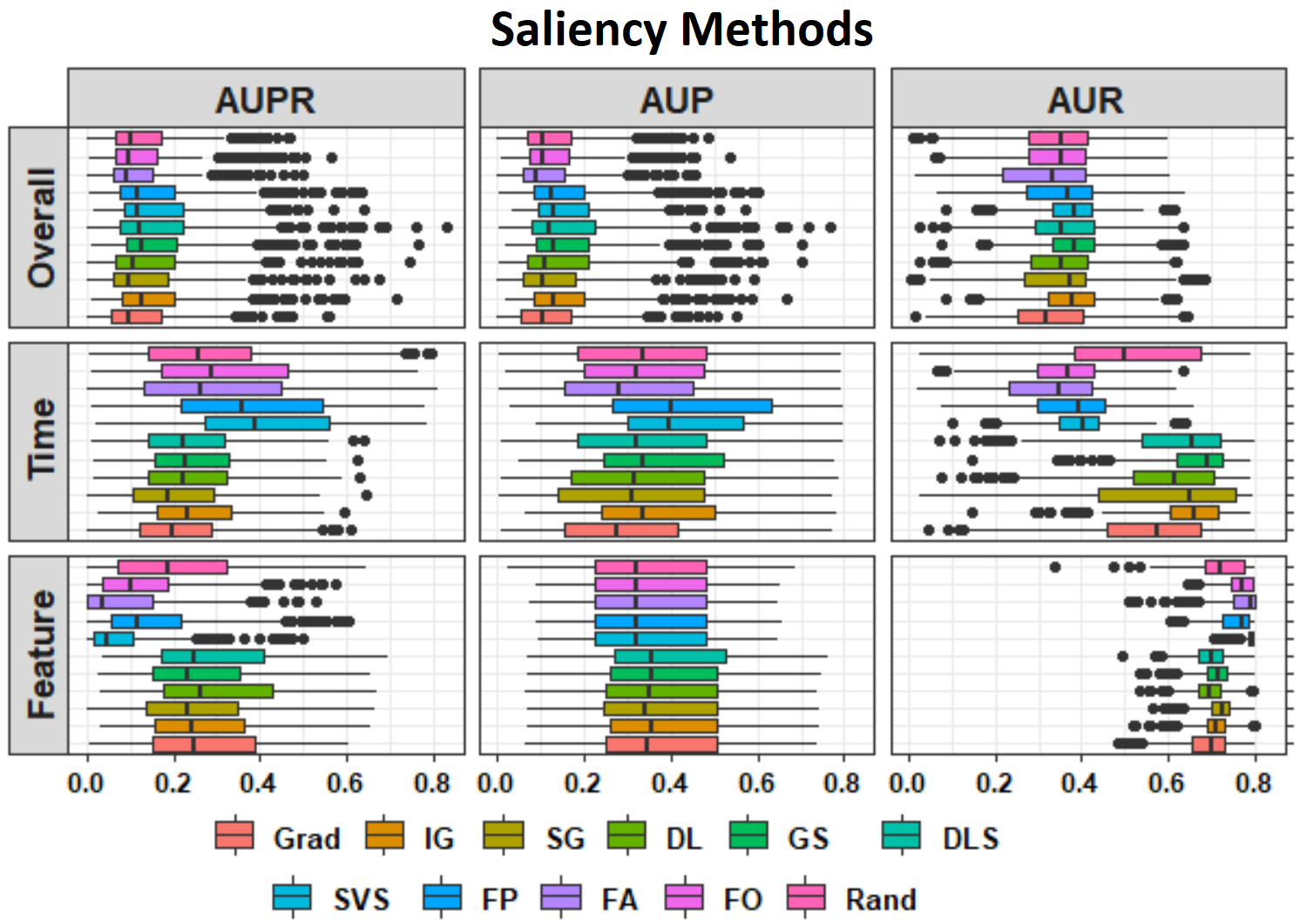}\\
\textbf{(a) }   & \textbf{(b)} 
\end{tabular}
\caption{Precision and Recall distribution box plots, the top row represents overall Precision/Recall, while the second two rows show  Precision/Recall distribution on time and feature axes (a) Distribution across architectures. (b) Distribution across saliency methods.
}
\label{fig:Precision/Recall}
\end{figure}

\section{Saliency Maps for Images versus Multivariate Time Series}\label{sec:imagesVerusTS}
\begin{figure}[htb!]
\centering
\includegraphics[width=1\textwidth]{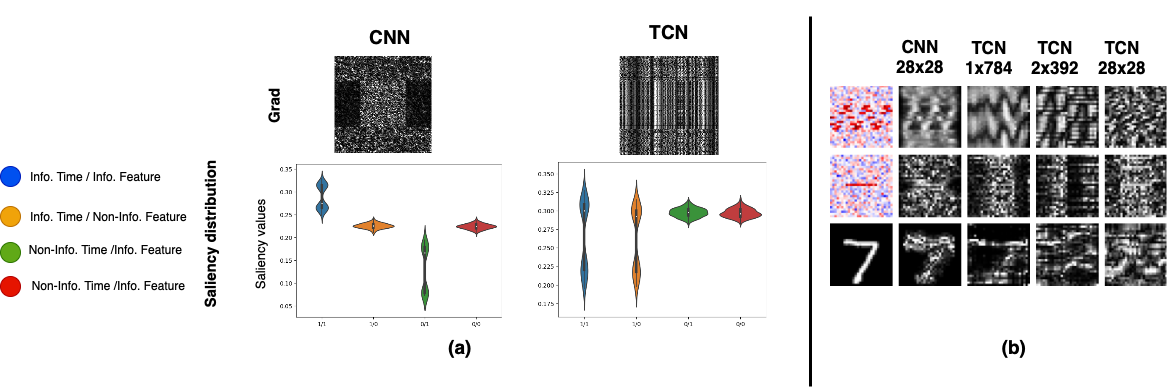}
\caption{\textbf{(a)} Saliency maps and distribution produced by CNN versus TCN for \textit{Middle Box}.  \textbf{(b)} Saliency Maps for samples treated as image (CNN) vs. uni-, bi- or multi-variate time series (TCN).}
\label{fig:multivars_sal}
\end{figure}

Since saliency methods are commonly evaluated on images, we compare the saliency maps produced from models like CNN, which fit images, to the maps produced by temporal models like TCN, over our evaluation datasets by treating the complete multivariate time series as an image.  Figure \ref{fig:multivars_sal}(a) shows two examples of such saliency maps. The maps produced by CNN can distinguish informative pixels corresponding to informative features in informative time steps. However, maps produced from TCN fall short in distinguishing important features within a given time step. Looking at the saliency distribution of gradients for each model, stratified by the category of each pixel with respect to its importance in both time and feature axes; we find that CNN correctly assigns higher saliency values to pixels with information in both feature and time axes compared to the other categories, which is not the case with TCN, that is biased in the time direction.
That observation supports the conclusion that even though most saliency methods we examine work for images, they generally fail for multivariate time series. It should be noted that this conclusion should not be misinterpreted as a suggestion to treat time series as images (in many cases this is not possible due to the decrease in model performance and increase in dimensionality).

Finally, we examine the effect of reshaping a multivariate time series into univariate or bivariate time series. Figure \ref{fig:multivars_sal} (b) shows a few examples of saliency maps produced by the various treatment approaches of the same sample (images for CNN, uni, bi, multivariate time series for TCN). One can see that CNN and univariate TCN produce interpretable maps, while the maps for the bivariate and multivariate TCN are harder to interpret. That is due to the failure of these methods to distinguish informative features within informative time steps, but rather focusing more on highlighting informative time steps. 

These observations suggest that saliency maps fail when feature and time domains are conflated. When the input is represented solely on the feature domain (as is the case of CNN), saliency maps are relatively accurate. When the input is represented solely on the time domain, maps are also accurate. However, when feature and time domains are both present, the saliency maps across these domains are conflated, leading to poor behavior. This observation motivates our proposed method to adapt existing saliency methods to multivariate time series data.


\section{Temporal Saliency Rescaling}
From the results presented in previous sections, we conclude that most saliency methods identify informative time steps successfully while they fail in identifying feature importance in those time steps. In this section, we propose a method that can be used on top of any generic interpretation method to boost its performance in time series applications. The key idea is to decouple the (time,feature) importance scores to time and feature relevance scores using a two-step procedure called {\bf T}emporal {\bf S}aliency {\bf R}escaling ({\bf TSR}). In the first step, we calculate the {\it time-relevance score} for each time by computing the total change in saliency values if that time step is masked. Based on our experiments presented in the last sections, many existing interpretation methods would provide reliable time-relevance scores. In the second step, in each time-step whose time-relevance score is above a certain threshold $\alpha$, we compute the {\it feature-relevance score} for each feature by computing the total change in saliency values if that feature is masked. By choosing a proper value for $\alpha$, the second step can be performed in a few highly-relevant time steps to reduce the overall computational complexity of the method. Then, the final (time, feature) importance score is the product of associated time and feature relevance scores. The method is formally presented in Algorithm \ref{algo:TSR}.

\begin{algorithm}[htpb!]
  \KwInput{ input $X$, a baseline interpretation method $R(.)$}
  \KwOutput{TSR interpretation method $R^{TSR}(.)$}

 \For{$t\gets0$ \KwTo $T$ }{
    Mask all features at time $t$: $\overbar X_{:,t}=0$, otherwise $\overbar X=X$\;
    Compute Time-Relevance Score $\Delta_t^{time} = \sum_{i,t} |R_{i,t}(X) -R_{i,t}(\overbar X)|$\;
    }
    
     \For{$t\gets0$ \KwTo $T$ }{

        \For{$i\gets0$ \KwTo $N$}{
         \eIf{$\Delta_t^{time}> \alpha$}{
        Mask feature $i$ at time $t$: $\overbar X_{i,:}=0$, otherwise $\overbar X=X$\;
         Compute Feature-Relevance Score $\Delta_i^{feature} = \sum_{i,t} |R_{i,t}(X) -R_{i,t}(\overbar X)|$\;
        }
        { Feature-Relevance Score
        $\Delta_i^{feature}=0$\;
        }
 Compute (time,feature) importance score $R^{TSR}_{i,t}=\Delta_i^{feature}\times \Delta_t^{time}$ \;
   }
    }
 \caption{Temporal Saliency Rescaling (TSR)}
 \label{algo:TSR}
\end{algorithm}

    

Figure \ref{fig:NewsaliencyMaps} shows updated saliency maps when applying \textbf{TSR} on the same examples in Figures \ref{fig:saliencyMaps}. There is a definite improvement in saliency quality across different architectures and interpretability methods except for SmoothGrad; this is probably because SmoothGrad adds noise to gradients, and using a noisy gradient as a baseline may not be appropriate. Table \ref{tab:Saliency_performance} shows the performance of \textbf{TSR} with simple Gradient compared to some standard saliency method on the benchmark metrics described in Section \ref{sec:Benchmark}.
\textbf{TSR + Grad} outpreforms other methods on all metrics.

The proposed rescaling approach improves the ability of saliency methods to capture feature importance over time but significantly increases the computational cost of producing a saliency map. Other approaches \cite{hooker2019benchmark,smilkov2017smoothgrad} have relied on a similar trade-off between interpretability and computational complexity.
In the supplementary material, we show the effect of applying temporal saliency rescaling on other datasets and provide possible optimizations.

\begin{table}[htb!]
    \centering
  \begin{tabular}{l|rrrr|rrrr}
          & \multicolumn{4}{c|}{Middle Box} & \multicolumn{4}{c}{Moving Box} \\
          \hline
    Saliency Methods & \multicolumn{1}{l}{AUPR}& \multicolumn{1}{l}{AUP} & \multicolumn{1}{l}{AUR}  & \multicolumn{1}{l|}{AUC} & \multicolumn{1}{l}{AUPR} & \multicolumn{1}{l}{AUP} & \multicolumn{1}{l}{AUR} & \multicolumn{1}{l}{AUC} \\
        \hline \hline
    Grad & 0.331 & 0.328 & 0.457  & 64.90 & 0.225& 0.229 & 0.394  & 95.35 \\
    DLS  & 0.344  & 0.344 & 0.452 & 68.30  & 0.288& 0.288 & 0.435 & 94.05 \\
    SG    & 0.294 & 0.300 & 0.451 & 64.00  & 0.241& 0.247 & 0.395 & 92.90 \\

       \hline
     TSR + Grad & \textbf{0.399} & \textbf{0.381} & \textbf{0.471}  & \textbf{62.20} & \textbf{0.335} & \textbf{0.326} & \textbf{0.456} & \textbf{84.00} \\
      \hline                                                                              
    \end{tabular}%
    
        \caption{Results from TCN on Middle Box and Moving Box synthetic datasets. Higher AUPR, AUP, and AUR values indicate better performance. AUC lower values are better as this indicates that the rate of accuracy drop is higher.}
    \label{tab:Saliency_performance}
\end{table}

\section{Summary and Conclusion}
In this work, we have studied deep learning interpretation methods when applied to multivariate time series data on various neural network architectures. To quantify the performance of each (interpretation method, architecture) pair, we have created a comprehensive synthetic benchmark where positions of informative features are known. We measure the quality of the generated interpretation by calculating the degradation of the trained model accuracy when inferred salient features are masked. These feature sets are then used to calculate the precision and recall for each pair.

Interestingly, we have found that commonly-used saliency methods, including both gradient-based, and perturbation-based methods, fail to produce high-quality interpretations when applied to multivariate time series data. However, they can produce accurate maps when multivariate time series are represented as either images or univariate time series. That is, when temporal and feature domains are combined in a multivariate time series, saliency methods break down in general. The exact mathematical mechanism underlying this result is an open question. Consequently, there is no clear distinction in performance between different interpretability methods on multiple evaluation metrics when applied to multivariate time series, and in many cases, the performance is similar to random saliency. Through experiments, we observe that methods generally identify salient time steps but cannot distinguish important vs. non-important features within a given time step. Building on this observation, we then propose a two-step temporal saliency rescaling approach to adapt existing saliency methods to time series data. This approach has led to substantial improvements in the quality of saliency maps produced by different methods.

\section{Broader Impact}
The challenge presented by meaningful interpretation of Deep Neural Networks (DNNs) is a technical barrier preventing their serious adoption by practitioners in fields such as Neuroscience, Medicine, and Finance~\cite{lipton2017doctor,tonekaboni2019clinicians}. 
Accurate DNNs are not, by themselves, sufficient to be used routinely in high stakes applications such as healthcare. For example, in clinical research, one might like to ask, "why did you predict this person as more likely to develop a certain disease?" Our work aims to answer such questions. Many critical applications involve time series data, e.g., electronic health records, functional Magnetic Resonance Imaging (fMRI) data, and market data; nevertheless, the majority of research on interpretability focuses on vision and language tasks. Our work aims to interpret DNNs applied to time series data.

Having interpretable DNNs has many positive outcomes. It will help increase the transparency of these models and ease their applications in a variety of research areas. Understanding how a model makes its decisions can help guide modifications to the model to produce better and fairer results. Critically, failure to provide faithful interpretations is a severe negative outcome. Having no interpretation at all is, in many situations, better than trusting an incorrect interpretation. Therefore, we believe this study can lead to significant positive and broad impacts in different applications.

\section*{Acknowledgements}
We thank Kalinda Vathupola for his thoughtful feedback on this work. This project was supported in part by NSF CAREER AWARD 1942230, a grant from NIST 303457-00001, AWS Machine Learning Research Award and Simons Fellowship on ``Foundations of Deep Learning.''
\bibliography{neurips_2020.bib}{}

\begin{thebibliography}{43}
\providecommand{\natexlab}[1]{#1}
\providecommand{\url}[1]{\texttt{#1}}
\expandafter\ifx\csname urlstyle\endcsname\relax
  \providecommand{\doi}[1]{doi: #1}\else
  \providecommand{\doi}{doi: \begingroup \urlstyle{rm}\Url}\fi

\bibitem[Rich(2016)]{rich2016machine}
Michael~L Rich.
\newblock Machine learning, automated suspicion algorithms, and the fourth
  amendment.
\newblock In \emph{University of Pennsylvania Law Review}, 2016.

\bibitem[Obermeyer and Emanuel(2016)]{obermeyer2016predicting}
Ziad Obermeyer and Ezekiel~J Emanuel.
\newblock Predicting the future—big data, machine learning, and clinical
  medicine.
\newblock In \emph{The New England journal of medicine}, 2016.

\bibitem[Caruana et~al.(2015)Caruana, Lou, Gehrke, Koch, Sturm, and
  Elhadad]{caruana2015intelligible}
Rich Caruana, Yin Lou, Johannes Gehrke, Paul Koch, Marc Sturm, and Noemie
  Elhadad.
\newblock Intelligible models for healthcare: Predicting pneumonia risk and
  hospital 30-day readmission.
\newblock In \emph{International conference on knowledge discovery and data
  mining}, 2015.

\bibitem[Lipton(2018)]{lipton2018mythos}
Zachary~C Lipton.
\newblock The mythos of model interpretability.
\newblock In \emph{Queue}, 2018.

\bibitem[Baehrens et~al.(2010)Baehrens, Schroeter, Harmeling, Kawanabe, Hansen,
  and M{\~A}{\v{z}}ller]{baehrens2010explain}
David Baehrens, Timon Schroeter, Stefan Harmeling, Motoaki Kawanabe, Katja
  Hansen, and Klaus-Robert M{\~A}{\v{z}}ller.
\newblock How to explain individual classification decisions.
\newblock In \emph{Journal of Machine Learning Research}, 2010.

\bibitem[Bach et~al.(2015)Bach, Binder, Montavon, Klauschen, M{\"u}ller, and
  Samek]{LRP}
Sebastian Bach, Alexander Binder, Gr{\'e}goire Montavon, Frederick Klauschen,
  Klaus-Robert M{\"u}ller, and Wojciech Samek.
\newblock On pixel-wise explanations for non-linear classifier decisions by
  layer-wise relevance propagation.
\newblock In \emph{PLoS ONE}, 2015.

\bibitem[Simonyan et~al.(2013)Simonyan, Vedaldi, and
  Zisserman]{Simonyan2013DeepIC}
Karen Simonyan, Andrea Vedaldi, and Andrew Zisserman.
\newblock Deep inside convolutional networks: Visualising image classification
  models and saliency maps.
\newblock \emph{CoRR}, 2013.

\bibitem[Kindermans et~al.(2016)Kindermans, Sch{\"u}tt, M{\"u}ller, and
  D{\"a}hne]{kindermans2016investigating}
Pieter-Jan Kindermans, Kristof Sch{\"u}tt, Klaus-Robert M{\"u}ller, and Sven
  D{\"a}hne.
\newblock Investigating the influence of noise and distractors on the
  interpretation of neural networks.
\newblock \emph{arXiv preprint arXiv:1611.07270}, 2016.

\bibitem[Sundararajan et~al.(2017)Sundararajan, Taly, and
  Yan]{sundararajan2017axiomatic}
Mukund Sundararajan, Ankur Taly, and Qiqi Yan.
\newblock Axiomatic attribution for deep networks.
\newblock In \emph{International Conference on Machine Learning}, 2017.

\bibitem[Smilkov et~al.(2017)Smilkov, Thorat, Kim, Vi{\'e}gas, and
  Wattenberg]{smilkov2017smoothgrad}
Daniel Smilkov, Nikhil Thorat, Been Kim, Fernanda Vi{\'e}gas, and Martin
  Wattenberg.
\newblock Smoothgrad: removing noise by adding noise.
\newblock \emph{arXiv preprint arXiv:1706.03825}, 2017.

\bibitem[Shrikumar et~al.(2017)Shrikumar, Greenside, and
  Kundaje]{shrikumar2017learning}
Avanti Shrikumar, Peyton Greenside, and Anshul Kundaje.
\newblock Learning important features through propagating activation
  differences.
\newblock In \emph{International Conference on Machine Learning}, 2017.

\bibitem[Lundberg and Lee(2017)]{lundberg2017unified}
Scott~M Lundberg and Su-In Lee.
\newblock A unified approach to interpreting model predictions.
\newblock In \emph{Advances in Neural Information Processing Systems}, 2017.

\bibitem[Levine et~al.(2019)Levine, Singla, and Feizi]{levine2019certifiably}
Alexander Levine, Sahil Singla, and Soheil Feizi.
\newblock Certifiably robust interpretation in deep learning.
\newblock \emph{arXiv preprint arXiv:1905.12105}, 2019.

\bibitem[Hooker et~al.(2019)Hooker, Erhan, Kindermans, and
  Kim]{hooker2019benchmark}
Sara Hooker, Dumitru Erhan, Pieter-Jan Kindermans, and Been Kim.
\newblock A benchmark for interpretability methods in deep neural networks.
\newblock In \emph{Advances in Neural Information Processing Systems}, 2019.

\bibitem[Ghorbani et~al.(2019)Ghorbani, Abid, and
  Zou]{ghorbani2019interpretation}
Amirata Ghorbani, Abubakar Abid, and James Zou.
\newblock Interpretation of neural networks is fragile.
\newblock In \emph{AAAI Conference on Artificial Intelligence}, 2019.

\bibitem[Adebayo et~al.(2018)Adebayo, Gilmer, Muelly, Goodfellow, Hardt, and
  Kim]{adebayo2018sanity}
Julius Adebayo, Justin Gilmer, Michael Muelly, Ian Goodfellow, Moritz Hardt,
  and Been Kim.
\newblock Sanity checks for saliency maps.
\newblock In \emph{Advances in Neural Information Processing Systems}, 2018.

\bibitem[Kindermans et~al.(2019)Kindermans, Hooker, Adebayo, Alber, Sch{\"u}tt,
  D{\"a}hne, Erhan, and Kim]{kindermans2019reliability}
Pieter-Jan Kindermans, Sara Hooker, Julius Adebayo, Maximilian Alber, Kristof~T
  Sch{\"u}tt, Sven D{\"a}hne, Dumitru Erhan, and Been Kim.
\newblock The (un) reliability of saliency methods.
\newblock In \emph{Explainable AI: Interpreting, Explaining and Visualizing
  Deep Learning}, 2019.

\bibitem[Singla et~al.(2019)Singla, Wallace, Feng, and
  Feizi]{singla2019understanding}
Sahil Singla, Eric Wallace, Shi Feng, and Soheil Feizi.
\newblock Understanding impacts of high-order loss approximations and features
  in deep learning interpretation.
\newblock In \emph{International Conference on Machine Learning}, 2019.

\bibitem[Ba and Caruana(2014)]{ba2014deep}
Jimmy Ba and Rich Caruana.
\newblock Do deep nets really need to be deep?
\newblock In \emph{Advances in Neural Information Processing Systems}, 2014.

\bibitem[Frosst and Hinton(2017)]{frosst2017distilling}
Nicholas Frosst and Geoffrey Hinton.
\newblock Distilling a neural network into a soft decision tree.
\newblock \emph{arXiv preprint arXiv:1711.09784}, 2017.

\bibitem[Ross et~al.(2017)Ross, Hughes, and Doshi-Velez]{ross2017right}
Andrew~Slavin Ross, Michael~C Hughes, and Finale Doshi-Velez.
\newblock Right for the right reasons: Training differentiable models by
  constraining their explanations.
\newblock \emph{arXiv preprint arXiv:1703.03717}, 2017.

\bibitem[Wu et~al.(2018)Wu, Hughes, Parbhoo, Zazzi, Roth, and
  Doshi-Velez]{wu2018beyond}
Mike Wu, Michael~C Hughes, Sonali Parbhoo, Maurizio Zazzi, Volker Roth, and
  Finale Doshi-Velez.
\newblock Beyond sparsity: Tree regularization of deep models for
  interpretability.
\newblock In \emph{AAAI Conference on Artificial Intelligence}, 2018.

\bibitem[Ismail et~al.(2019)Ismail, Gunady, Pessoa, Corrada~Bravo, and
  Feizi]{inputCellAttention}
Aya~Abdelsalam Ismail, Mohamed Gunady, Luiz Pessoa, Hector Corrada~Bravo, and
  Soheil Feizi.
\newblock Input-cell attention reduces vanishing saliency of recurrent neural
  networks.
\newblock In \emph{Advances in Neural Information Processing Systems}, 2019.

\bibitem[Zeiler and Fergus(2014)]{zeiler2014visualizing}
Matthew~D Zeiler and Rob Fergus.
\newblock Visualizing and understanding convolutional networks.
\newblock In \emph{European conference on computer vision}, 2014.

\bibitem[Ancona et~al.(2018)Ancona, Ceolini, {\"O}ztireli, and
  Gross]{ancona2017towards}
Marco Ancona, Enea Ceolini, Cengiz {\"O}ztireli, and Markus Gross.
\newblock Towards better understanding of gradient-based attribution methods
  for deep neural networks.
\newblock In \emph{International Conference on Learning Representations}, 2018.

\bibitem[Kim et~al.(2018)Kim, Wattenberg, Gilmer, Cai, Wexler, Viegas, and
  Sayres]{kim2017interpretability}
Been Kim, Martin Wattenberg, Justin Gilmer, Carrie Cai, James Wexler, Fernanda
  Viegas, and Rory Sayres.
\newblock Interpretability beyond feature attribution: Quantitative testing
  with concept activation vectors (tcav).
\newblock In \emph{International Conference on Machine Learning}, 2018.

\bibitem[Samek et~al.(2016)Samek, Binder, Montavon, Lapuschkin, and
  M{\"u}ller]{samek2016evaluating}
Wojciech Samek, Alexander Binder, Gr{\'e}goire Montavon, Sebastian Lapuschkin,
  and Klaus-Robert M{\"u}ller.
\newblock Evaluating the visualization of what a deep neural network has
  learned.
\newblock \emph{IEEE transactions on neural networks and learning systems},
  2016.

\bibitem[Petsiuk et~al.(2018)Petsiuk, Das, and Saenko]{petsiuk2018rise}
Vitali Petsiuk, Abir Das, and Kate Saenko.
\newblock Rise: Randomized input sampling for explanation of black-box models.
\newblock \emph{arXiv preprint arXiv:1806.07421}, 2018.

\bibitem[Kindermans et~al.(2017)Kindermans, Sch{\"u}tt, Alber, M{\"u}ller,
  Erhan, Kim, and D{\"a}hne]{kindermans2017learning}
Pieter-Jan Kindermans, Kristof~T Sch{\"u}tt, Maximilian Alber, Klaus-Robert
  M{\"u}ller, Dumitru Erhan, Been Kim, and Sven D{\"a}hne.
\newblock Learning how to explain neural networks: Patternnet and
  patternattribution.
\newblock \emph{arXiv preprint arXiv:1705.05598}, 2017.

\bibitem[Tonekaboni et~al.(2020)Tonekaboni, Joshi, Duvenaud, and
  Goldenberg]{tonekaboni2020went}
Sana Tonekaboni, Shalmali Joshi, David Duvenaud, and Anna Goldenberg.
\newblock What went wrong and when? instance-wise feature importance for
  time-series models.
\newblock \emph{arXiv preprint arXiv:2003.02821}, 2020.

\bibitem[Hardt et~al.(2020)Hardt, Rajkomar, Flores, Dai, Howell, Corrado, Cui,
  and Hardt]{hardt2020explaining}
Michaela Hardt, Alvin Rajkomar, Gerardo Flores, Andrew Dai, Michael Howell,
  Greg Corrado, Claire Cui, and Moritz Hardt.
\newblock Explaining an increase in predicted risk for clinical alerts.
\newblock In \emph{ACM Conference on Health, Inference, and Learning}, 2020.

\bibitem[Suresh et~al.(2017)Suresh, Hunt, Johnson, Celi, Szolovits, and
  Ghassemi]{suresh2017clinical}
Harini Suresh, Nathan Hunt, Alistair Johnson, Leo~Anthony Celi, Peter
  Szolovits, and Marzyeh Ghassemi.
\newblock Clinical intervention prediction and understanding using deep
  networks.
\newblock \emph{arXiv preprint arXiv:1705.08498}, 2017.

\bibitem[Molnar(2020)]{molnar2020interpretable}
Christoph Molnar.
\newblock \emph{Interpretable Machine Learning}.
\newblock Lulu. com, 2020.

\bibitem[Castro et~al.(2009)Castro, G{\'o}mez, and
  Tejada]{castro2009polynomial}
Javier Castro, Daniel G{\'o}mez, and Juan Tejada.
\newblock Polynomial calculation of the shapley value based on sampling.
\newblock \emph{Computers \& Operations Research}, 36\penalty0 (5):\penalty0
  1726--1730, 2009.

\bibitem[Hochreiter and Schmidhuber(1997)]{hochreiter1997long}
Sepp Hochreiter and J{\"u}rgen Schmidhuber.
\newblock Long short-term memory.
\newblock In \emph{Neural computation}, 1997.

\bibitem[Oord et~al.(2016)Oord, Dieleman, Zen, Simonyan, Vinyals, Graves,
  Kalchbrenner, Senior, and Kavukcuoglu]{oord2016wavenet}
Aaron van~den Oord, Sander Dieleman, Heiga Zen, Karen Simonyan, Oriol Vinyals,
  Alex Graves, Nal Kalchbrenner, Andrew Senior, and Koray Kavukcuoglu.
\newblock Wavenet: A generative model for raw audio.
\newblock \emph{arXiv preprint arXiv:1609.03499}, 2016.

\bibitem[Lea et~al.(2017)Lea, Flynn, Vidal, Reiter, and Hager]{TCN2017}
Colin Lea, Michael Flynn, Rene Vidal, Austin Reiter, and Gregory Hager.
\newblock Temporal convolutional networks for action segmentation and
  detection.
\newblock In \emph{Conference on Computer Vision and Pattern Recognition},
  2017.

\bibitem[Bai et~al.(2018)Bai, Kolter, and Koltun]{bai2018empirical}
Shaojie Bai, J~Zico Kolter, and Vladlen Koltun.
\newblock An empirical evaluation of generic convolutional and recurrent
  networks for sequence modeling.
\newblock \emph{arXiv preprint arXiv:1803.01271}, 2018.

\bibitem[Vaswani et~al.(2017)Vaswani, Shazeer, Parmar, Uszkoreit, Jones, Gomez,
  Kaiser, and Polosukhin]{vaswani2017attention}
Ashish Vaswani, Noam Shazeer, Niki Parmar, Jakob Uszkoreit, Llion Jones,
  Aidan~N Gomez, {\L}ukasz Kaiser, and Illia Polosukhin.
\newblock Attention is all you need.
\newblock In \emph{Advances in Neural Information Processing Systems}, 2017.

\bibitem[Lipton(2017)]{lipton2017doctor}
Zachary~C Lipton.
\newblock The doctor just won't accept that!
\newblock \emph{arXiv preprint arXiv:1711.08037}, 2017.

\bibitem[Tonekaboni et~al.(2019)Tonekaboni, Joshi, McCradden, and
  Goldenberg]{tonekaboni2019clinicians}
Sana Tonekaboni, Shalmali Joshi, Melissa~D McCradden, and Anna Goldenberg.
\newblock What clinicians want: contextualizing explainable machine learning
  for clinical end use.
\newblock \emph{arXiv preprint arXiv:1905.05134}, 2019.

\bibitem[Rasmussen(2003)]{rasmussen2003gaussian}
Carl~Edward Rasmussen.
\newblock Gaussian processes in machine learning.
\newblock In \emph{Summer School on Machine Learning}, 2003.

\bibitem[Van~Essen et~al.(2013)Van~Essen, Smith, Barch, Behrens, Yacoub,
  Ugurbil, Consortium, et~al.]{van2013wu}
David~C Van~Essen, Stephen~M Smith, Deanna~M Barch, Timothy~EJ Behrens, Essa
  Yacoub, Kamil Ugurbil, Wu-Minn~HCP Consortium, et~al.
\newblock The wu-minn human connectome project: an overview.
\newblock In \emph{Neuroimage}, 2013.

\end{thebibliography}
\bibliographystyle{unsrtnat}
\newpage
\section*{Supplementary Material}
 \subsection*{Benchmark Details: Saliency Methods}
 We compare popular backpropagation-based  and  perturbation based post-hoc saliency methods; each method provides feature importance, or "relevance", at a given time step to each input feature in a network.The relevance $ R^c(x_{i,t})$ produced by saliency methods can be defined as:
 \begin{itemize}
     \item \textbf{Backpropagation-based methods}:

 \begin{itemize}
     \item \textit{\textbf{Gradient (GRAD)}}\cite{baehrens2010explain} the gradient of the output with respect to $x_{i,t}$:
     $$\frac{\partial S_c(X)}{\partial  x_{i,t}} $$

     \item \textit{\textbf{Integrated Gradients (IG)}} \cite{sundararajan2017axiomatic} uses the average
gradient while input changes from a non-informative reference point $\overline{X}$ to ${X}$. The relevance $ R^c(x_{i,t})$ will depend upon the choice the reference point $\overline{X}$ (which is often set to zero).
\begin{equation*}
 \left( x_{i,t} -  \overline{x}_{t_{i}} \right) \times \int_{\alpha=0}^{1} \frac{\partial S_c\left(\overline{X}+\alpha \left(X-\overline{X}\right)\right)}{\partial  x_{i,t}} d\alpha
\end{equation*}

\item \textit{\textbf{SmoothGrad (SG)}} \cite{smilkov2017smoothgrad}
computes the gradient $n$ times adding Gaussian noise $\mathcal{N}(0,\,\sigma^{2})$  with standard deviation $\sigma$ to the input at each time.

\begin{equation*}
\frac{1}{n} \sum_1^n \frac{\partial S_c(X +  \mathcal{N}(0,\,\sigma^{2}))}{\partial  x_{i,t}}
\end{equation*}

    \item  \textit{\textbf{DeepLIFT (DL)}} \cite{shrikumar2017learning}
     a back-propagation based approach that defines a reference point and  compares the activation of each neuron to its reference activation; assigning relevance according to the difference.
    
     \item \textit{\textbf{Gradient SHAP (GS)}} \cite{lundberg2017unified} relevance is computed by adding Gaussian noise to each input sample multiple times (similar to SmoothGrad), selects a point along the path between a reference point and input, and computes the gradient of outputs with respect to those selected points. The Shapley value is the expected value of the gradients multiplied by the difference between input and reference point.

     \item \textit{\textbf{Deep SHAP (DeepLIFT + Shapley values) (DLS)}} \cite{lundberg2017unified}  Approximates the SHAP values using DeepLIFT; instead of a single reference point DeepLIFT takes a distribution of baselines computes the attribution for each input-baseline pair and averages the resulting attributions per input example; Shapley equations are used to linearize components such as max, softmax, products, divisions, etc..
     
    \end{itemize}

    \item \textbf{Perturbation-based:}
     \begin{itemize}
          \item \textit{\textbf{Feature Occlusion (FO)}} \cite{zeiler2014visualizing}  computes attribution as the difference in output after replacing each contiguous region with a given baseline. For time series we considered continuous regions as features with in same time step or multiple continuous  time steps.
      \item  \textit{\textbf{Feature Ablation (FA)}} \cite{suresh2017clinical}
 involves replacing each input feature with a given baseline, and computing the difference in output. Input features can also be grouped and ablated together rather than individually. 

  \item  \textit{\textbf{Feature permutation (FP)}} \cite{molnar2020interpretable}
  randomly permutes the feature values within a batch and computes the change in output as a result of this modification.  Similarly, to  feature ablation input features can also be grouped and ablated together rather than individually. 
     
    \end{itemize}
    
    \item \textbf{Others:}
    \begin{itemize}
  \item  \textit{\textbf{Shapley Value Sampling (SVS)}} \cite{castro2009polynomial} 
        Shapley value measure the contribution of each input features by taking each permutation of the feature and adding them one-by-one to a given baseline and measuring the difference in the output after adding the features. Shapley Value Sampling is an approximation of Shapley values that  involves sampling some random permutations  of the input features and average the marginal contribution of features based the differences on these permutations.

\item \textbf{Random} as a control; we compare methods to a random assignment of importance.

    \end{itemize}
 
 \end{itemize}
 \subsection*{Benchmark Details: Dataset Design}
We design multiple synthetic datasets where we can control and examine different design aspects that emerge in typical time series datasets. Different dataset combinations are shown in Figure \ref{fig:DS_DIST}.  The specific features and the time intervals (dark red/blue areas) that are considered informative is varied between datasets to capture different scenarios of how features vary over time. 
As shown in Figure \ref{fig:DS_DIST}, we consider the following sub-levels:

\begin{itemize}
    \item \textbf{Shape Normal/Small:} We modify the classification difficulty by decreasing the number of informative features. For \textit{Middle box} and \textit{Moving box} datasets we consider two scenarios:  \textbf{Normal shape} where more than $35\%$ of overall features are informative. \textbf{Small shape} less then $10\%$ of overall features are informative.
    \item \textbf{Signal Normal/Moving:} The location of the importance box differs in each sample.
    \item \textbf{Positional Temporal/Feature:} The classification does not depend on the value of informative signal $\mu$, rather the position of informative features. \textbf{Temporal position} each class has a constant temporal position; however, the informative features in the informative temporal window change in between samples.  \textbf{Feature position} each class has a constant group of features that are informative; however, the time at which these groups are informative is different between samples. 
    \item \textbf{Rare Time/Feature:} Mimic anomalies in time series variables, identification of such deviations is important in anomaly detection tasks. \textbf{Rare Time} Most features in a small temporal window are informative; this can be static or moving, i.e., N/M.  \textbf{Rare Feature} a small group features are informative in most time steps. Note that in both rare cases, less than $5\%$ of overall features are informative. 
\end{itemize}

Each synthetic dataset is generated by seven different processes as shown in Figure \ref{fig:generation_methods}. Data generation and time sampling was done in an non-uniform manner using python TimeSynth \footnote{https://github.com/TimeSynth/TimeSynth} package.
The base time series were generation by the following processes note that $\varepsilon _{t} \sim \mathcal{N}(0,1)$

\begin{itemize}
    \item Gaussian noise with zero mean and unit variance. $$X_{t}=\varepsilon _{t}$$
     \item   Independent sequences sampled from a harmonic function. A sinusoidal wave was used with $f=2$.
     $$X(t) = \sin(2 \pi f t)+\varepsilon _{t} $$
       \item  Independent sequences sampled from a pseudo period function, where, $A_{t} \sim \mathcal{N}(0,0.5)$ and $f_{t} \sim \mathcal{N}(2,0.01)$
       $$X(t) = A_{t}\sin(2 \pi f_t t)+\varepsilon _{t} $$
      
    \item Independent sequences of an autoregressive time series process, where, $p=1$ and $\varphi=0.9$
    $$X_{t}=\sum _{{i=1}}^{p}\varphi _{i}X_{{t-i}}+\varepsilon _{t}$$
          \item Independent sequences of a continuous autoregressive  time series process, where,  $\varphi=0.9$ and 
          $\sigma=0.1$.
             $$X_{t}=\varphi X_{t-1} +\sigma ({1-\varphi})^2*\varepsilon +\varepsilon_t$$ 
     \item  Independent sequences of non–linear autoregressive moving average (NARMA) time series, where, the equation is given below, where $n=10$ and $U\sim U(0,0.5)$ is a uniform distribution.
 $$X_{t} = 0.3X_{t-1} + 0.05X_{t-1} \sum_{i=0}^{n-1}X_{t-i} +  1.5 \space  U\left(t-\left(n-1\right)\right) * U(t) + 0.1 + \varepsilon _{t}$$
        \item   Independent sequences sampled  according to a Gaussian Process mixture  model with selected covariance function \cite{rasmussen2003gaussian}. 

\end{itemize}

 Informative features are then highlighted by the addition of a constant $\mu$
to positive class and subtraction of $\mu$ from negative class (unless specified, $\mu=1$).

 \textbf{Multivariate MNIST time series} is included as a more general case, each sample has 28 time steps, and the feature embedding size is 28.
 \subsection*{Benchmark Details: Saliency Distribution}
\subsubsection*{Real Data}
\begin{figure*}[tbh!]
\centering
\includegraphics[width=\textwidth]{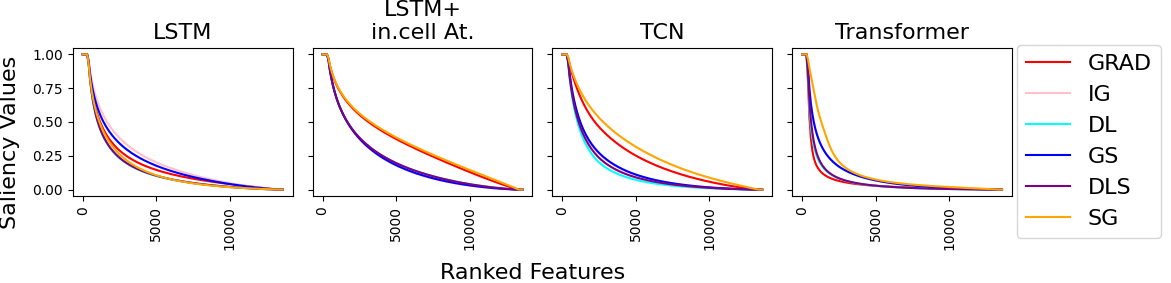}
\caption{The distribution of saliency values of ranked features produced by different saliency methods for HCP fMRI data. }
\label{fig:HCP_DS}
\end{figure*}

\paragraph{Human Connectome Project fMRI Data:}
To inspect saliency distribution in a more realistic setting, we apply different saliency methods and plot the distribution of ranked features on an openly available fMRI dataset of the Human Connectome Project (HCP) \cite{van2013wu}.  In this dataset, subjects are performing specific tasks while scanned by an fMRI machine. Our classification problem is to identify the task performed, given the fMRI scans. The distribution of different saliency methods across multiple neural architectures is shown in Figure \ref{fig:HCP_DS}.
Similar to synthetic data, saliency exponentially decays with feature ranking.

\begin{figure*}[tbh!]
\centering
\includegraphics[width=\textwidth]{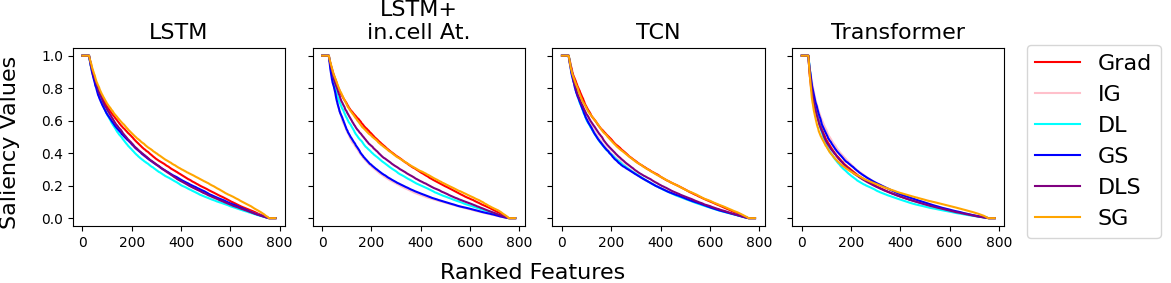}
\caption{The distribution of saliency values of ranked features produced by different saliency methods for Time Series MNIST. }
\label{fig:MNIST_DS}
\end{figure*}

\begin{figure}[htbp!]
\centering
\begin{tabular}{cc}
\includegraphics[width=.4\textwidth]{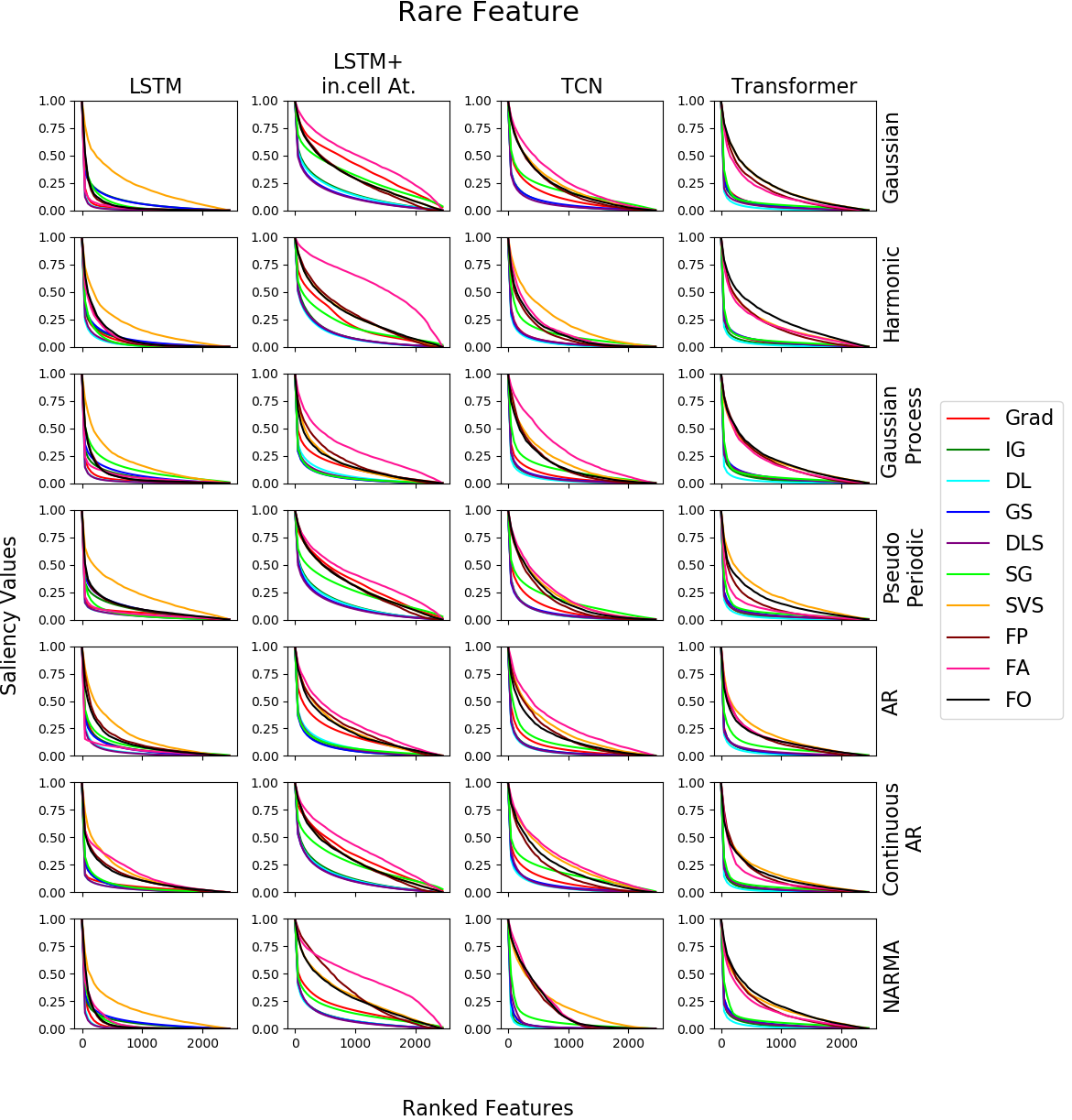}&
\includegraphics[width=.4\textwidth]{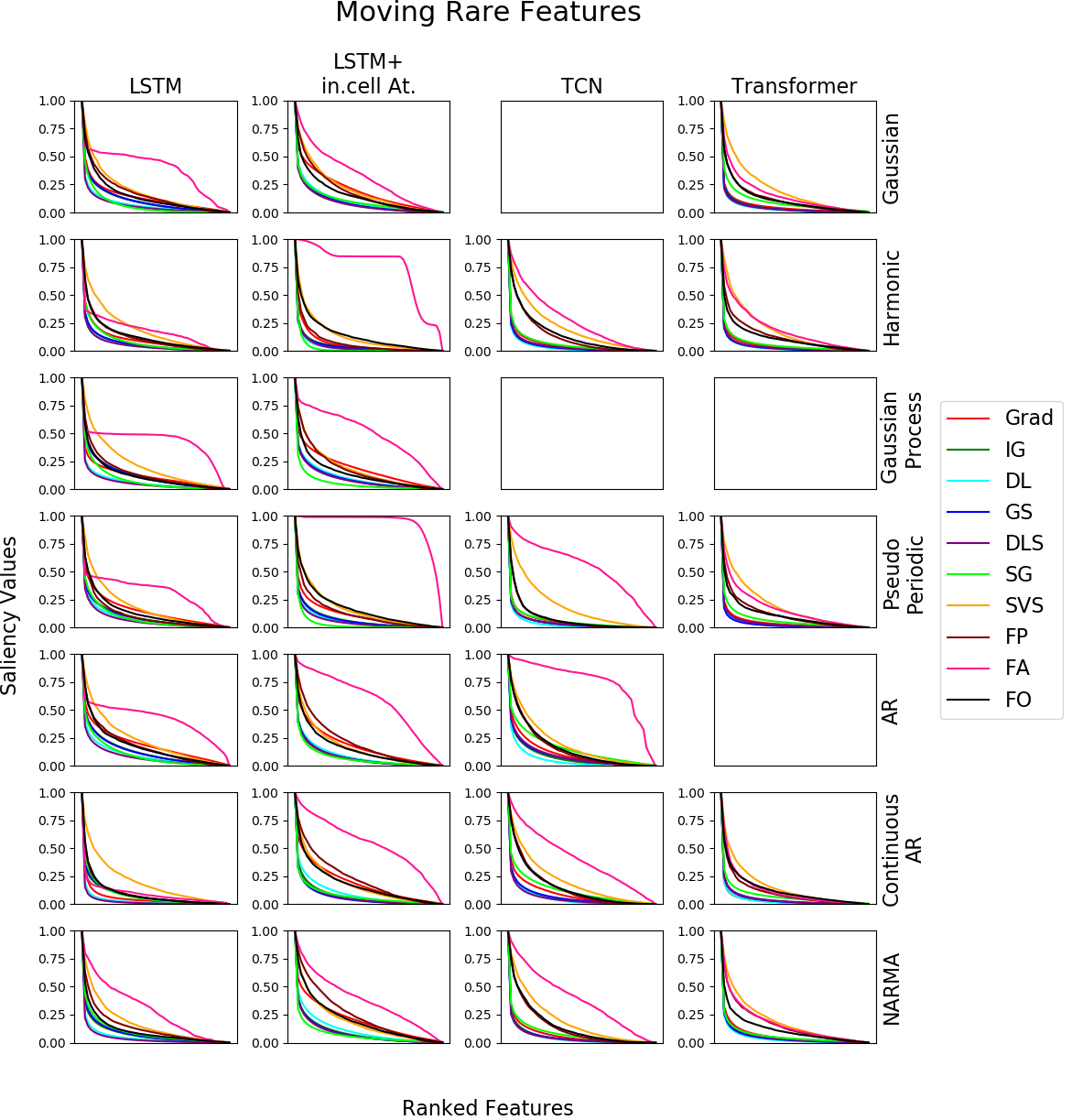}\\

\includegraphics[width=.4\textwidth]{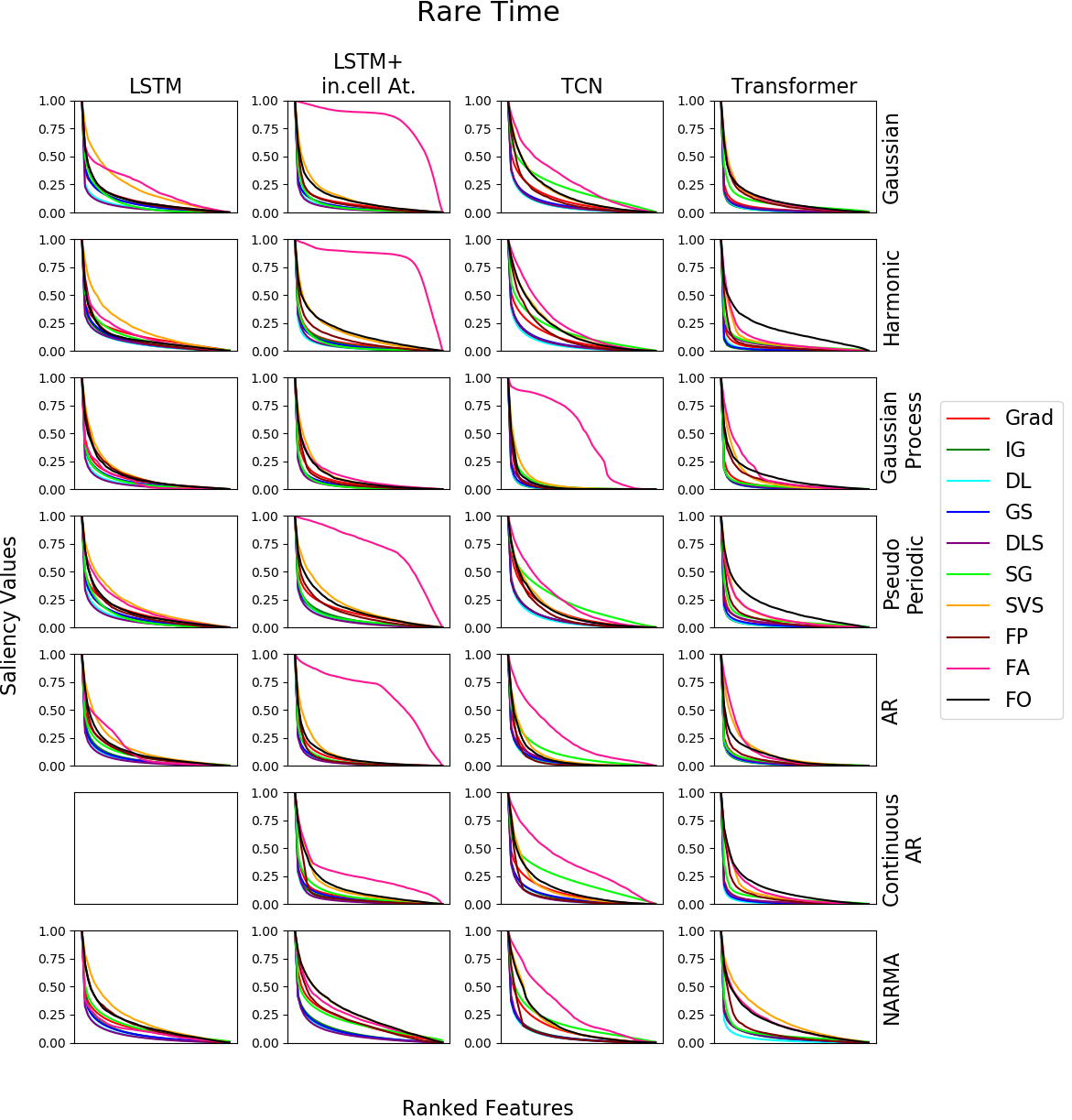}&
\includegraphics[width=.4\textwidth]{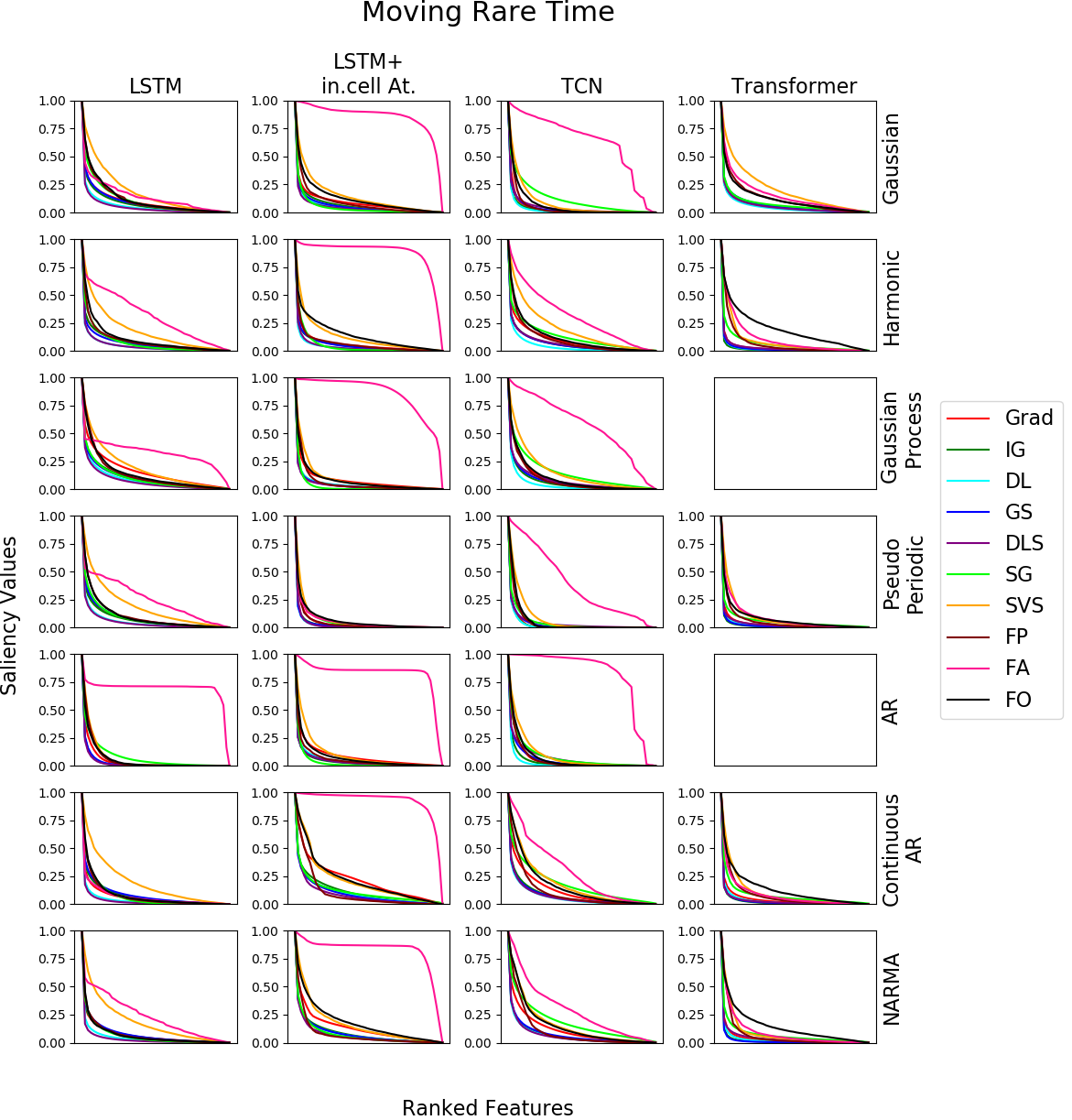}\\
\includegraphics[width=.4\textwidth]{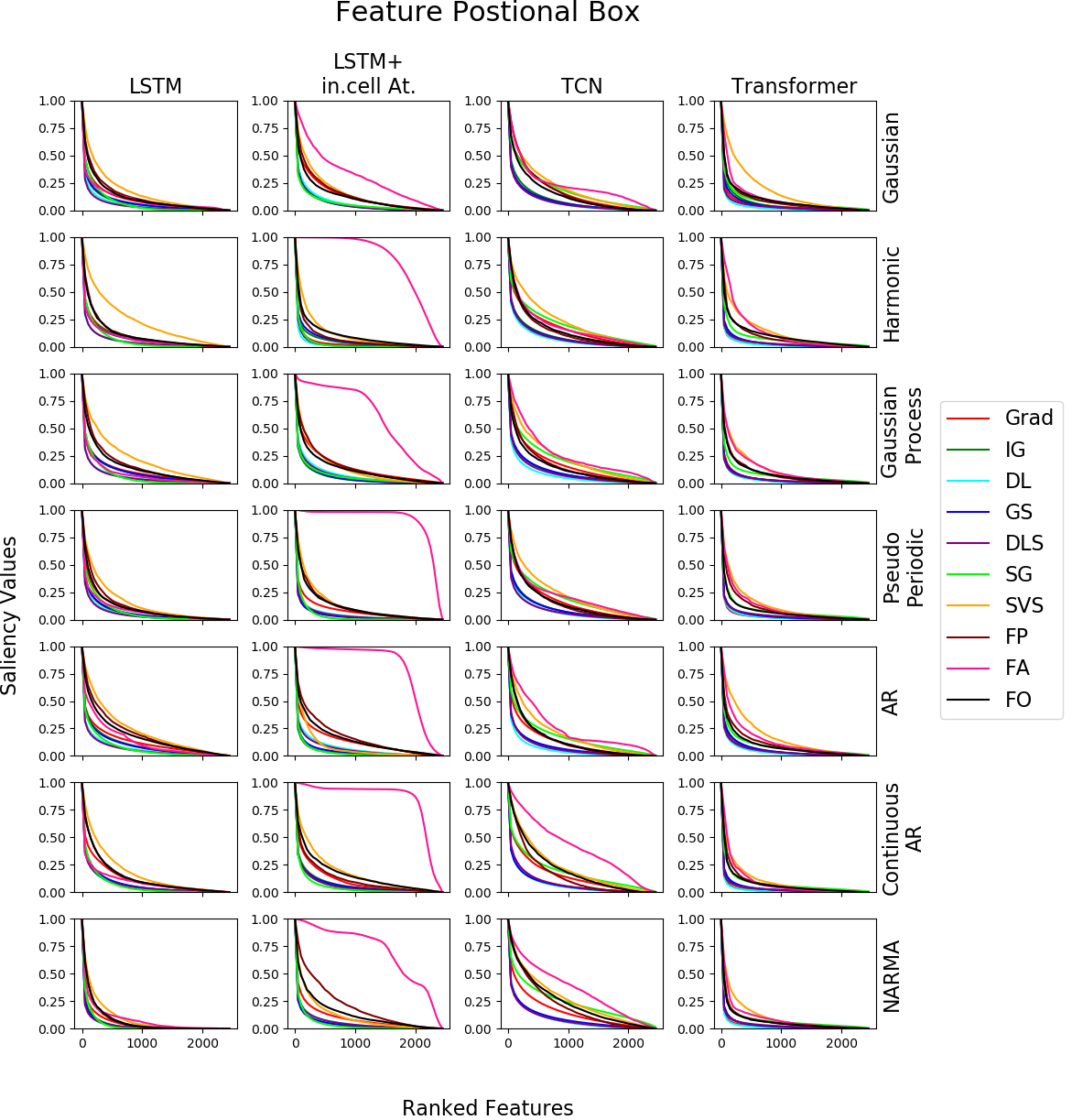}&
\includegraphics[width=.4\textwidth]{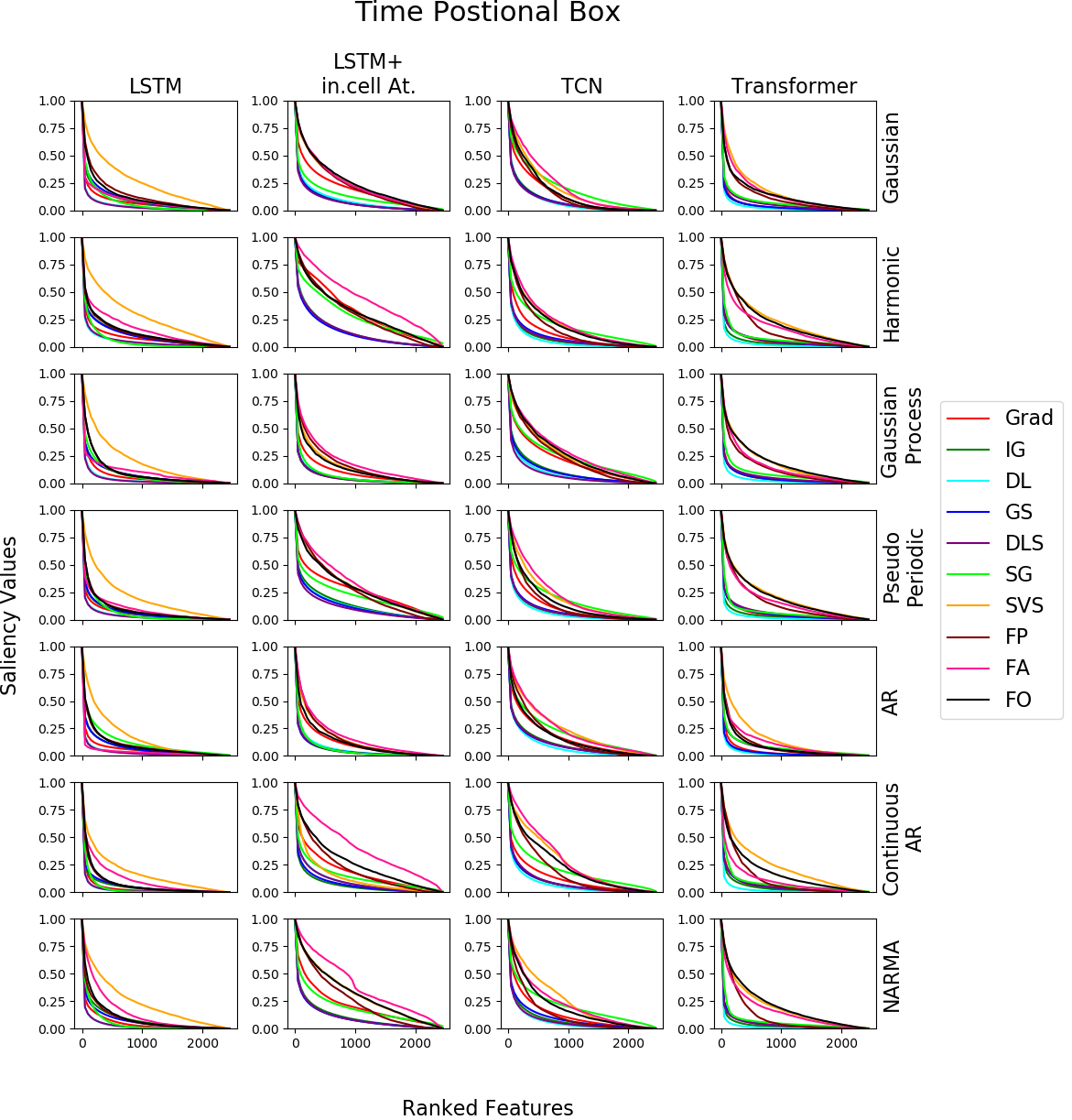}\\
\end{tabular}
\caption{The distribution of saliency values of ranked features produced by different saliency methods for various synthetic datasets. Empty spaces indicates that model was not able to learn classification task for the given dataset.
}
\label{fig:Synthetic_DS}
\end{figure}

\paragraph{Time Series MNIST:} Figure \ref{fig:MNIST_DS} shows the saliency distribution for different neural architecture on multivariate MNIST.
\paragraph{ Synthetic Data:}
Figure \ref{fig:Synthetic_DS} shows the saliency distribution for different \textit{(neural architecture, saliency method)} pairs. Aside from feature ablation, saliency decays exponentially with feature ranking, the distribution across different methods and datasets seem to be similar for a neural architecture.

 \subsection*{Benchmark Details: Performance Evaluation Metrics}

Given the synthetic data described earlier, informative features are known (dark areas in Figure \ref{fig:DS_DIST}), and we can calculate precision and recall of each \textit{(neural architecture, saliency method)} pair using the confusion matrix in Table \ref{tab:Confusion}.
 
\begin{table}[htb]
    \centering
\begin{tabular}{c||c|c}
        \backslashbox{Saliency}{Actual}  & Informative & Noise \\
          \hline
    High  &  \small{True Positive (TP)} &  \small{False Positive (FP)}  \\
          \hline
    Low& \small{False Negative (FN)} &  \small{True Negative (TN)}\\
    \end{tabular}%
    \caption{Confusion Matrix, for precision and recall calculation.}
    \label{tab:Confusion}
\end{table}

\subsubsection*{Precision} 
The fraction of informative high saliency features among all high saliency features. Since the saliency value varies dramatically across features, we do not look at the number of true positive and false negative instead their saliency value; the (weighted) precision is calculated by:
\begin{equation*}
  \frac{\sum R\left(x_{t_{i}}\right) \{x_{t_{i}} : x_{t_{i}} \in TP\} } {\sum R\left(x_{t_{i}}\right) \{x_{t_{i}} : x_{t_{i}} \in TP\} + \sum R\left(x_{t_{i}}\right) \{x_{t_{i}} : x_{t_{i}} \in FP\}}
  \end{equation*}
\subsubsection*{Recall}
The fraction of the total informative features that had high saliency, similar to the precision we use the saliency value rather than the count. (Weighted) the recall is defined as:
\begin{equation*}
    \frac{\sum R\left(x_{t_{i}}\right) \{x_{t_{i}} : x_{t_{i}} \in TP\} } {\sum R\left(x_{t_{i}}\right) \{x_{t_{i}} : x_{t_{i}} \in TP\} + \sum R\left(x_{t_{i}}\right) \{x_{t_{i}} : x_{t_{i}} \in FN\}}
\end{equation*}

Through our experiments, we report area under the precision curve (AUP), the area under the recall curve (AUR), and area under precision and recall (AUPR). The curves are calculated by the precision/recall values at different levels of degradation. We also consider feature/time precision and recall (a feature is considered informative if it has information at any time step and vice versa). For the random baseline, we stochastically select a saliency method then permute the saliency values producing arbitrary ranking.
\subsubsection*{Temporal Saliency Rescaling Optimizations and Complexity}
The main back draw of {\bf T}emporal {\bf S}aliency {\bf R}escaling (Algorithm \ref{algo:TSR}) is the increase in complexity that is a result of performing multiple gradient calculations.  Algorithm \ref{algo:TSR_withFeatureGrouping} shows a variation of the algorithm that calculates the contribution of a group of features within a time step. Algorithm \ref{algo:TFSR} calculates the contribution of each time step and feature independently; the total contribution of a single feature at a given time is the product of feature and time contributions.

\begin{algorithm}[ht!]
  \KwInput{ input $X$, a baseline interpretation method $R(.)$, feature group size $G$}
  \KwOutput{TSR interpretation method $R^{TSR+FG}(.)$}

 \For{$t\gets0$ \KwTo $T$ }{
    Mask all features at time $t$: $\overbar X_{:,t}=0$, otherwise $\overbar X=X$\;
    Compute Time-Relevance Score $\Delta_t^{time} = \sum_{i,t} |R_{i,t}(X) -R_{i,t}(\overbar X)|$\;
    }
    
     \For{$t\gets0$ \KwTo $T$ }{

        \For{$i\gets0,G,2G,\dots, N$}{
         \eIf{$\Delta_t^{time}> \alpha$}{
        Mask features $i:i+G$ at time $t$: $\overbar X_{i:i+G,t}=0$, otherwise $\overbar X=X$\;
         Compute Feature-Relevance Score $\Delta_{i:i+G}^{feature} = \sum_{i,t} |R_{i,t}(X) -R_{i,t}(\overbar X)|$\;
        }
        { Feature-Relevance Score
        $\Delta_{i:i+G}^{feature}=0$\;
        }
 Compute (time,feature) importance score $R^{TSR+FG}_{i,t}=\Delta_{i:i+G}^{feature}\times \Delta_t^{time}$ \;
   }
    }
 \caption{Temporal Saliency Rescaling (TSR) With Feature Grouping}
 \label{algo:TSR_withFeatureGrouping}
\end{algorithm}

\begin{algorithm}[htb!]
  \KwInput{ input $X$, a baseline interpretation method $R(.)$}
  \KwOutput{TFSR interpretation method $R^{TFSR}(.)$}

 \For{$t\gets0$ \KwTo $T$ }{
    Mask all features at time $t$: $\overbar X_{:,t}=0$, otherwise $\overbar X=X$\;
    Compute Time-Relevance Score $\Delta_t^{time} = \sum_{i,t} |R_{i,t}(X) -R_{i,t}(\overbar X)|$\;
    }
   
    \For{$i\gets0$ \KwTo $N$ }{
    Mask all time steps for feature $i$: $\overbar X_{i,:}=0$, otherwise $\overbar X=X$\;
    Compute Feature-Relevance Score $\Delta_i^{feature} = \sum_{i,t} |R_{i,t}(X) -R_{i,t}(\overbar X)|$\;
    }

     \For{$t\gets0$ \KwTo $T$ }{

        \For{$i\gets0$ \KwTo $N$}{
         
 Compute (time,feature) importance score $R^{TFSR}_{i,t}=\Delta_i^{feature}\times \Delta_t^{time}$ \;
   }
    }
 \caption{Temporal Feature Saliency Rescaling (TFSR)}
 \label{algo:TFSR}
\end{algorithm}
The approximate relevance calculations needed for each variation is shown in table \ref{tab:TSR_complexity}. The complexity \textbf{TSR} and \textbf{TSR With Feature Grouping} highly depends on $\alpha$. In many time series applications such as anomaly detection, $\alpha$ can be set to be close to 1.
\textbf{TFSR}  complexity is comparable to SmoothGrad. Other approaches have proposed similar trade-offs between interpretability and computational complexity, i.e., \citet{hooker2019benchmark} proposed retraining the entire network after removing salient features, retraining even most simple networks is very expensive in comparison to extra gradient calculations.

\begin{table}[htb!]
    \centering
  \begin{tabular}{l|c}
  Algorithm  & \multicolumn{1}{l}{Approximate number of Relevance Calculations} \\
        \hline \hline
    Algorithm \ref{algo:TSR}: $R^{TSR}(.)$ & $T+(T*(1-\alpha)*N)$ \\
     Algorithm \ref{algo:TSR_withFeatureGrouping}: $R^{TSR+FG}(.)$  &$T+(T*(1-\alpha)*N/G)$   \\
     Algorithm \ref{algo:TFSR}: $R^{TFSR}(.)$    &$T+N$  \\
      \hline                                                                              
    \end{tabular}%
    
        \caption{Complexity analysis of different varaitions of \textbf{TSR}}
    \label{tab:TSR_complexity}
\end{table}

\subsection*{Experiments and Results}

\subsubsection*{Saliency Map Quality}
\begin{figure*}[htb!]
\centering
\includegraphics[width=1\textwidth]{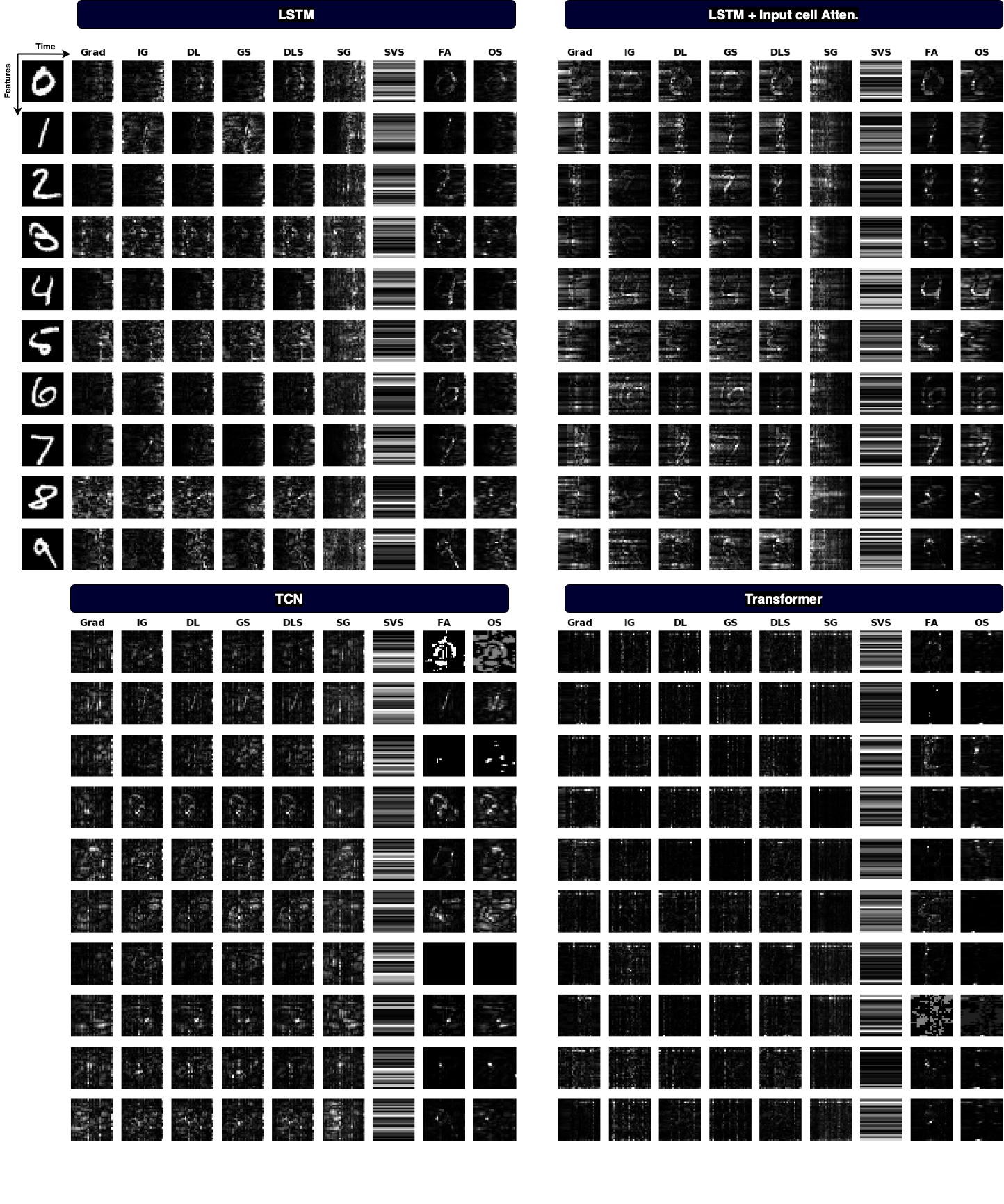}
\caption{Saliency maps produced by Gradient-based saliency methods including Grad, Integrated Gradients, DeepLIFT, Gradient SHAP, DeepSHAP and SmoothGrad and non-gradient-based saliency method including Shap value sampling, Feature Ablation and Feature Occlusion for 4 different models on time series MNIST (white represents high saliency).}
\label{fig:saliencyMaps_before}
\end{figure*}

\begin{figure*}[htb!]
\centering
\includegraphics[width=1\textwidth]{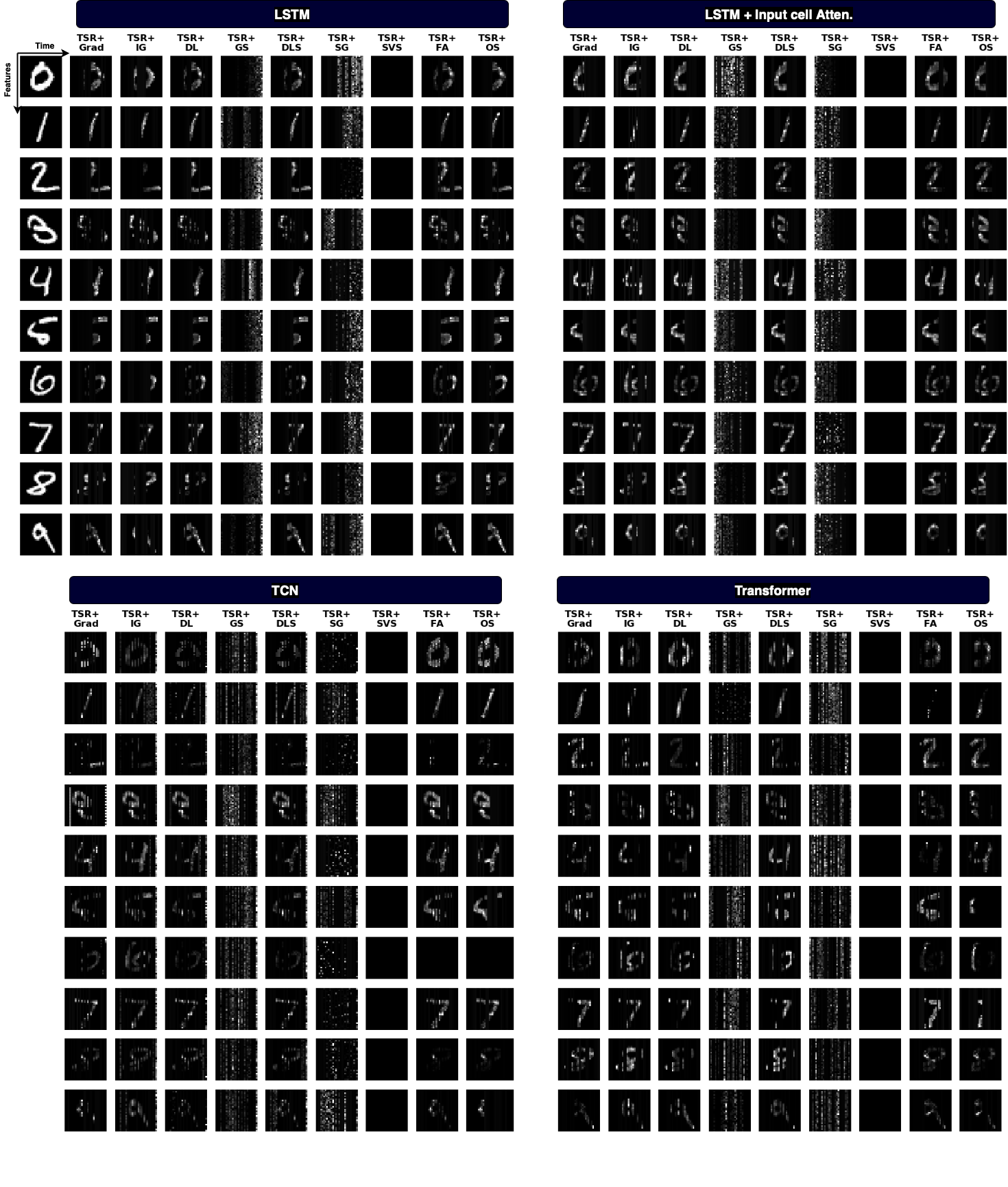}
\caption{Saliency maps when applying the proposed Temporal Saliency Rescaling (TSR) approach on different saliency methods.}
\label{fig:saliencyMaps_after}
\end{figure*}

\begin{figure}[htb!]
\centering
\begin{tabular}{cc}
\includegraphics[width=.47\textwidth]{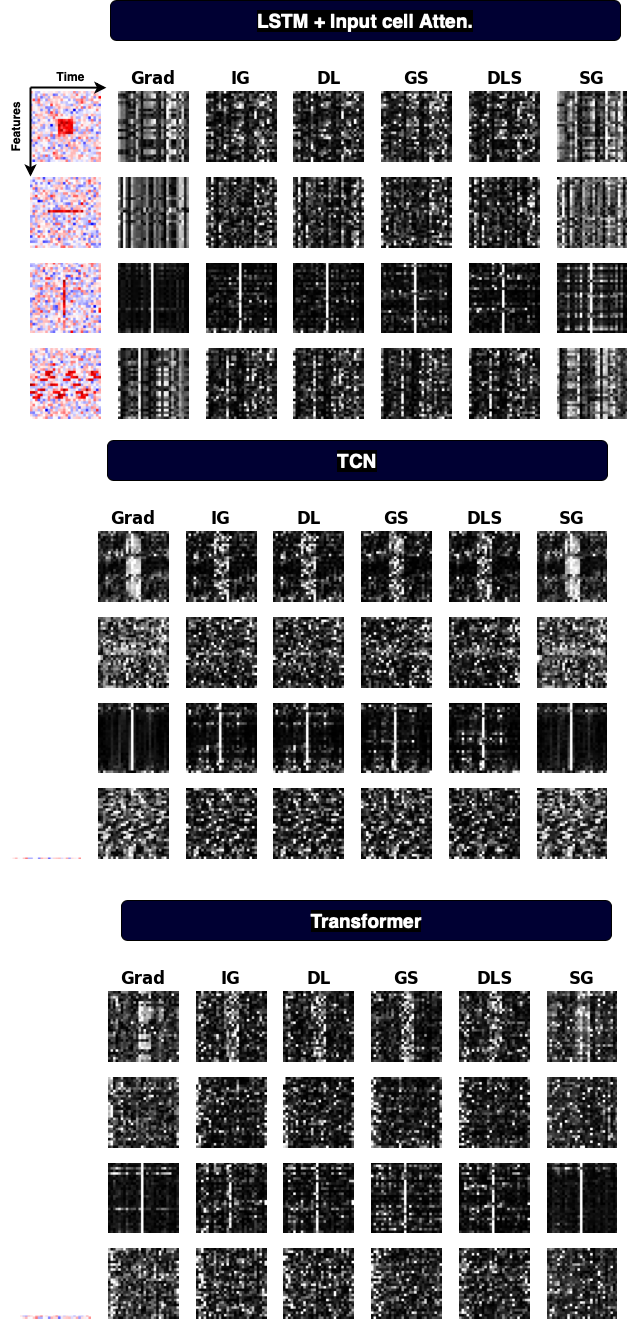}&
\includegraphics[width=.47\textwidth]{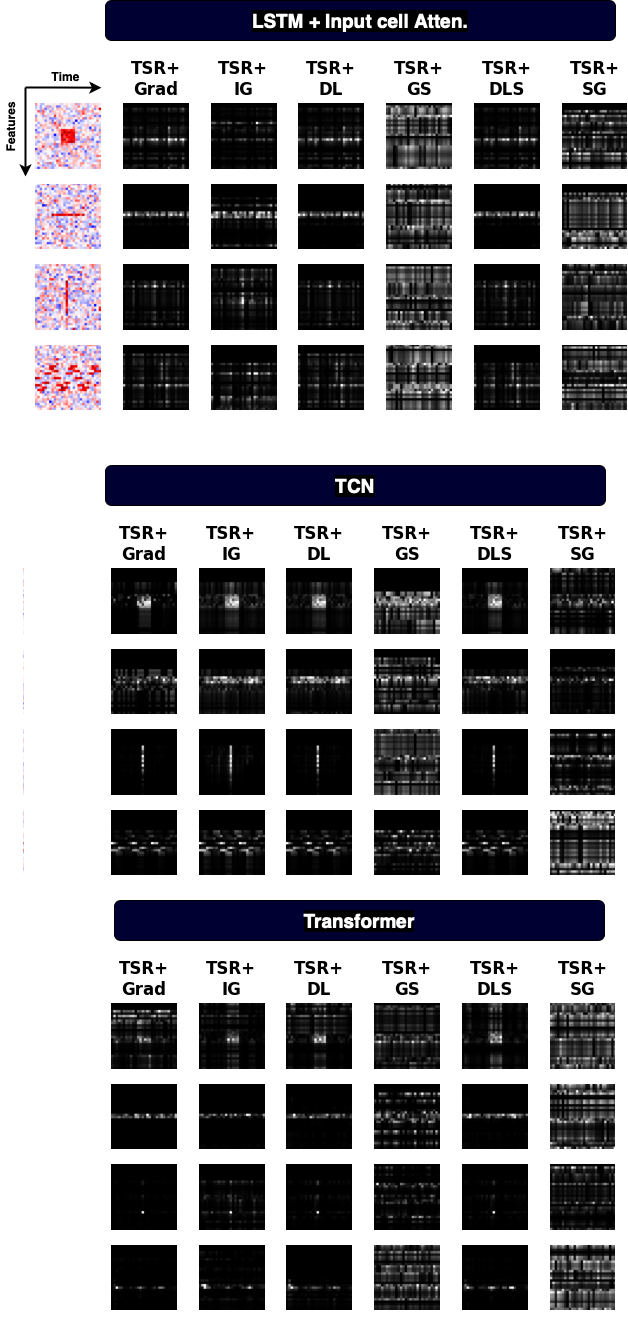}\\
\textbf{(a) }   & \textbf{(b)} 
\end{tabular}
\caption{Saliency maps produced by Grad, Integrated Gradients, DeepLIFT, Gradient SHAP, DeepSHAP, and SmoothGrad for three different models on static synthetic datasets. (b)Saliency maps when applying the proposed Temporal Saliency Rescaling (TSR) approach.
}
\label{fig:saliencyMaps_boxes}
\end{figure}

From figures  \ref{fig:saliencyMaps_before}, \ref{fig:saliencyMaps_after} and  \ref{fig:saliencyMaps_boxes}, when applying temporal saliency rescaling we observe a definite improvement in saliency quality across different architectures and interpretability methods except for Gradient SHAP and SmoothGrad.
\paragraph{MNIST}
 Figure \ref{fig:saliencyMaps_before} shows saliency maps produced by each \textit{(neural architecture, saliency method)} pair on samples from time series MNIST; Figure \ref{fig:saliencyMaps_after}, show the samples after applying \textbf{TSR}. There is a significant improvement in the quality of the saliency map after applying the temporal saliency rescaling approach.

\paragraph{Synthetic Datasets}
Figure \ref{fig:saliencyMaps_boxes} shows saliency maps produced by each \textit{(neural architecture, saliency method)} pair on samples from different synthetic datasets before and after applying \textbf{TSR}.

\FloatBarrier
\subsubsection*{Saliency Methods versus Random Ranking}
\begin{figure}[htb!]
\centering
\includegraphics[width=.75\textwidth]{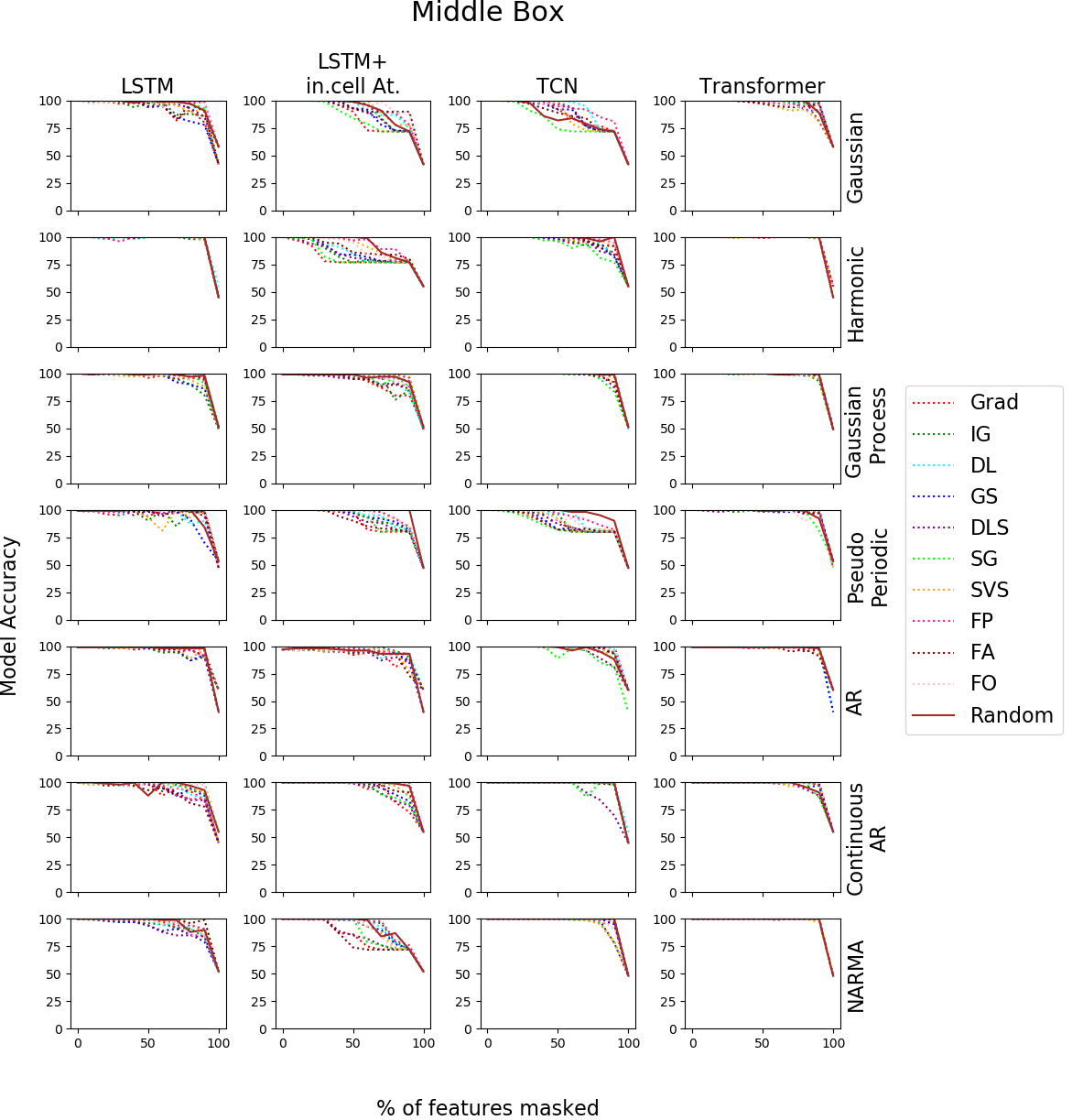} \hfill
\includegraphics[width=.5\textwidth]{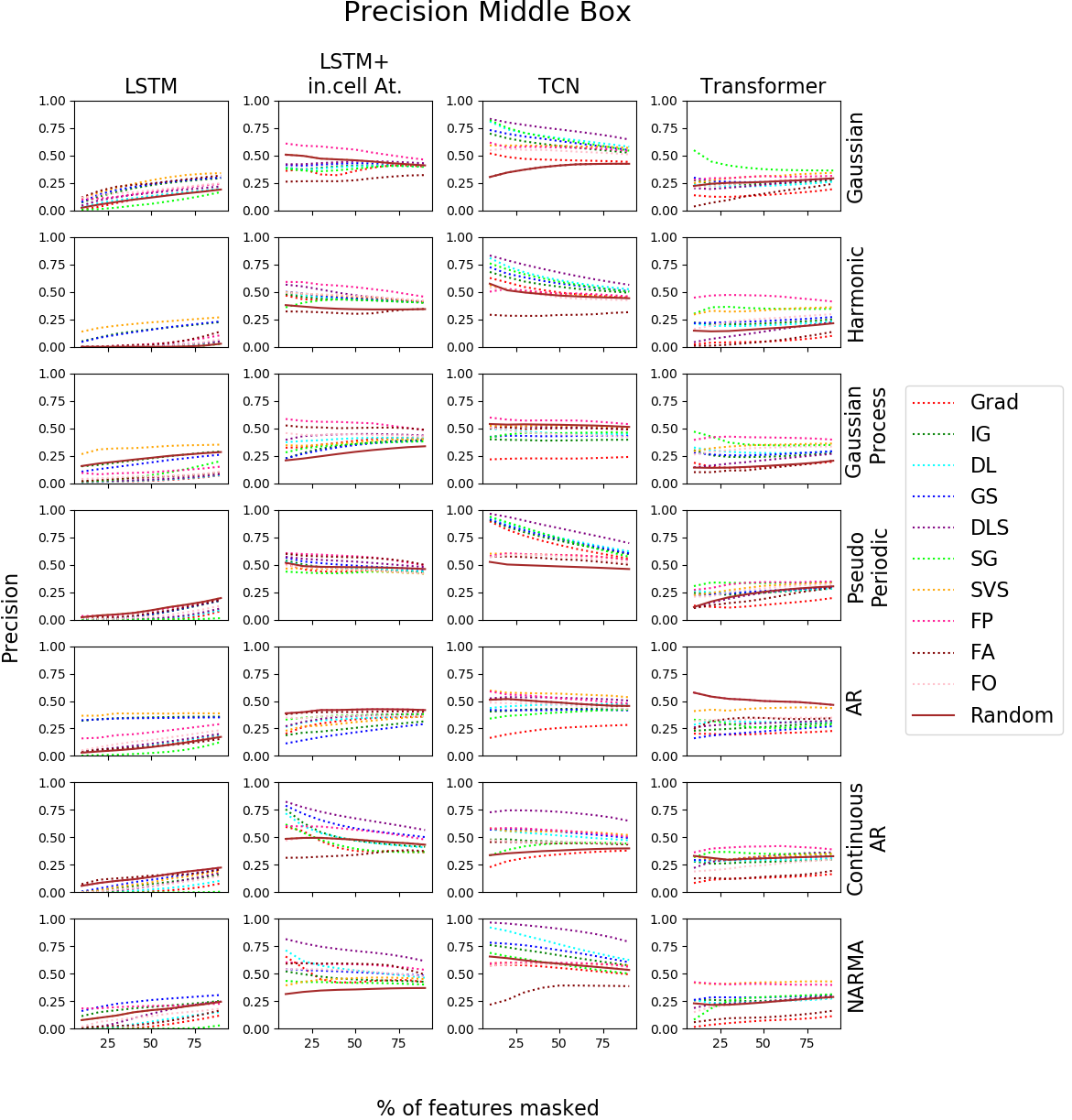}\hfill \includegraphics[width=.5\textwidth]{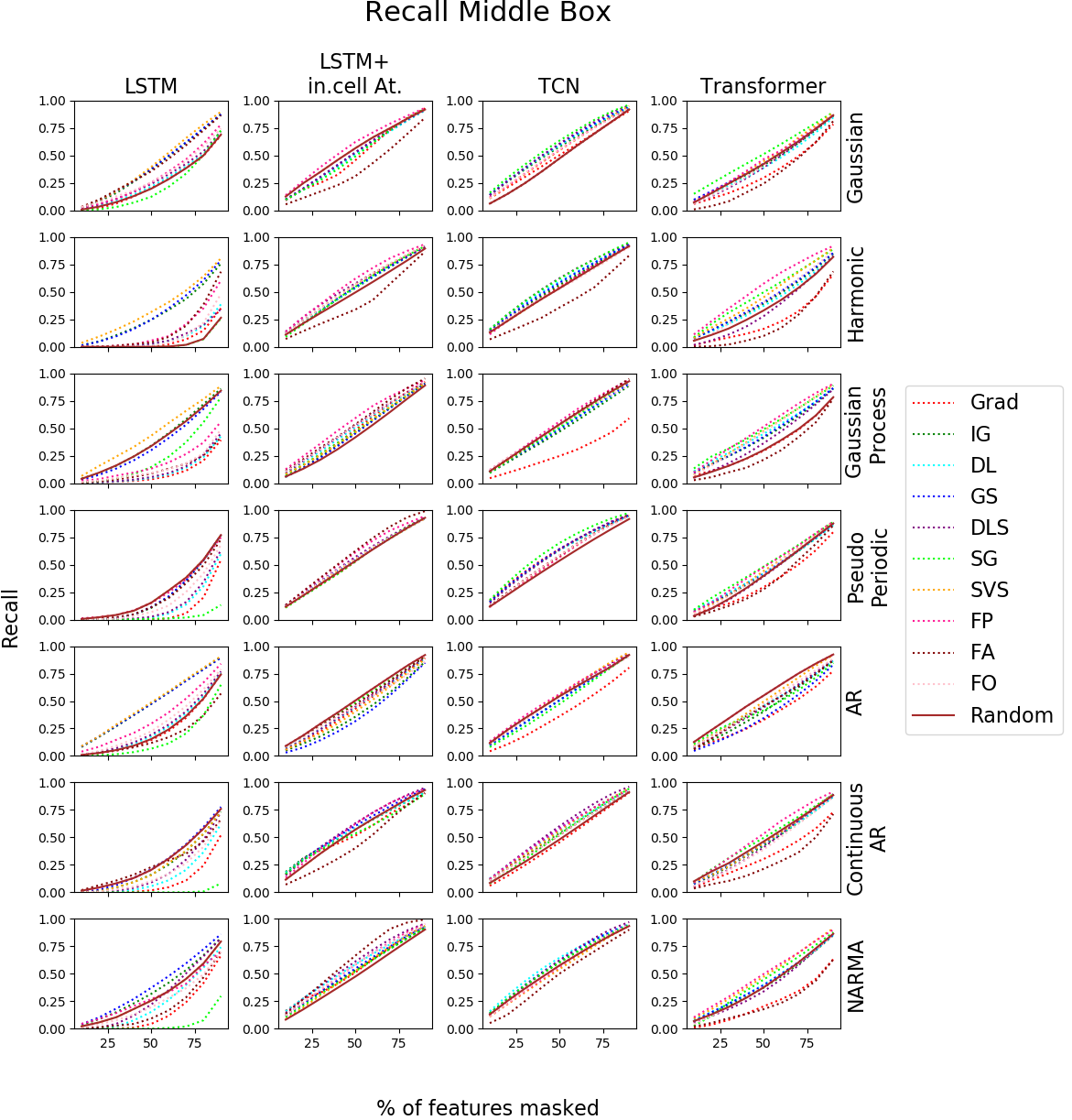}

\caption{Accuracy drop, precision and recall for \textit{\textbf{Middle box}} datasets
}
\newpage
\label{fig:MiddleBox}
\end{figure}

\begin{figure}[htb!]
\centering
\includegraphics[width=.75\textwidth]{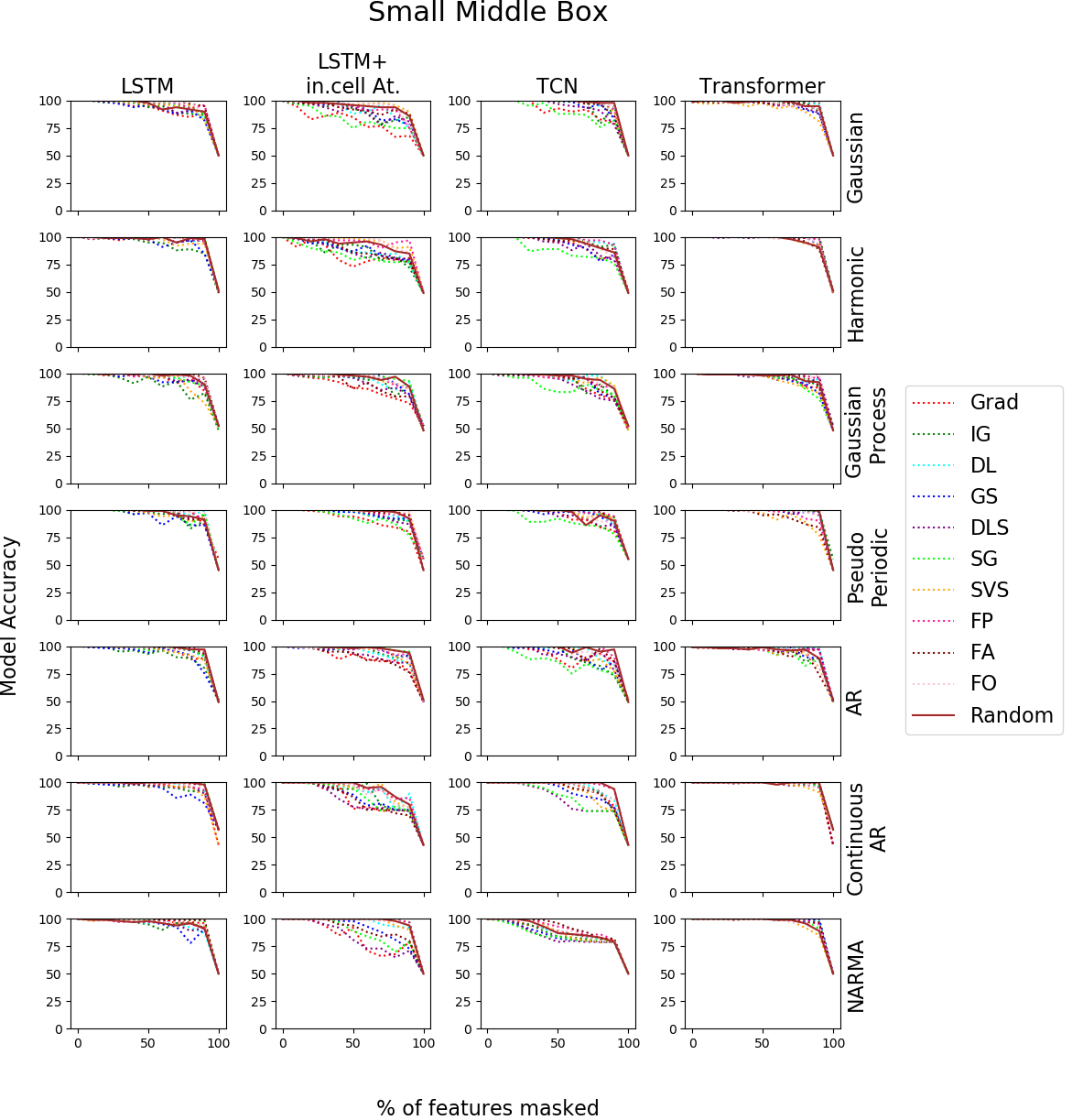} \hfill
\includegraphics[width=.5\textwidth]{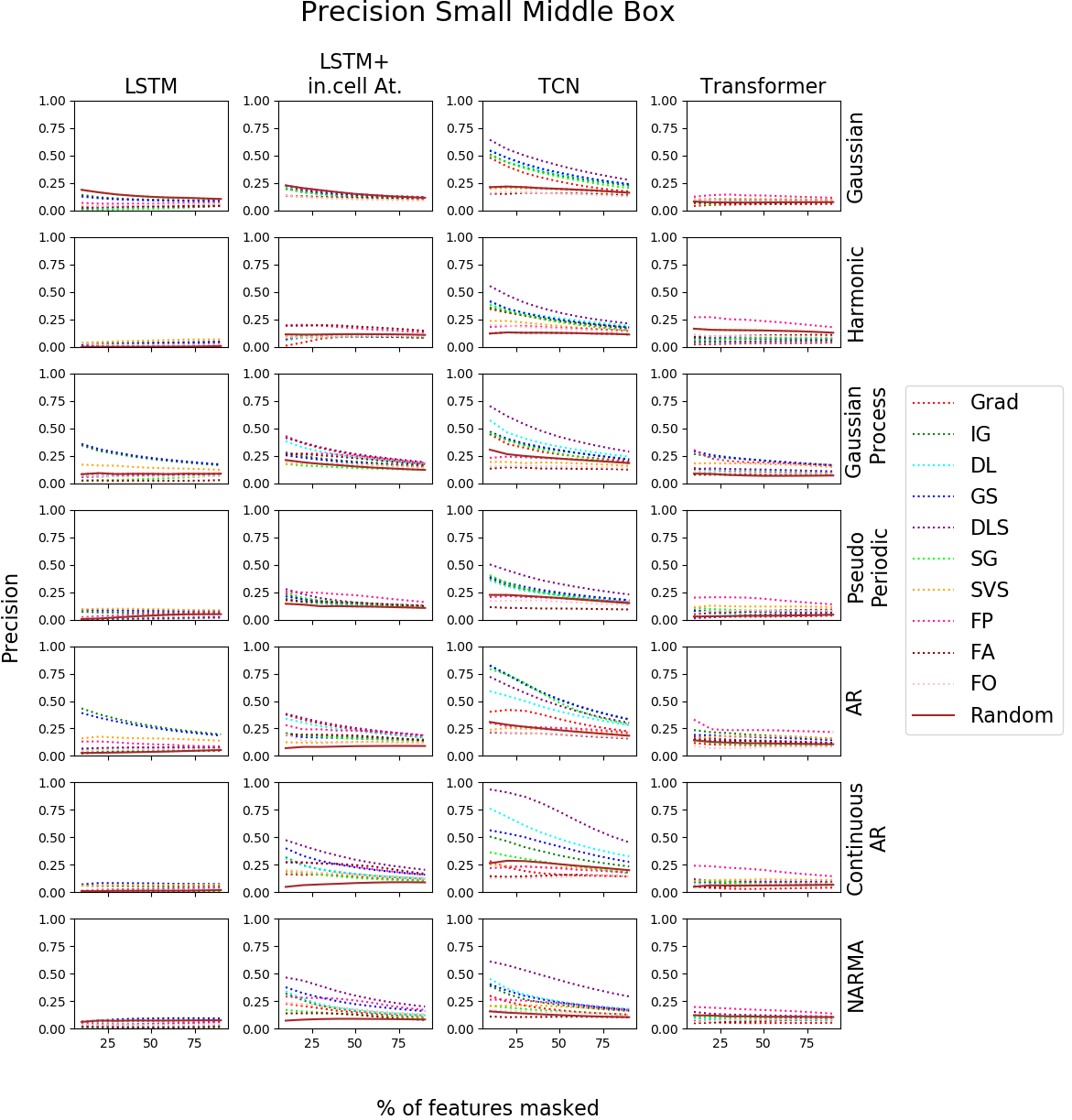}\hfill \includegraphics[width=.5\textwidth]{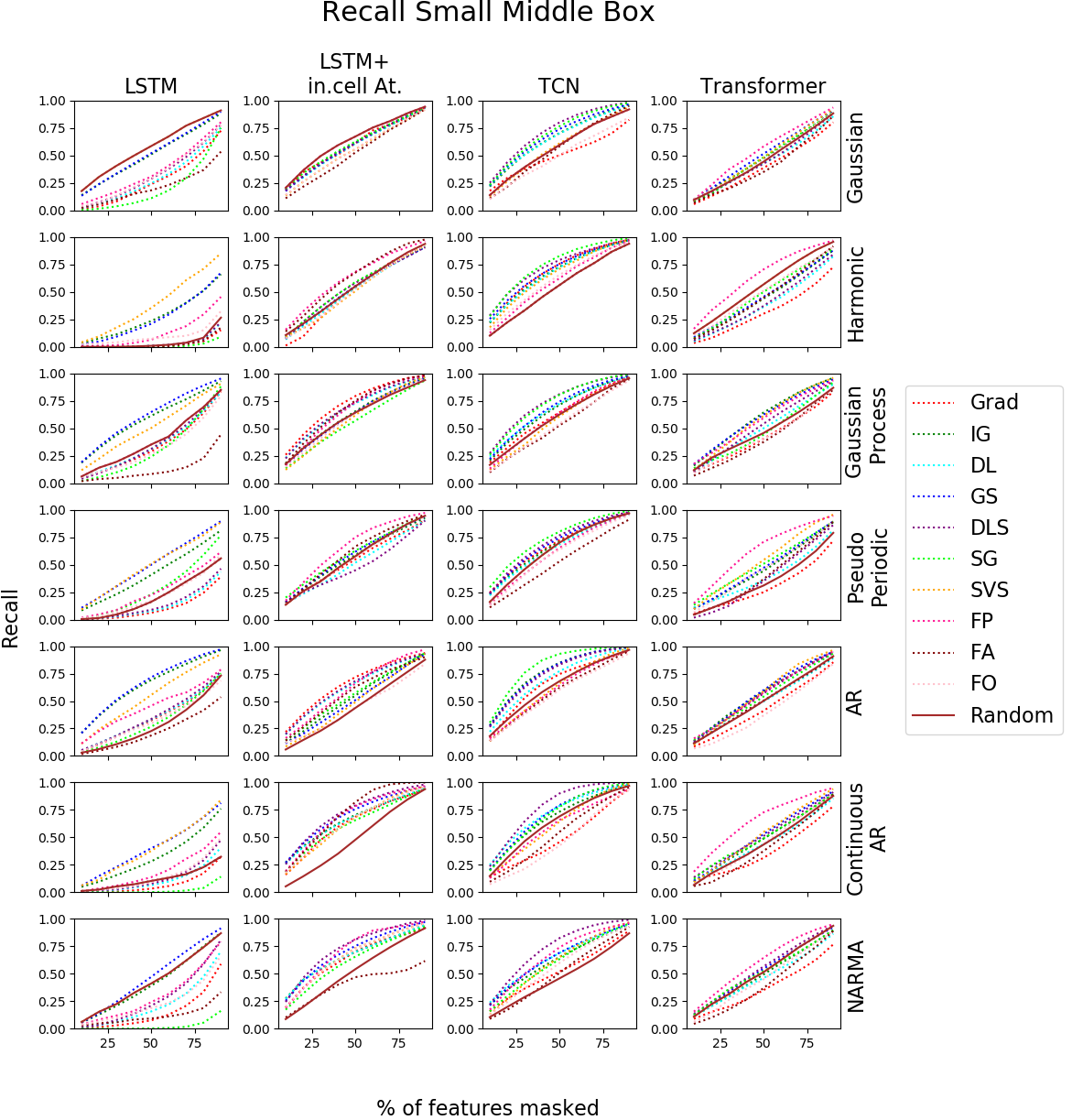}

\caption{Accuracy drop, precision and recall for  \textit{\textbf{Small Middle box}} datasets
}
\label{fig:SmallMiddleBox}
\end{figure}

\begin{figure}[htb!]
\centering
\includegraphics[width=.75\textwidth]{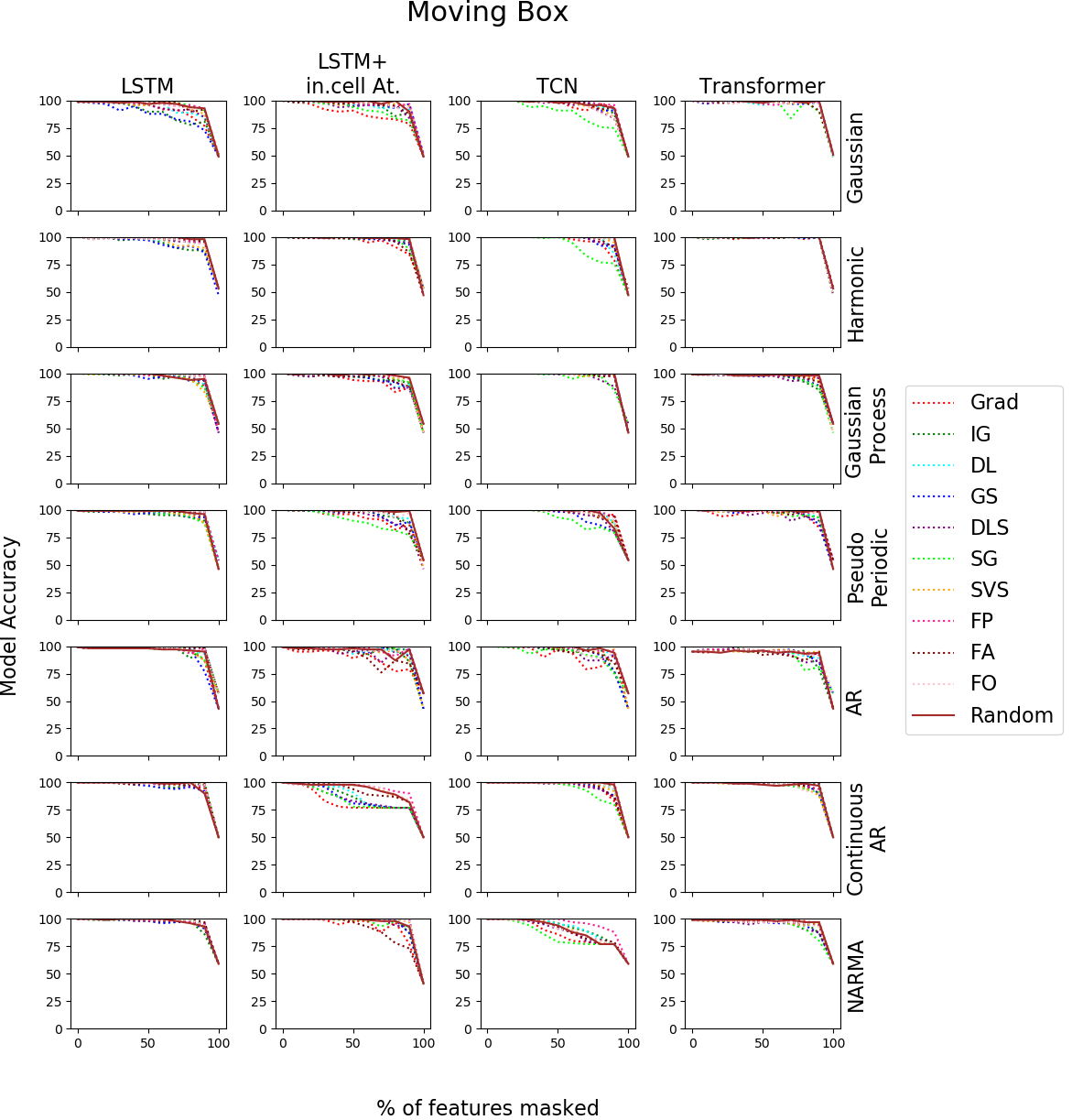} \hfill
\includegraphics[width=.5\textwidth]{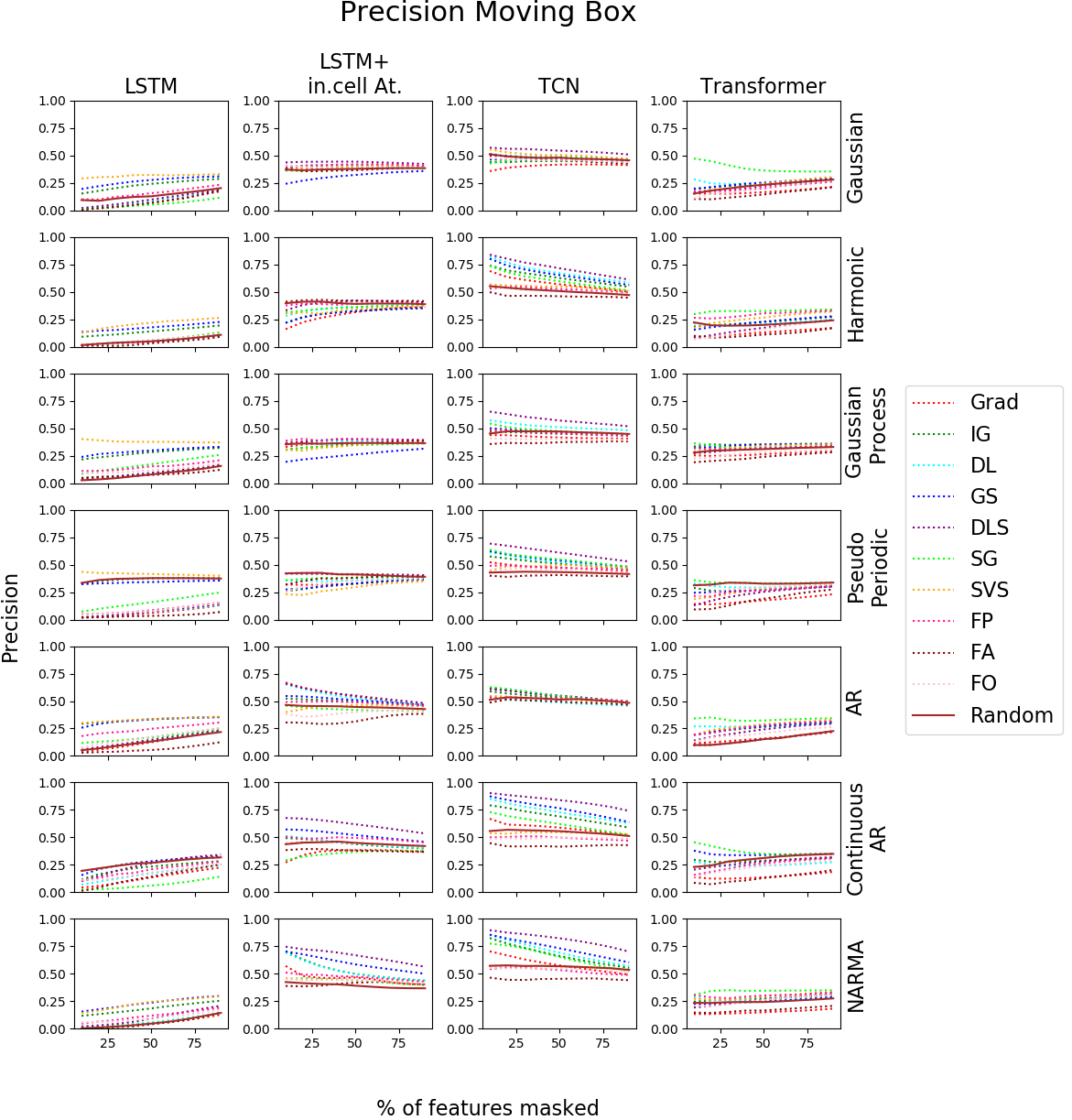}\hfill \includegraphics[width=.5\textwidth]{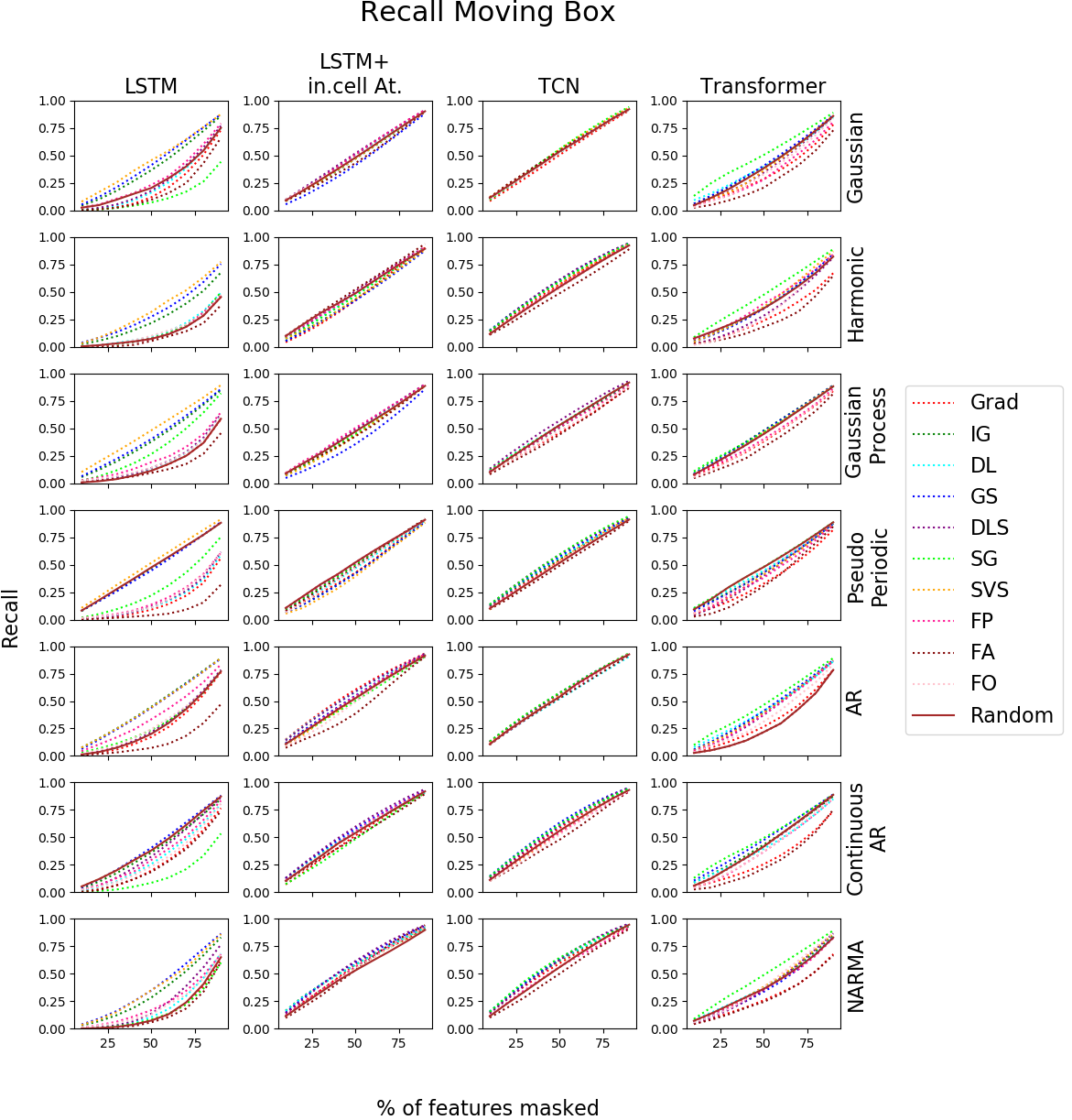}

\caption{Accuracy drop, precision and recall for  \textit{\textbf{Moving Middle box}} datasets
}
\label{fig:MovingMiddleBox}
\end{figure}

\begin{figure}[htb!]
\centering
\includegraphics[width=.75\textwidth]{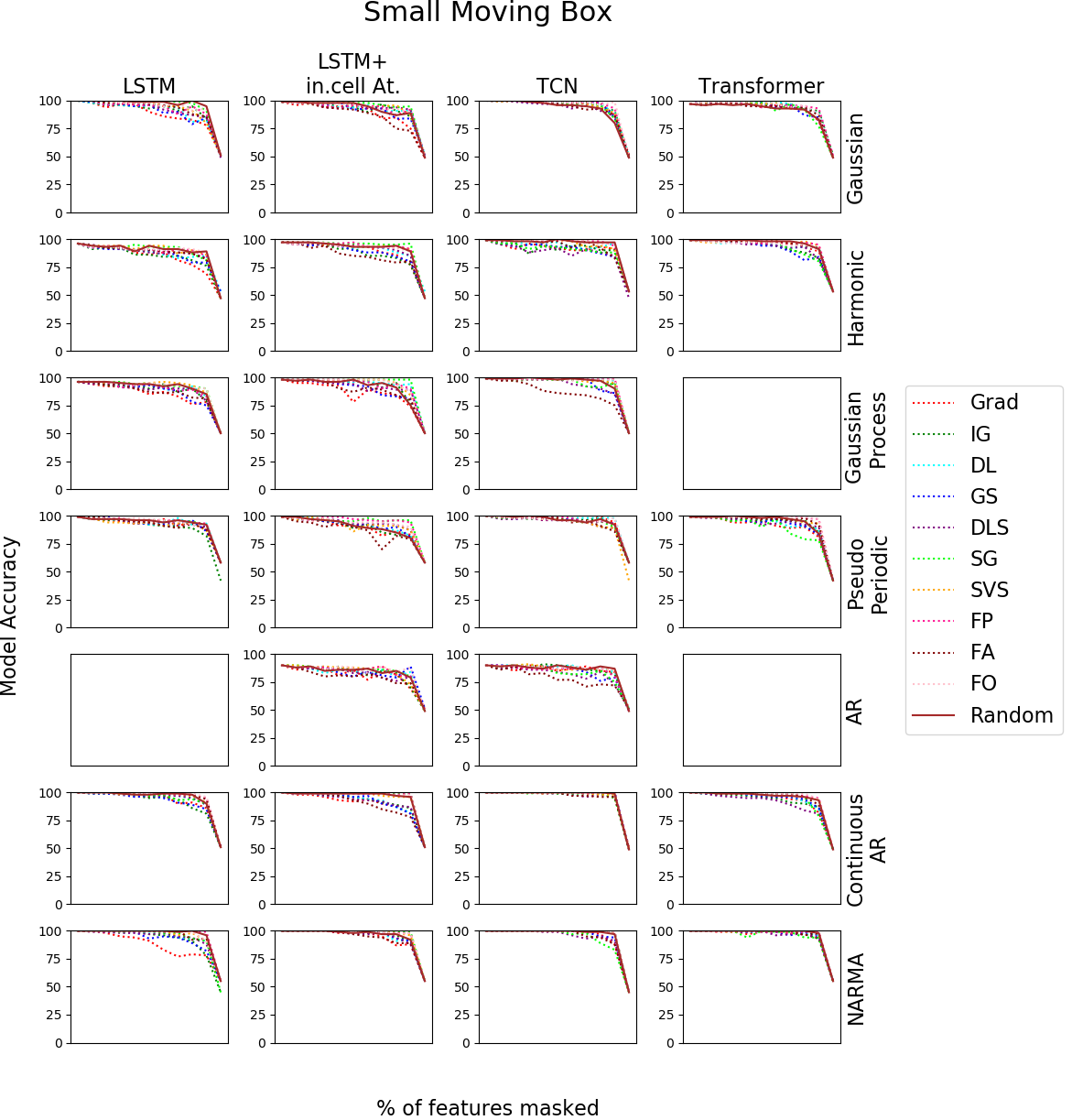} \hfill
\includegraphics[width=.5\textwidth]{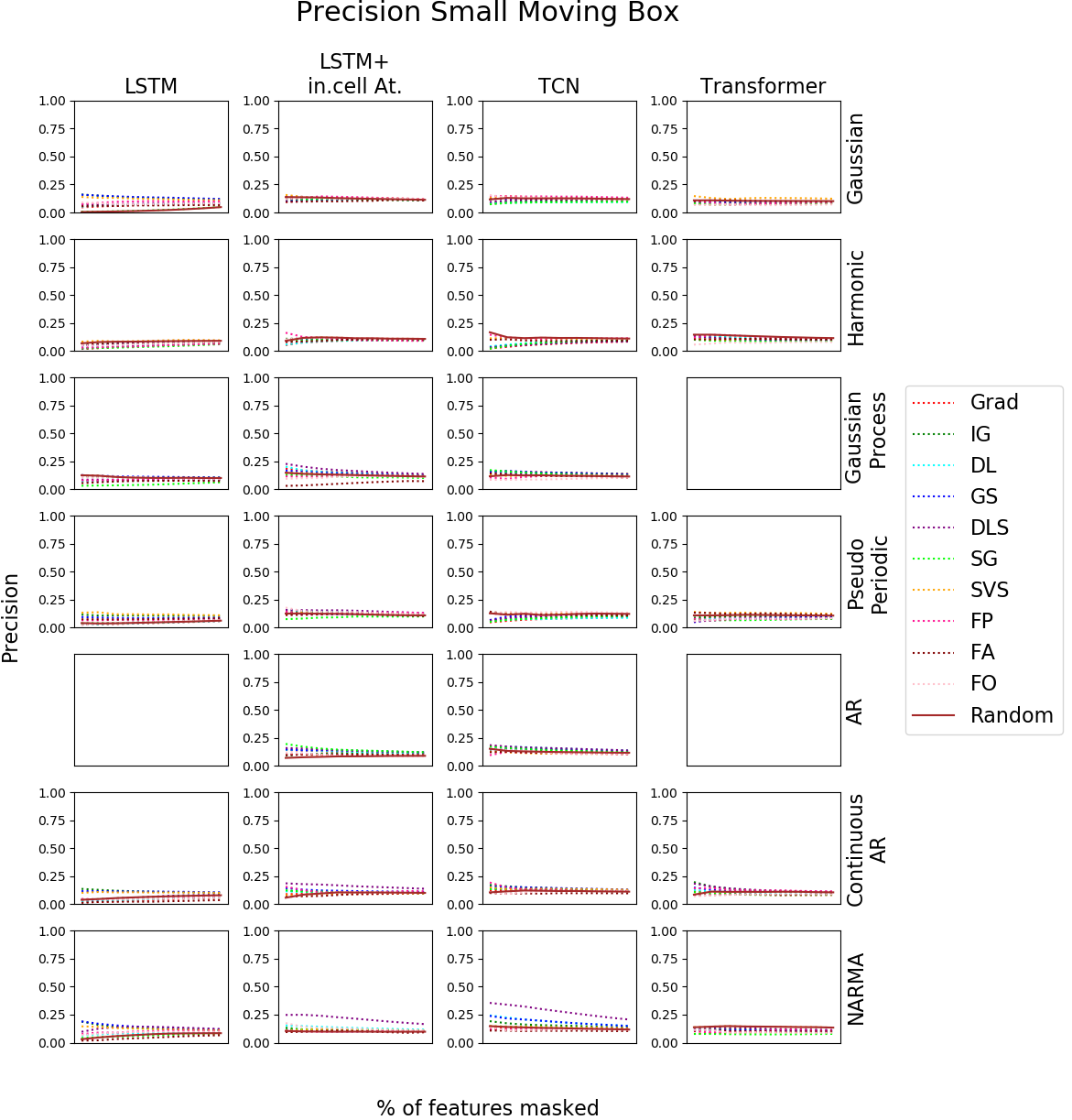}\hfill \includegraphics[width=.5\textwidth]{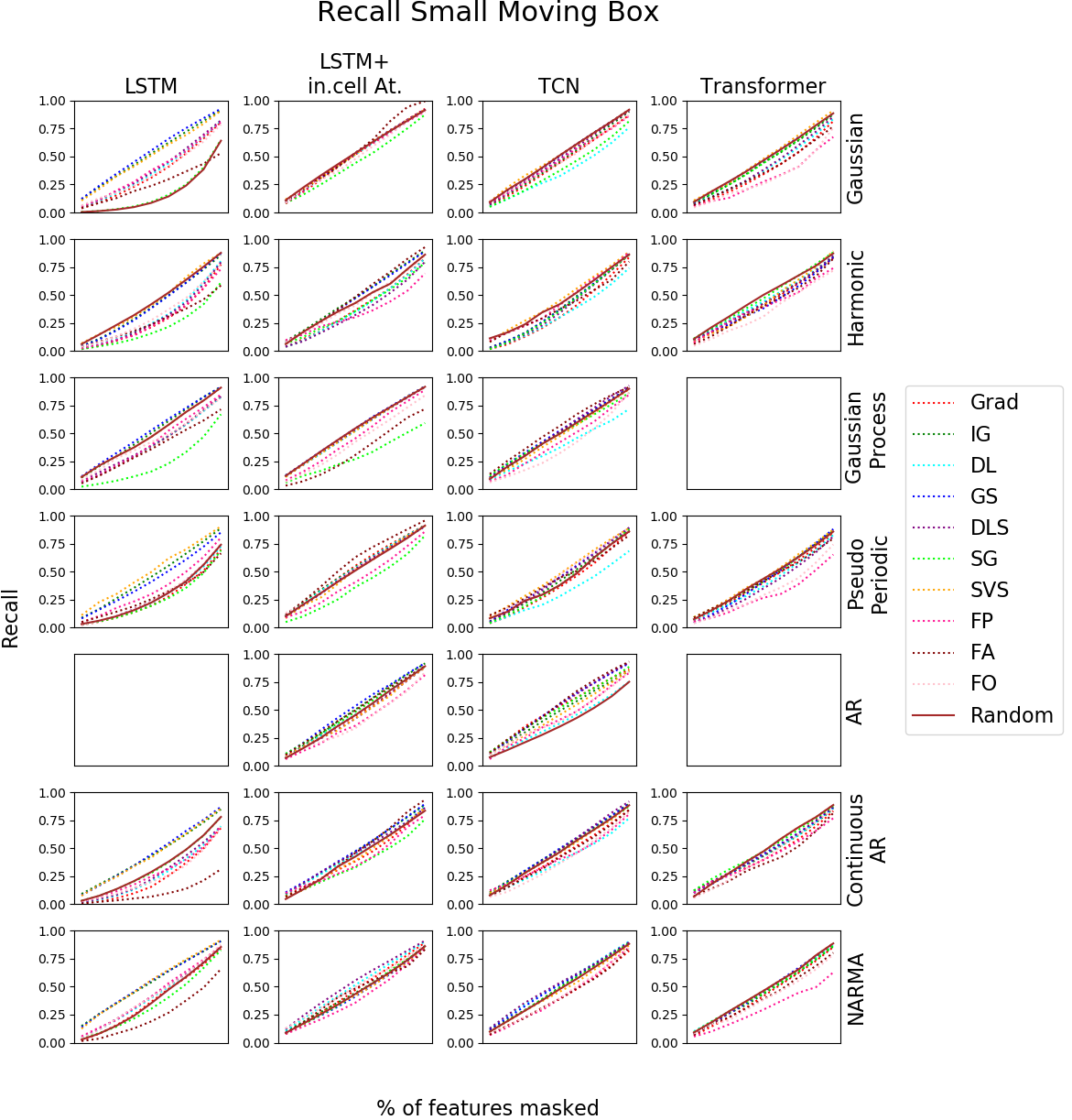}

\caption{Accuracy drop, precision and recall for  \textit{\textbf{Small Moving Middle box}} datasets
}
\label{fig:MovingSmallMiddleBox}
\end{figure}

\begin{figure}[htb!]
\centering
\includegraphics[width=.75\textwidth]{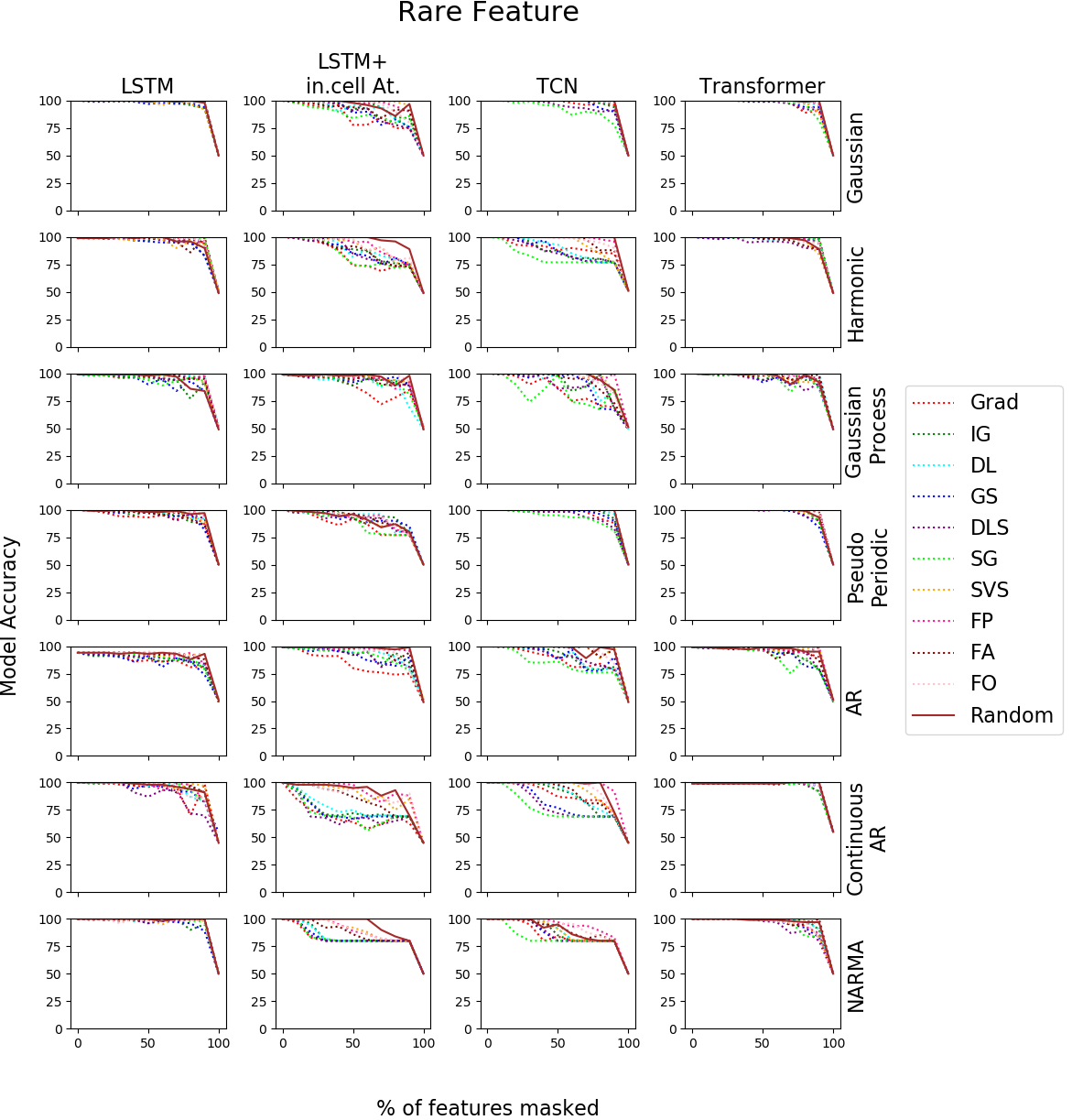} \hfill
\includegraphics[width=.5\textwidth]{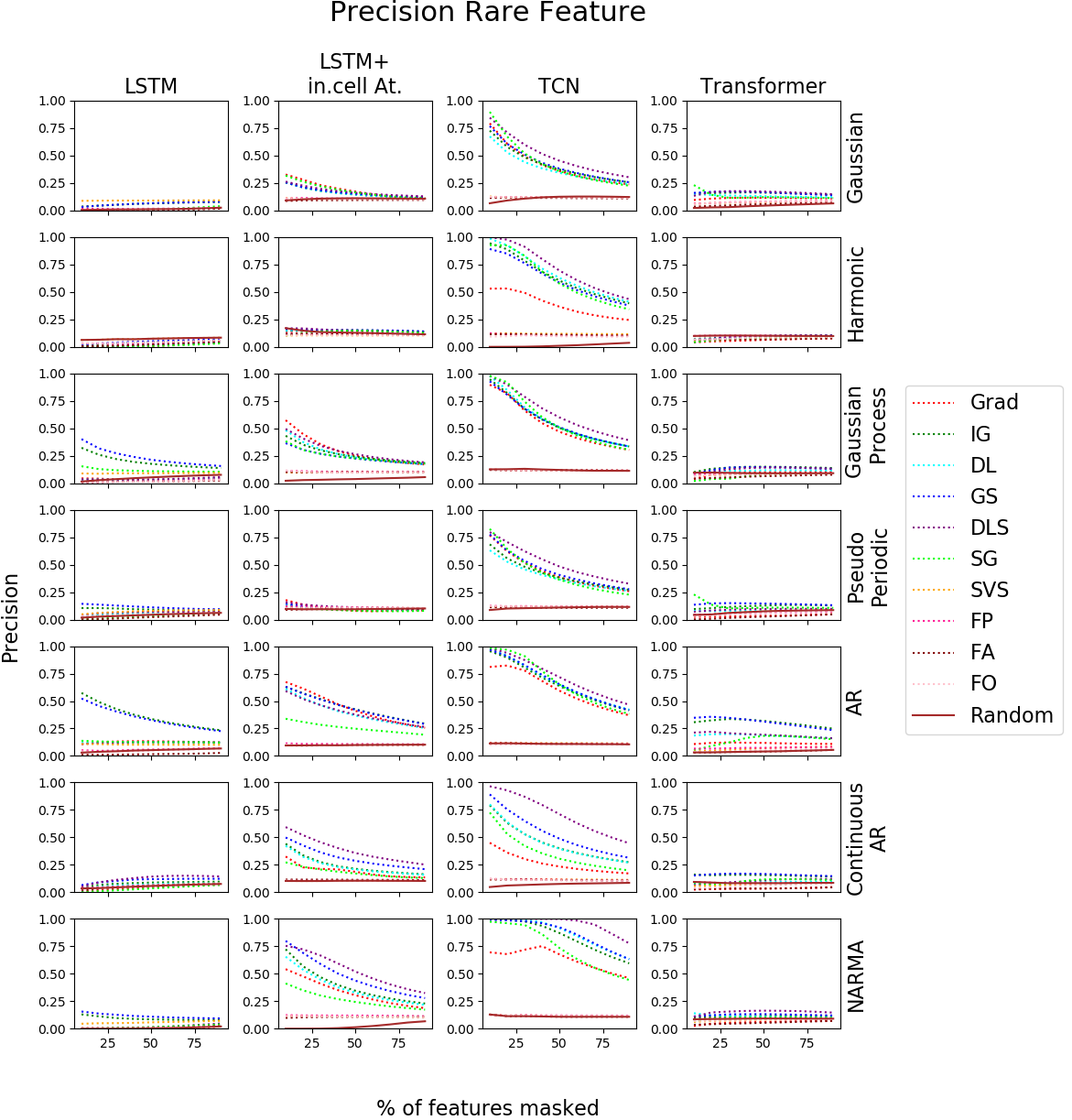}\hfill \includegraphics[width=.5\textwidth]{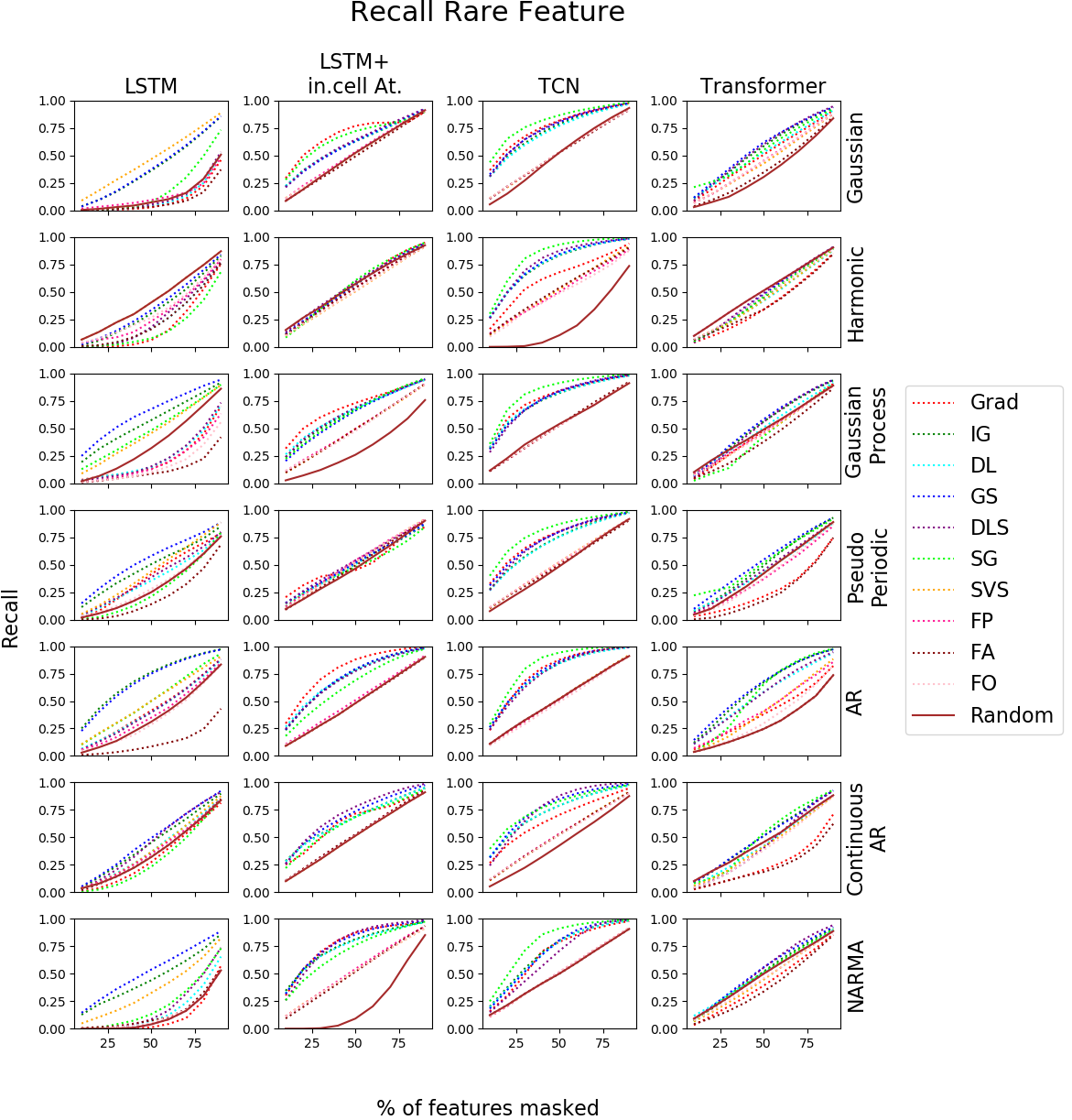}

\caption{Accuracy drop, precision and recall for  \textit{\textbf{Rare Feature}} datasets
}
\label{fig:RareFeature}
\end{figure}

\begin{figure}[htb!]
\centering
\includegraphics[width=.75\textwidth]{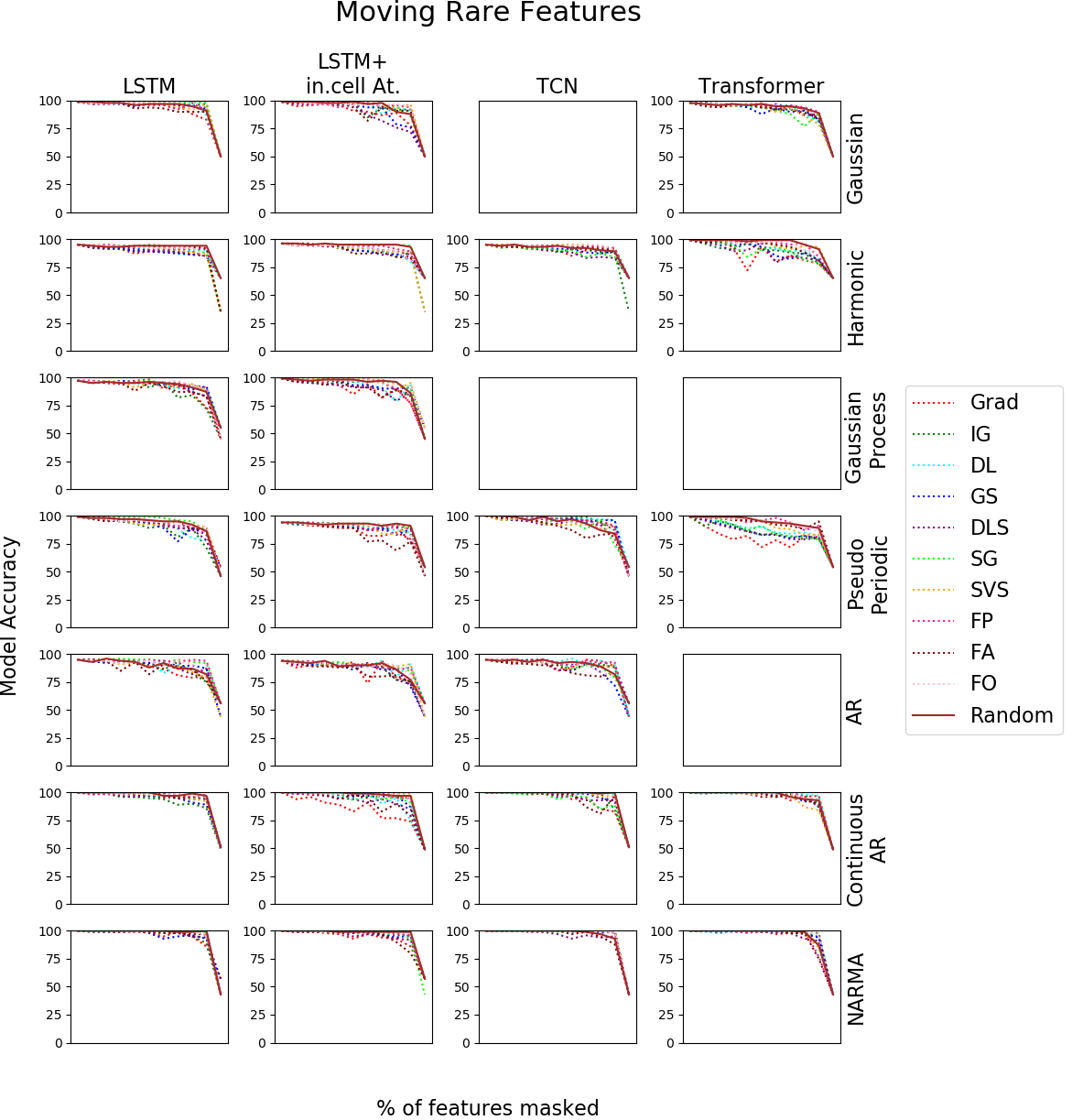} \hfill
\includegraphics[width=.5\textwidth]{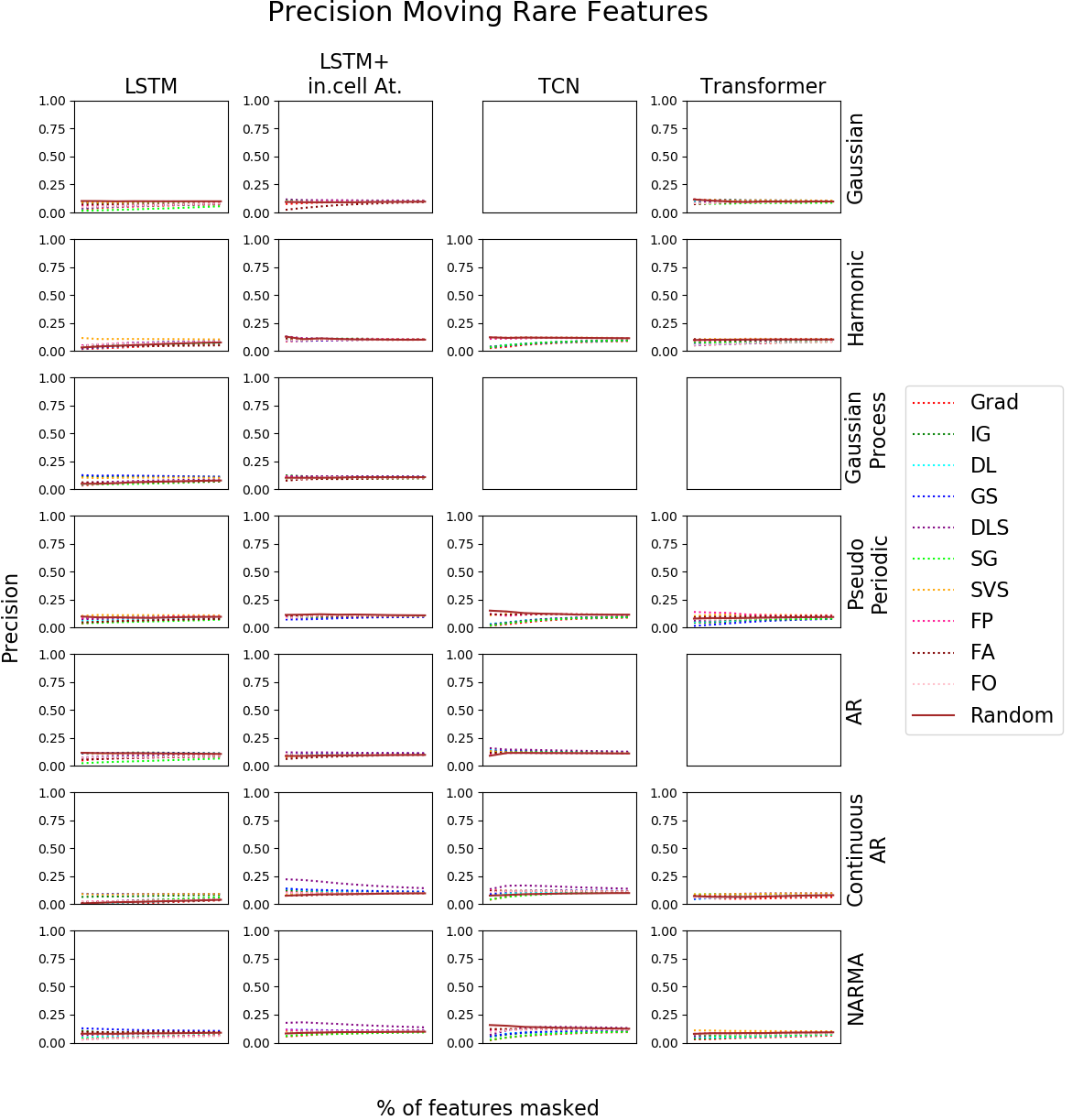}\hfill \includegraphics[width=.5\textwidth]{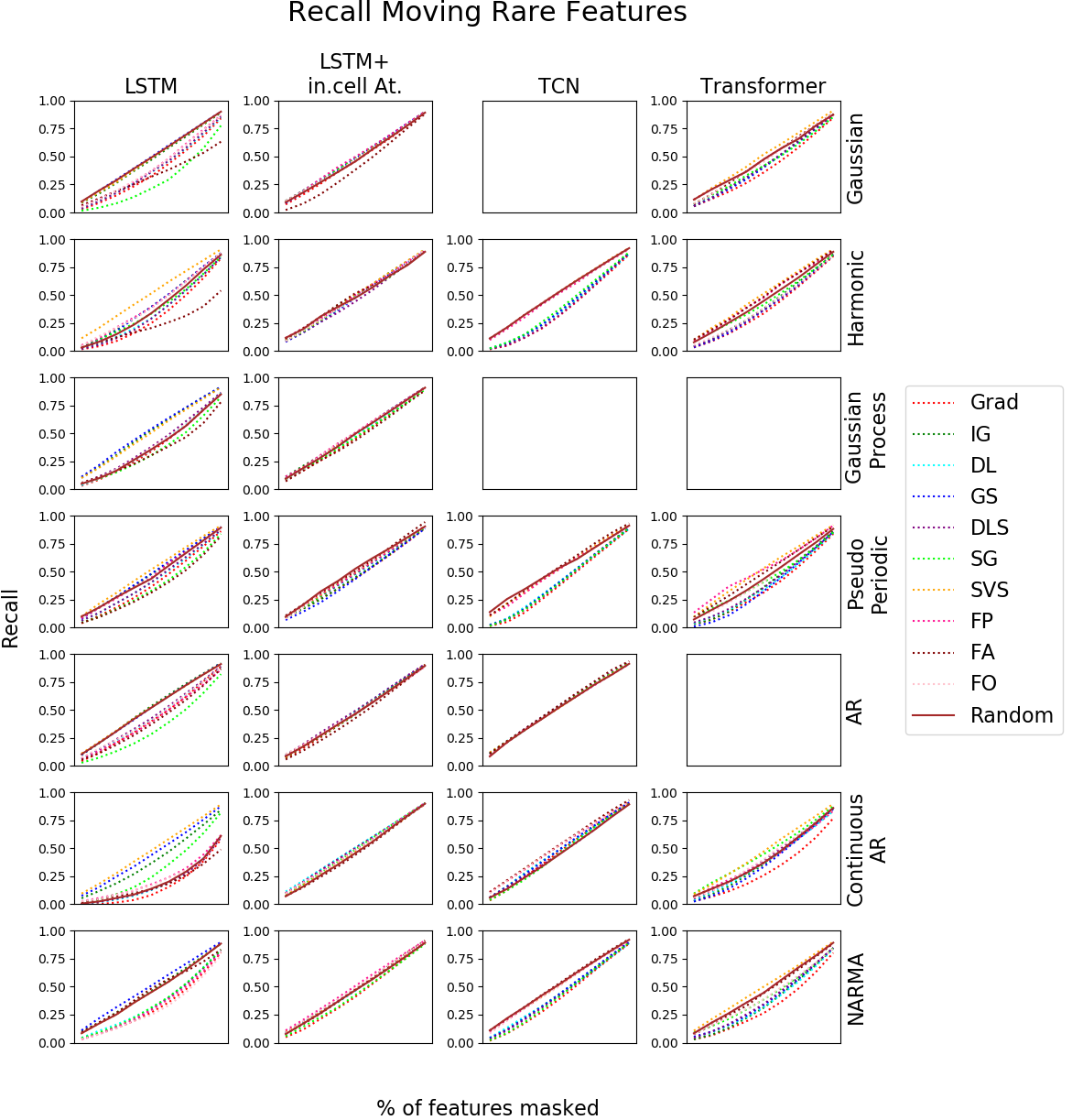}

\caption{Accuracy drop, precision and recall for  \textit{\textbf{Moving Rare Feature}} datasets
}
\label{fig:MovingRareFeature}
\end{figure}

\begin{figure}[htb!]
\centering
\includegraphics[width=.75\textwidth]{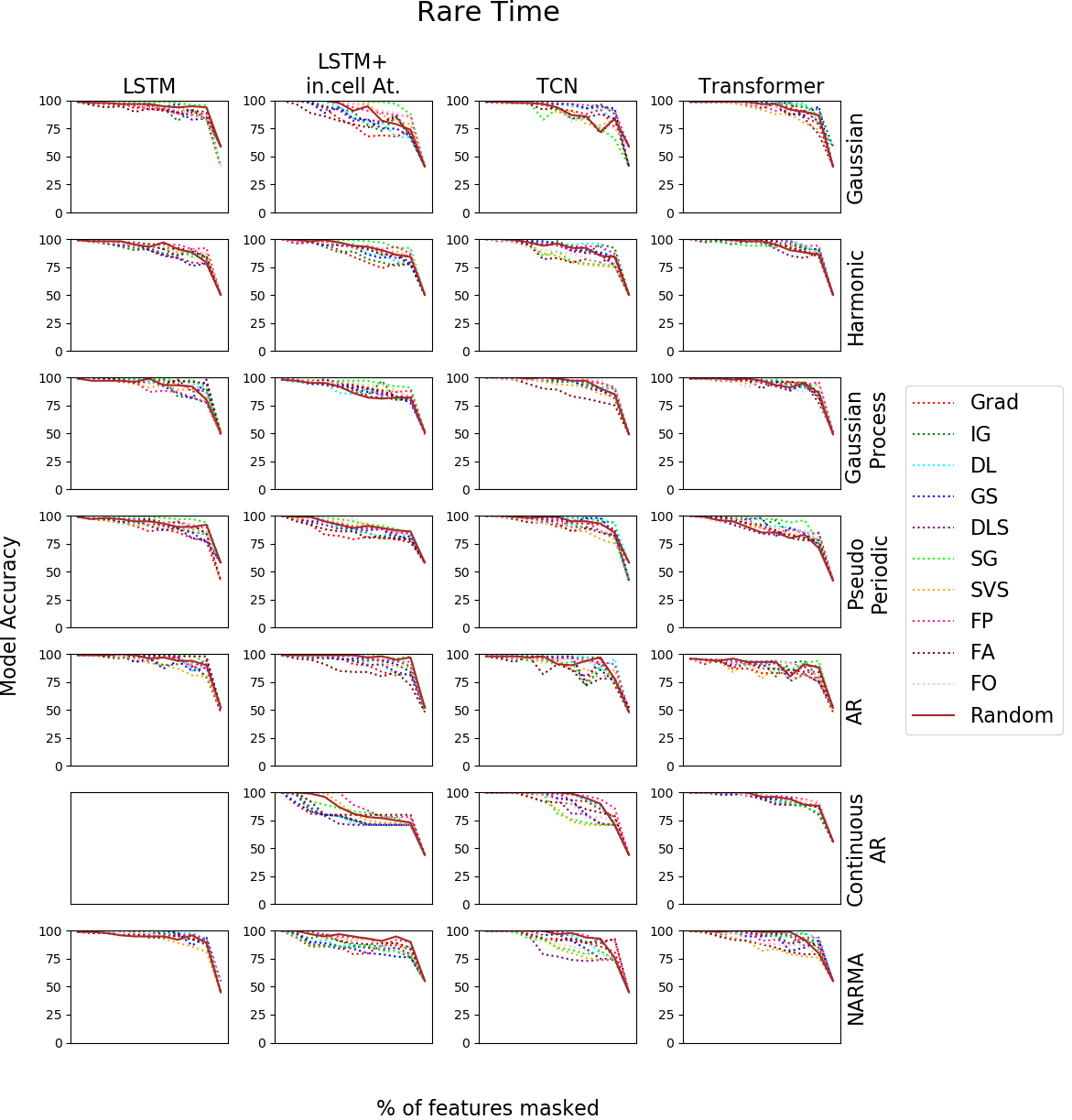} \hfill
\includegraphics[width=.5\textwidth]{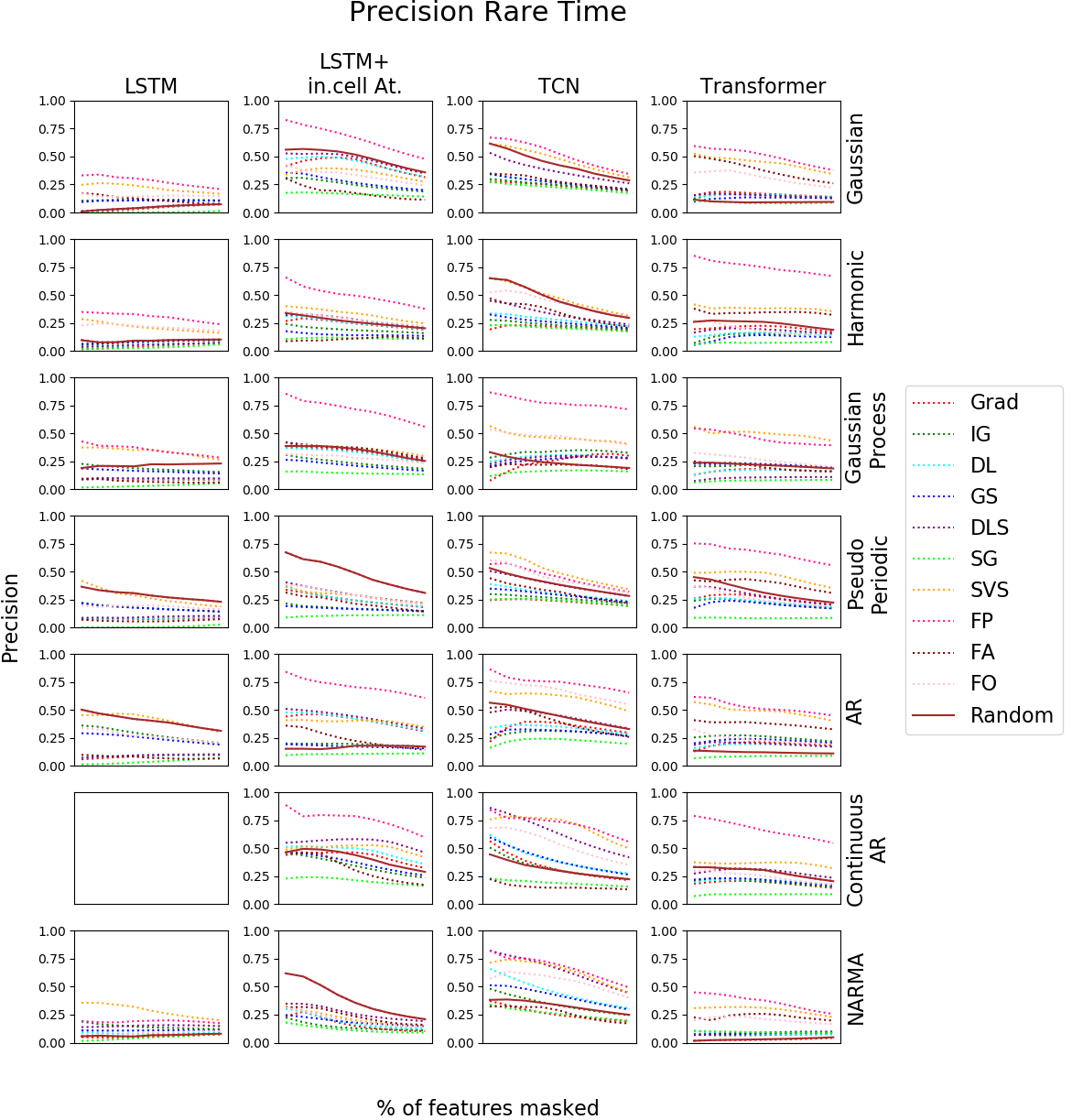}\hfill \includegraphics[width=.5\textwidth]{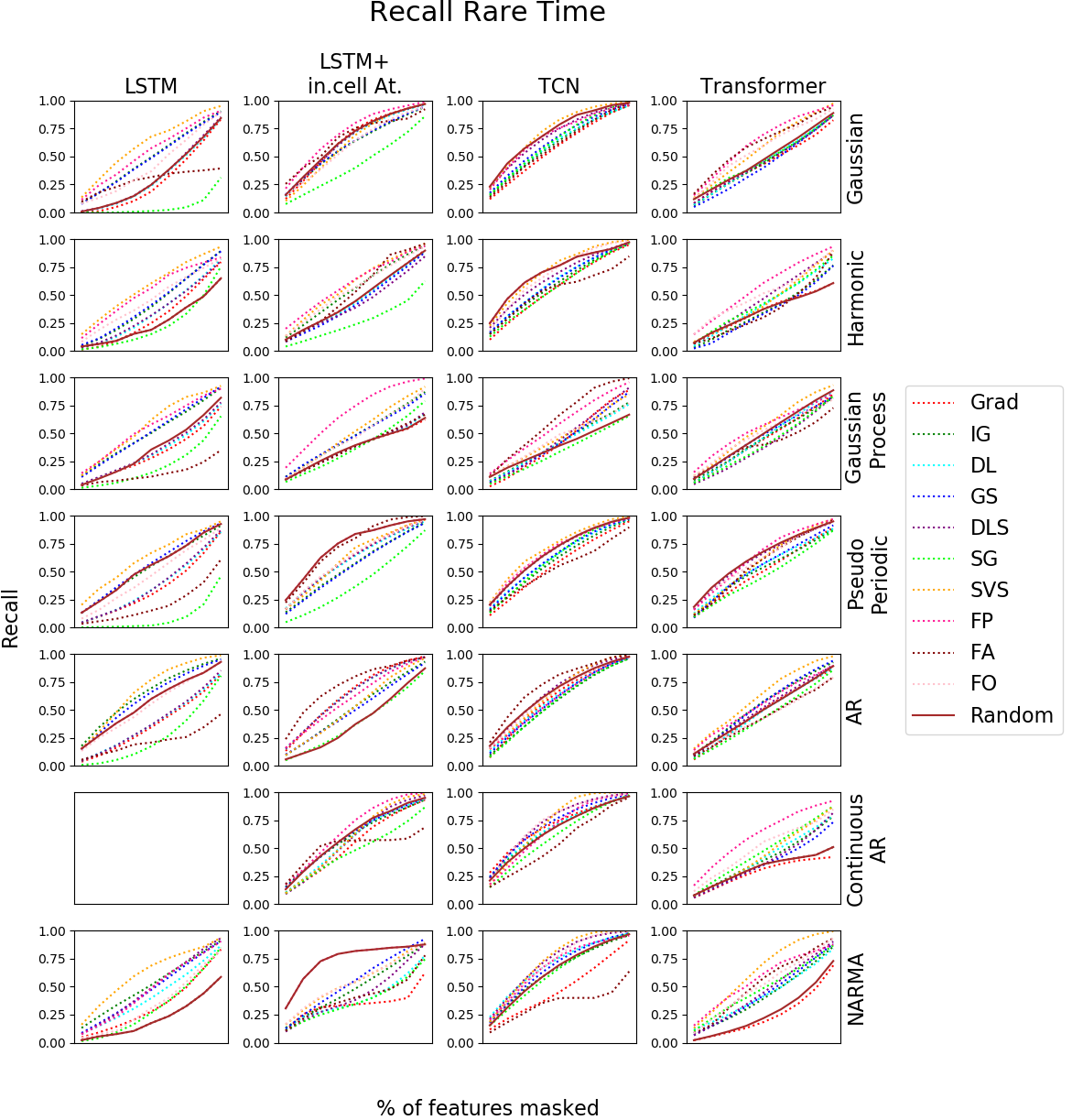}

\caption{Accuracy drop, precision and recall for  \textit{\textbf{Rare Time}} datasets
}
\label{fig:RareTime}
\end{figure}

\begin{figure}[htb!]
\centering
\includegraphics[width=.75\textwidth]{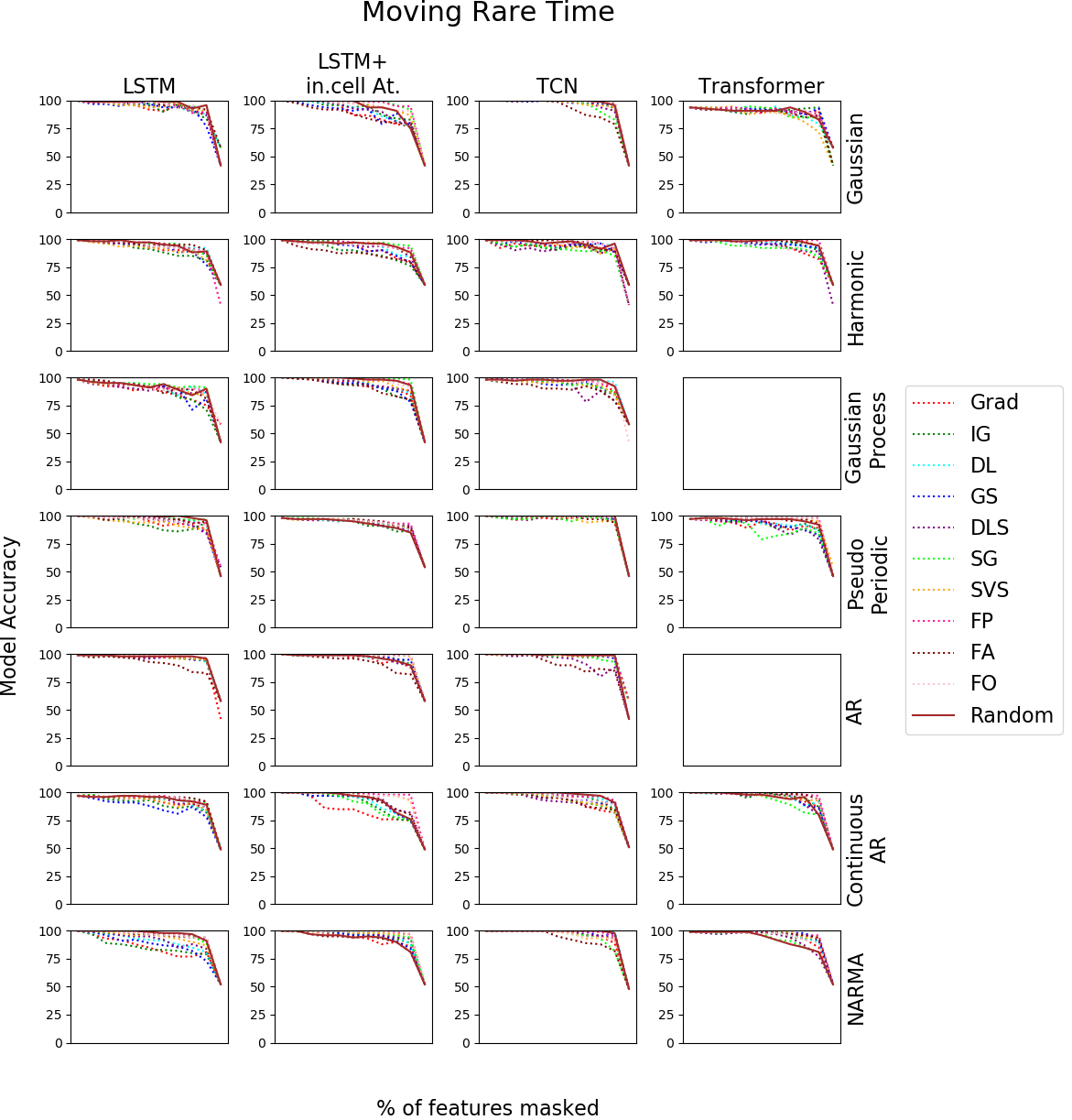} \hfill
\includegraphics[width=.5\textwidth]{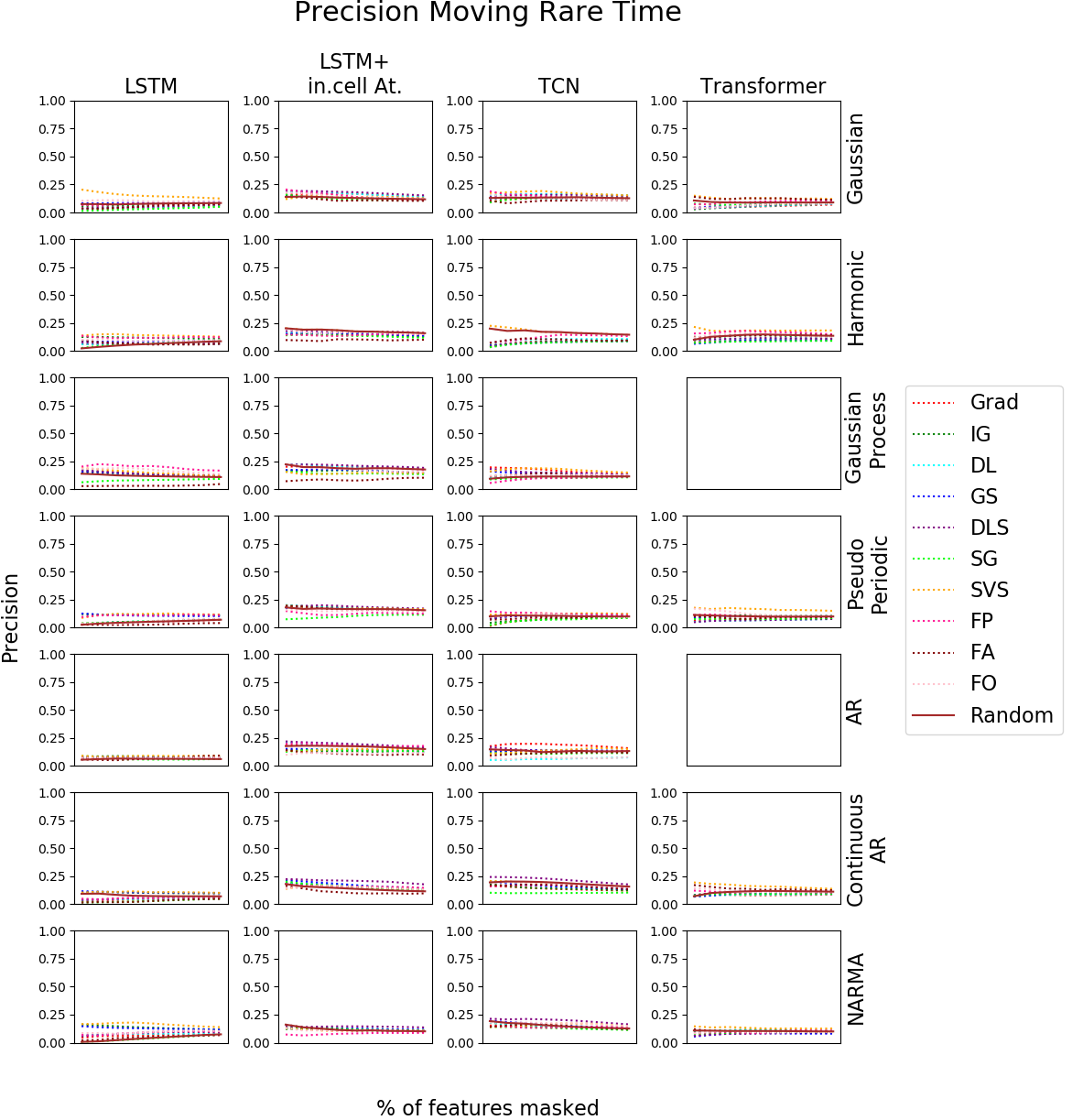}\hfill \includegraphics[width=.5\textwidth]{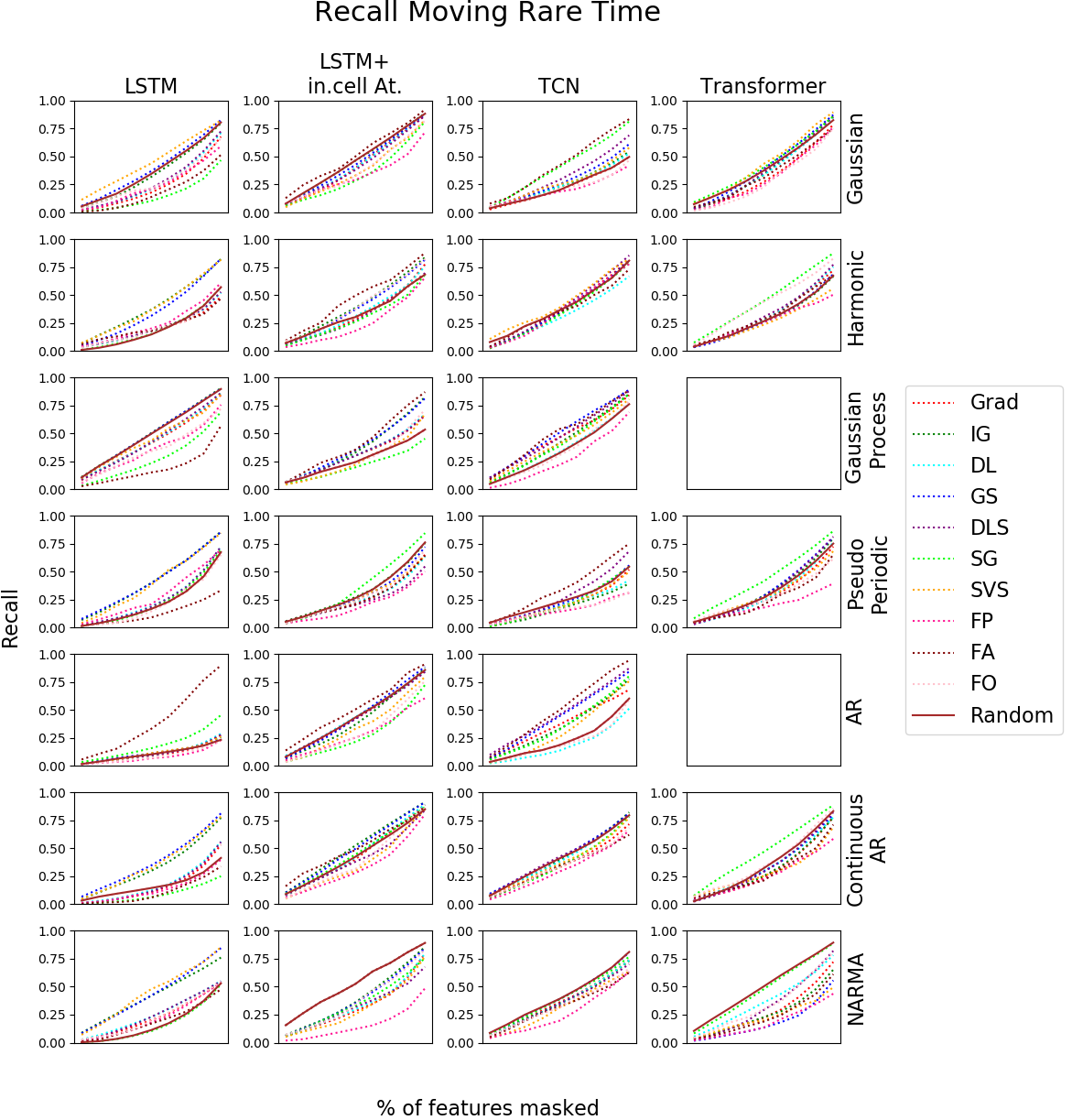}

\caption{Accuracy drop, precision and recall for  \textit{\textbf{Moving Rare Time}} datasets
}
\label{fig:MovingRareTime}
\end{figure}

\begin{figure}[htb!]
\centering
\includegraphics[width=.75\textwidth]{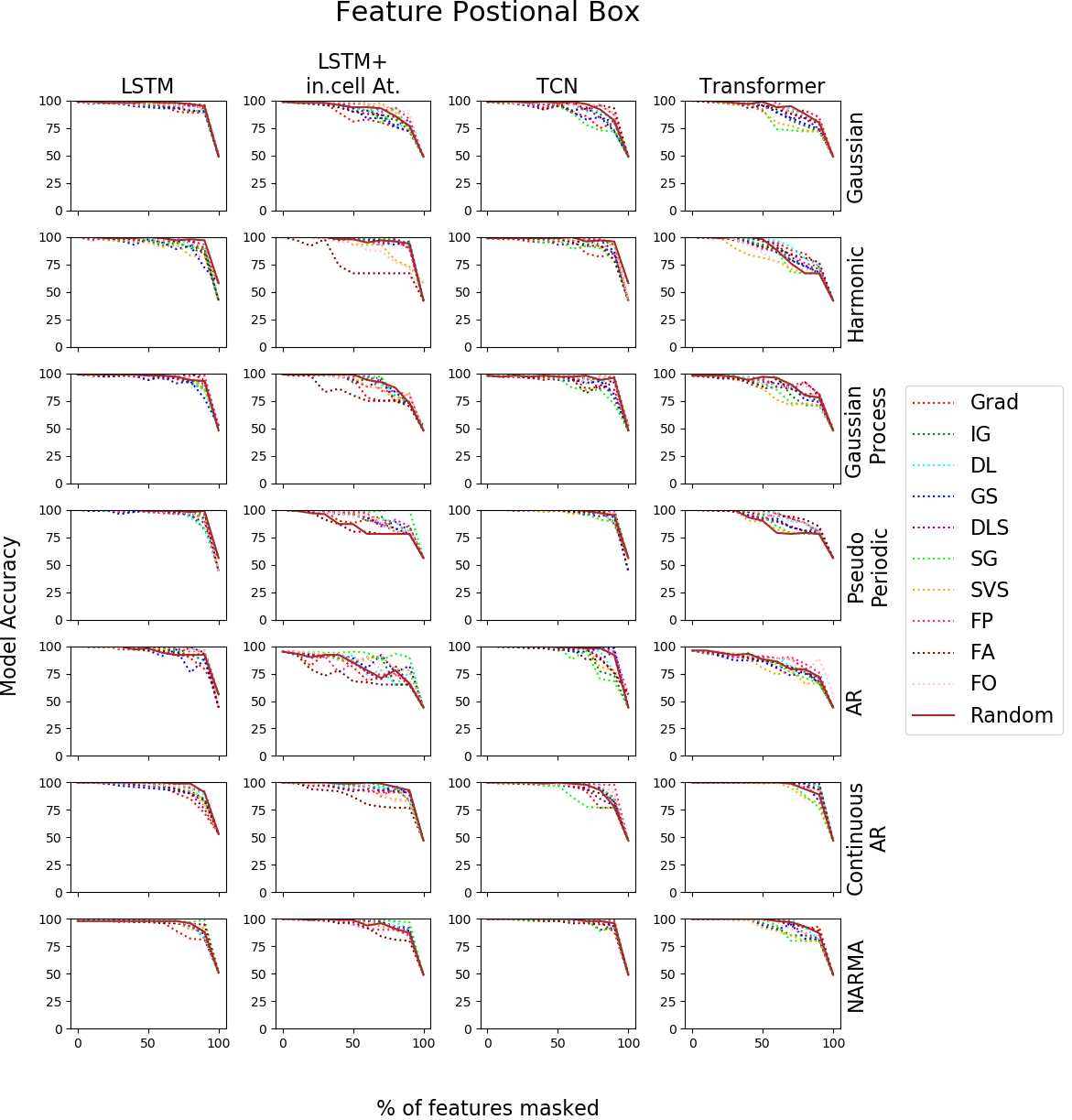} \hfill
\includegraphics[width=.5\textwidth]{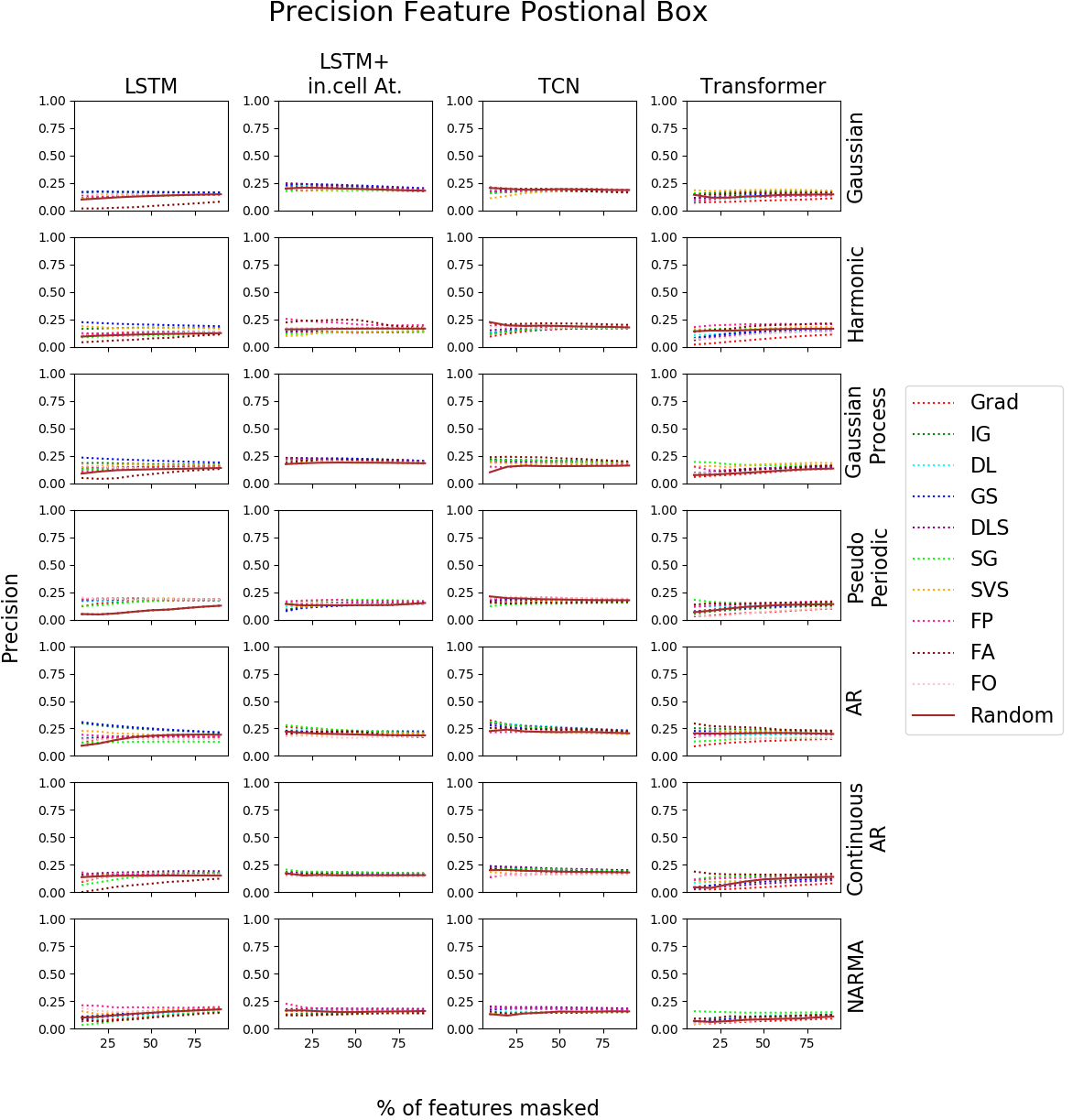}\hfill \includegraphics[width=.5\textwidth]{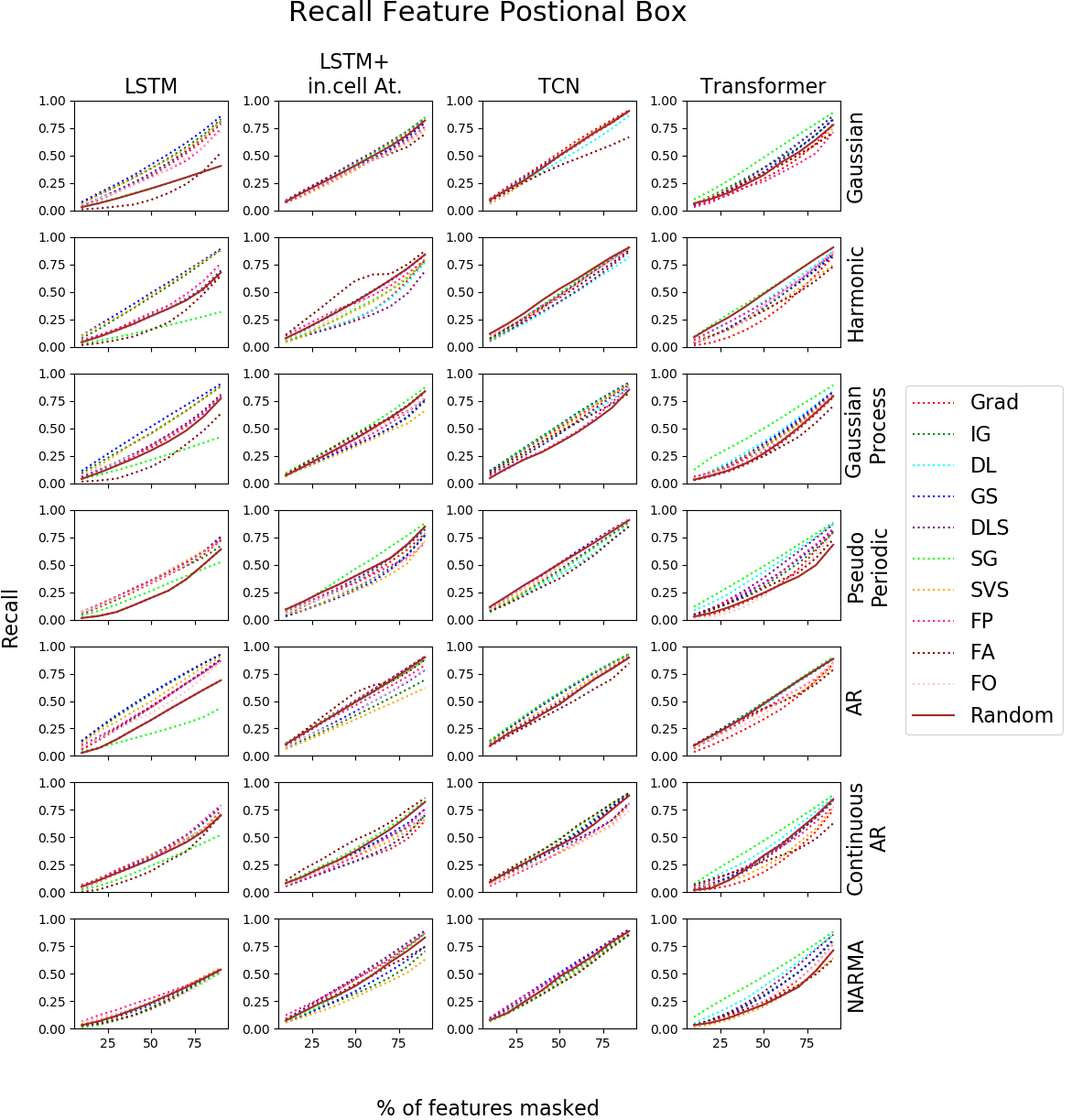}

\caption{Accuracy drop, precision and recall for  \textit{\textbf{Positional Feature}} datasets
}
\label{fig:PostionalFeature}
\end{figure}

\begin{figure}[htb!]
\centering
\includegraphics[width=.75\textwidth]{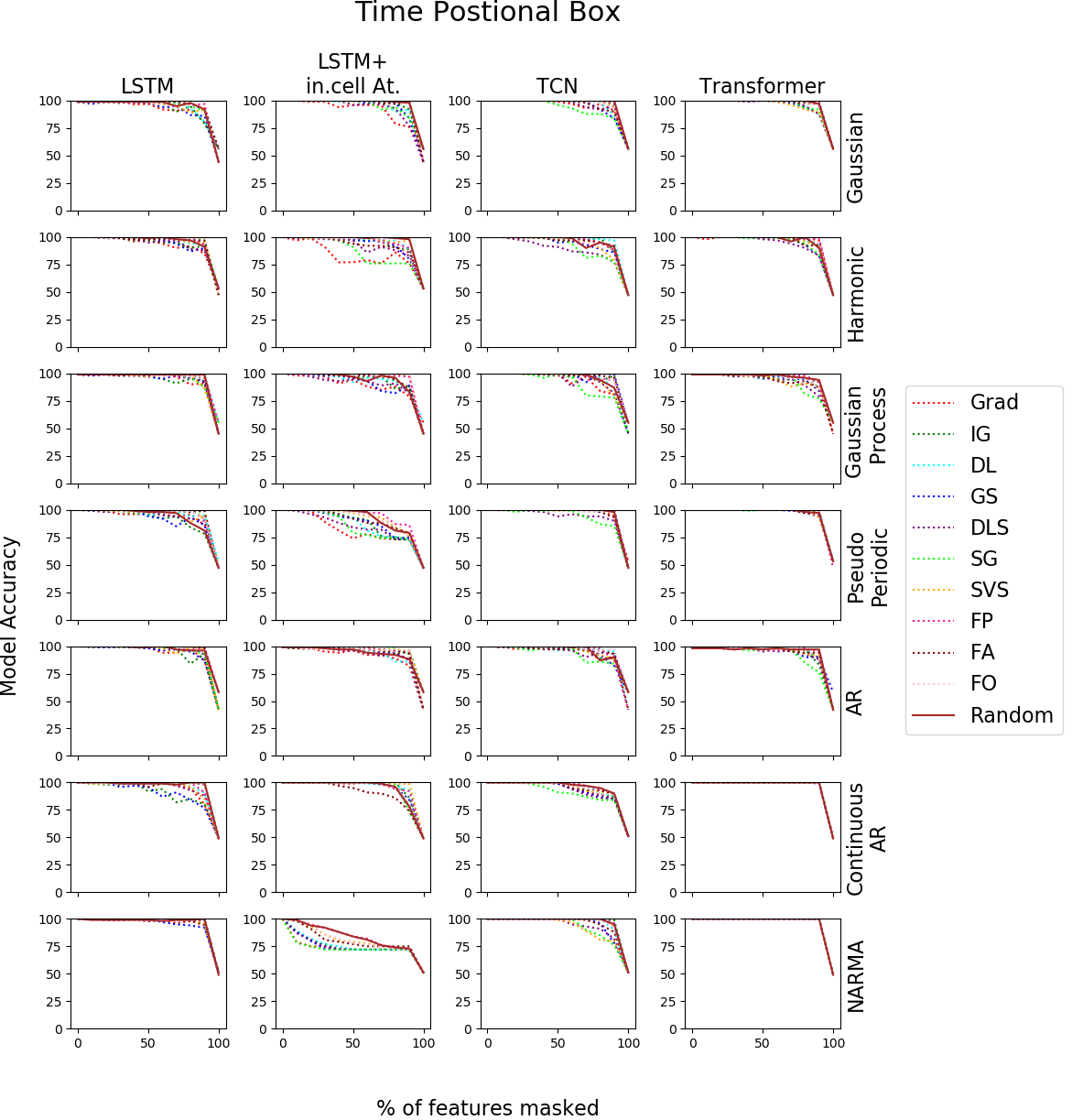}\hfill
\includegraphics[width=.5\textwidth]{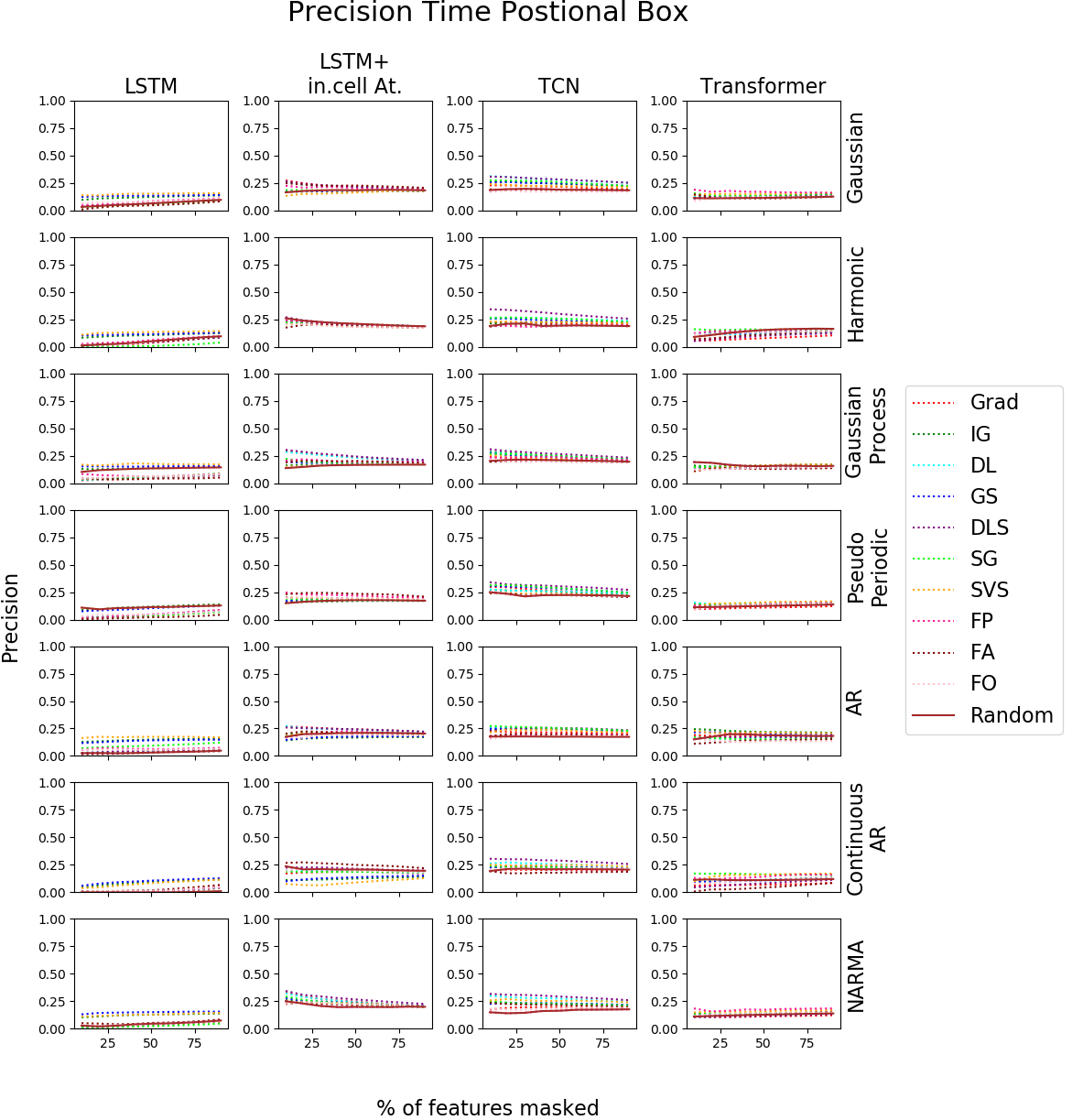}\hfill \includegraphics[width=.5\textwidth]{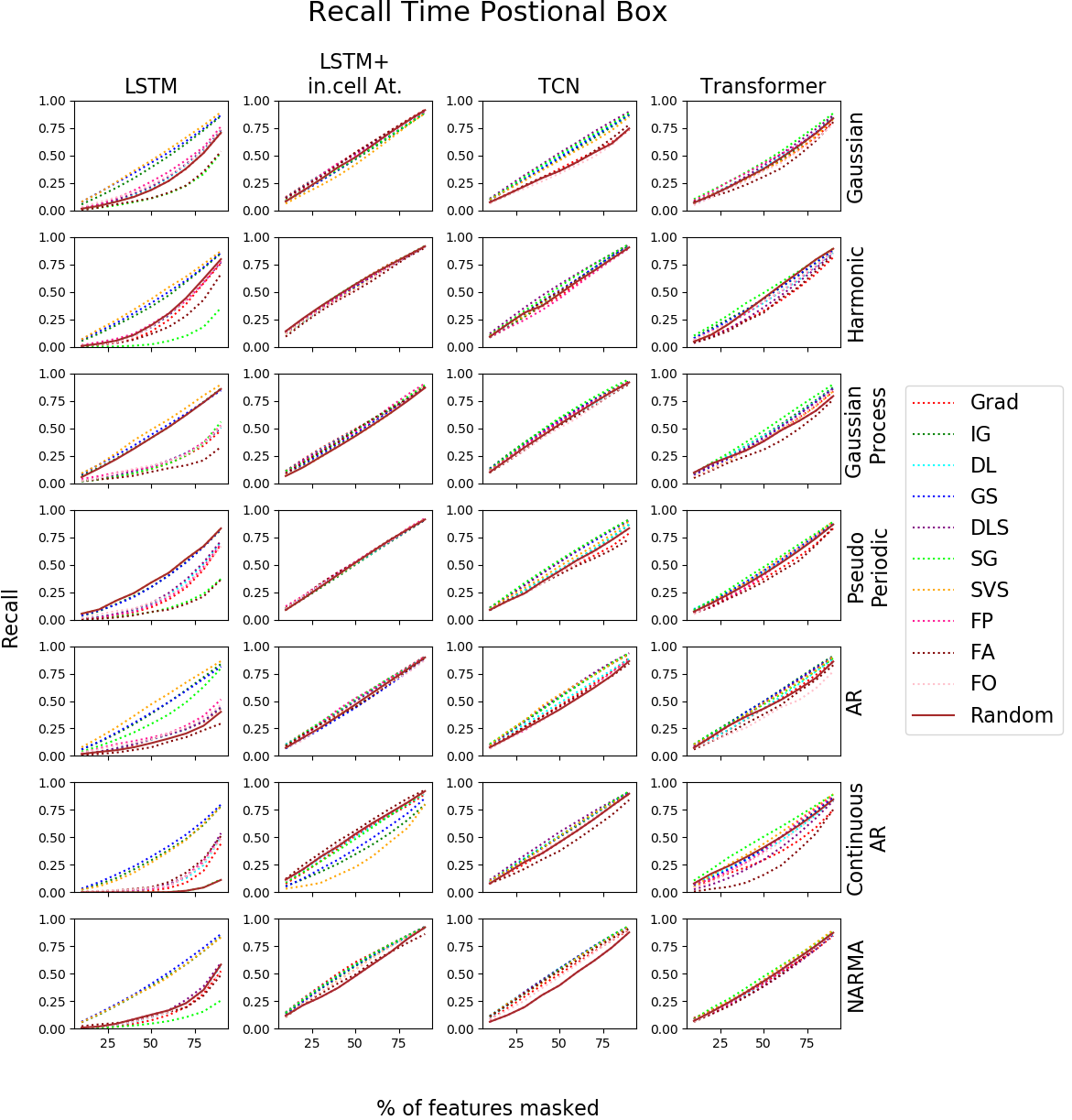}

\caption{Accuracy drop, precision and recall for  \textit{\textbf{Positional Time}} datasets
}
\label{fig:PostionalTime}
\end{figure}

\paragraph{Model Accuracy Drop, Precision and Recall}
The effect of masking salient features on the model accuracy is shown in the first row of  Figures [\ref{fig:MiddleBox}-\ref{fig:PostionalTime}]. Similarly, precision and recall at different levels of degradtion are shown in second row of  Figures [\ref{fig:MiddleBox}-\ref{fig:PostionalTime}].


\newpage

\newpage
\newpage

\newpage
\newpage

\newpage
\newpage

\newpage
\newpage

\FloatBarrier

\subsubsection*{Saliency Maps for Images versus Multivariate Time Series}
\begin{figure*}[htb!]
\centering
\includegraphics[width=0.85\textwidth]{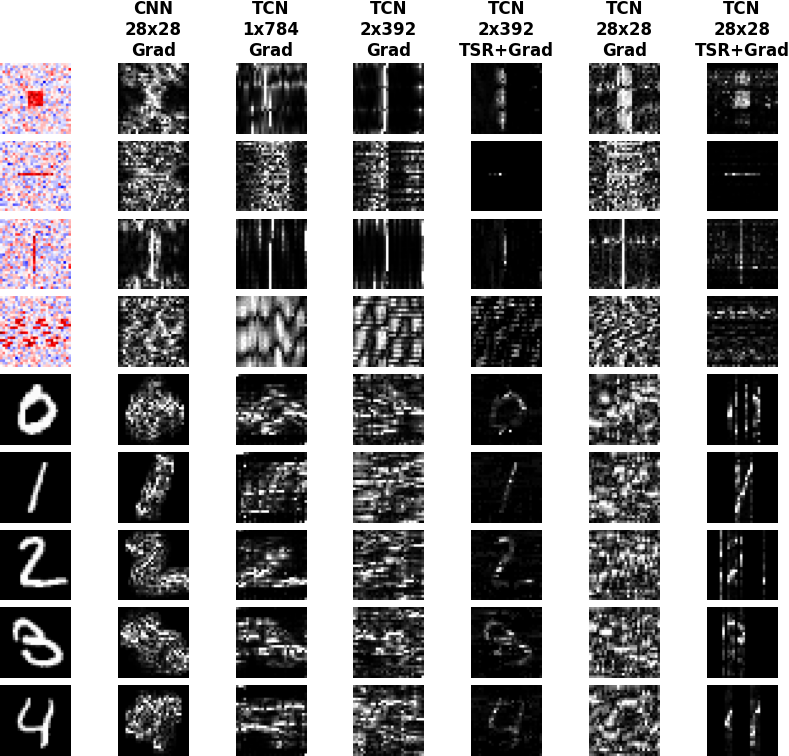}
\caption{Saliency Maps for samples when treated as an image (CNN) versus univariate (1 feature x 784 time steps), bivariate (2 features x 392 time steps), or multivariate  (28 features x 28 time steps) time series (TCN) before and after applying TSR.}
\label{fig:uniVSMulti}
\end{figure*}

Figure \ref{fig:uniVSMulti} shows a few examples of saliency maps produced by the various treatment approaches of the same sample (images for CNN, uni, bi, multivariate time series for TCN). One can see that CNN and univariate TCN produce interpretable maps. In contrast, the maps for the bivariate and multivariate Grad are harder to interpret, applying the proposed temporal saliency rescaling approach on bivariate and multivariate time series significantly improves the quality of saliency maps and in some cases even better than images or univariate time series.

\end{document}